\documentclass[journal]{IEEEtran}
\usepackage{cite}

\ifCLASSINFOpdf
\else
\fi
\usepackage{amsmath}
\usepackage{algorithmic}

\usepackage{array}

\ifCLASSOPTIONcompsoc
 \usepackage[caption=false,font=normalsize,labelfont=sf,textfont=sf]{subfig}
\else
 \usepackage[caption=false,font=footnotesize]{subfig}
\fi
\usepackage{url}

\usepackage{graphicx}

\usepackage{xcolor}
\usepackage{amsmath}
\usepackage{hyperref}
\usepackage{amssymb}
\usepackage{cleveref}

\usepackage{algorithm}

\usepackage{multicol}

\usepackage{changepage}

\usepackage{makecell}

\usepackage{scrextend}

\usepackage{diagbox}

\renewcommand{\algorithmiccomment}[1]{{\color{blue}\texttt{\bgroup\hfill//~#1\egroup}}}

\newlength{\minuslength}
\DeclareMathOperator*{\argmin}{arg\,min}

\newcommand*{\tran}{^{\mkern-1.5mu\mathsf{T}}}

\begin{document}
%

\title{Hyperparameter Learning under Data Poisoning: Analysis of the Influence of Regularization via Multiobjective Bilevel Optimization}

%
%
%

\author{Javier~Carnerero-Cano,~\IEEEmembership{Student Member,~IEEE,}
         Luis~Mu\~noz-Gonz\'alez, Phillippa Spencer,
        and~Emil~C.~Lupu
\thanks{J. Carnerero-Cano, L. Mu\~noz-Gonz\'alez and E. C. Lupu are with Imperial College London, South Kensington Campus,  London, SW7 2AZ, United Kingdom. E-mail: \{j.cano, l.munoz, e.c.lupu\}@imperial.ac.uk}\protect\\
\thanks{P. Spencer is with the Defence Science and Technology Laboratory (Dstl), Porton Down, Salisbury, United Kingdom.}
}

%
%

\markboth{IEEE TRANSACTIONS ON NEURAL NETWORKS AND LEARNING SYSTEMS}
{IEEE TRANSACTIONS ON NEURAL NETWORKS AND LEARNING SYSTEMS}
%



\maketitle

\begin{abstract}
  Machine Learning (ML) algorithms are vulnerable to poisoning attacks, where a fraction of the training data is manipulated to deliberately degrade the algorithms' performance. Optimal attacks can be formulated as bilevel optimization problems and help to assess their robustness in worst-case scenarios. We show that current approaches, which typically assume that hyperparameters remain constant, lead to an overly pessimistic view of the algorithms' robustness and of the impact of regularization. We propose a novel optimal attack formulation that considers the effect of the attack on the hyperparameters and models the attack as a \emph{multiobjective} bilevel optimization problem. This allows to formulate optimal attacks, learn hyperparameters and evaluate robustness under worst-case conditions. We apply this attack formulation to several ML classifiers using $L_2$ and $L_1$  regularization. Our evaluation on multiple datasets shows that choosing {an ``a priori'' constant} value for the regularization hyperparameter can be detrimental to the performance of the algorithms. This confirms the limitations of previous strategies and evidences the benefits of using $L_2$ and $L_1$ regularization to dampen the effect of poisoning attacks, when hyperparameters are learned using a small trusted dataset.
Additionally, our results show that the use of regularization plays an important robustness and stability role in complex models, such as Deep Neural Networks, where the attacker can have more flexibility to manipulate the decision boundary.
\end{abstract}


%
\IEEEpeerreviewmaketitle

\section{Introduction}
%
%
%
%

\IEEEPARstart{I}{n} many applications,  Machine Learning (ML) systems rely on data collected from \emph{untrusted} data sources, such as humans, machines, sensors, or IoT devices that can be compromised and manipulated. Malicious data from these compromised sources can then be used to poison the learning algorithms themselves. These scenarios expose ML algorithms to data poisoning attacks, where adversaries manipulate a fraction of the training data to subvert the learning process, either to decrease its overall performance or to produce a particular kind of error in the system \cite{barreno2010security, huang2011adversarial, munoz2019challenges}. Poisoning attacks can also facilitate subsequent evasion attacks or produce \emph{backdoor} (or \emph{Trojan}) attacks \cite{gu2019badnets, liu2018trojaning, xiang2020detection, li2022backdoor, liu2022adaptive, jiang2022critical}.

Several systematic optimal poisoning attacks have already been proposed to analyze different families of ML algorithms under worst-case scenarios, including Support Vector Machines (SVMs) \cite{biggio2012poisoning}, other linear classifiers \cite{xiao2015feature, mei2015using, koh2018stronger, carnerero2021regularization}, and neural networks \cite{koh2017understanding, munoz2017towards, huang2020metapoison}. These attack strategies are formulated as a bilevel optimization problem, i.e., an optimization problem that \emph{depends} on another optimization problem. In these cases, the attacker typically aims to maximize a malicious objective (e.g., to maximize the error for a set of target points) by manipulating a fraction of the training data. At the same time, the defender aims to optimize a different objective function to learn the model's parameters, typically by minimizing some loss function evaluated on the poisoned training set.

Some of the previous attacks target algorithms that have hyperparameters, but the hyperparameters are considered constant regardless of the fraction of poisoning points injected in the training dataset. This can provide a misleading analysis of the robustness of the algorithms against such attacks, as the value of the hyperparameters can change depending on the type and strength of the attack. For example, Xiao \MakeLowercase{\textit{et al.}} \cite{xiao2015feature} presented a poisoning attack against embedded feature selection methods, including $L_1$, $L_2$ and \emph{elastic-net} regularization. Their results show that the attacker can completely control the selection of the features to significantly increase the overall test error of linear classifiers. However, they assume a constant regularization hyperparameter regardless of the attack considered. We show that this approach provides overly pessimistic results on the ML algorithms' robustness to poisoning attacks.

In our prior work \cite{carnerero2021regularization} we reported a limited case study that  analyzes the influence of the $L_2$ regularization hyperparameter on the effect of poisoning attacks against Logistic Regression (LR). In this paper we provide a comprehensive analysis of the influence of the hyperparameters on the effect of poisoning attacks when using different regularization techniques, including $L_2$ and $L_1$ regularization. We also propose a more general optimal indiscriminate poisoning attack formulation to test worst-case scenarios against ML algorithms that contain hyperparameters. For this, we model the attack as a \emph{multiobjective bilevel optimization problem}, where the outer objective includes both the learning of the poisoning points and that of the hyperparameters, while the inner problem involves learning the model's parameters. This attack formulation allows us to model an adversary aware not only of the training algorithm, but also of the procedure used to select the model's hyperparameters. Thus, this formulation considers a more realistic attacker and allows to assess in a more comprehensive way the robustness of the algorithms to poisoning attacks in worst-case scenarios.  In scenarios where the attacker is aware of the dataset used to learn the model's hyperparameters and aims to maximize the overall error, the outer objective can be modeled as a \emph{minimax} problem.   

We used \emph{hypergradient} (i.e., the gradient in the outer problem \cite{maclaurin2015gradient, franceschi2017forward, franceschi2018bilevel, grazzi2020iteration}) descent/ascent to solve the multiobjective bilevel optimization problem. As the computation of the exact hypergradients can be computationally expensive, especially for neural networks, we used Reverse-Mode Differentiation (RMD) \cite{domke2012generic,maclaurin2015gradient, franceschi2017forward, munoz2017towards,franceschi2018bilevel, grazzi2020iteration} to approximate the hypergradients.  We conduct an exhaustive experimental analysis on Logistic Regression (LR) and Deep Neural Networks
 (DNNs), using different datasets including MNIST \cite{lecun1998gradient}, Fashion-MNIST (FMNIST) \cite{xiao2017fashion} and CIFAR-10 \cite{krizhevsky2009learning}, and attacks with both small and large fractions of poisoning points.\footnote{The PyTorch implementation of the algorithms used for the experiments is available at \href{https://github.com/javiccano/hyperparameter-learning-and-poisoning---tnnls/}{https://github.com/javiccano/hyperparameter-learning-and-poisoning---tnnls/}.}

We show that choosing (\emph{a priori}) a constant value for the regularization hyperparameter, $\lambda$,  can be detrimental: if the value is too high it damages accuracy ({i.e.,} it produces underfitting when there is no attack), if the value is too low it damages robustness (the algorithm is more brittle in the presence of an adversary). In contrast, selecting $\lambda$ appropriately by, for example, using a small trusted validation set, provides both accuracy and robustness regardless of the presence or absence of poisoning points in the training dataset and of the attack strength. Our empirical evaluation also reveals that the value of the regularization hyperparameter increases with the number of poisoning points injected in the training set. The algorithm automatically tries to compensate the negative effect of the poisoning points by increasing the strength of the regularization term. For the DNNs, we show that the attack can have a more pronounced effect in the later layers of the network, and that the use of different regularization hyperparameters for the different layers in the DNN can be beneficial to mitigate the impact of data poisoning. In the case of embedded feature selection methods, we confirm the stabilizing effect of regularization against poisoning.

The rest of the paper is organized as follows: In Sect.~\ref{sec:related} we describe the related work. In Sect.~\ref{sec:generalAttacks} we introduce our novel formulation for optimal poisoning attacks against learning algorithms with hyperparameters. In Sect.~\ref{sec:L2} we discuss how regularization can help mitigate poisoning attacks by enhancing algorithms' stability. In Sect.~\ref{sec:experiment} we present our experimental evaluation on different datasets. Finally, Sect.~\ref{sec:conclusion} concludes the paper. 

\section{Related Work} 
\label{sec:related}

The first poisoning attacks reported in the literature targeted specific applications, such as spam filtering \cite{nelson2008exploiting,barreno2010security} or anomaly detection \cite{barreno2006can,kloft2012security}. A more systematic approach was introduced by Biggio \MakeLowercase{\textit{et al.}} \cite{biggio2012poisoning} to poison SVMs, modeling the attack as a bilevel optimization problem. Subsequent works extended this approach to other families of ML algorithms, including linear and other convex classifiers \cite{mei2015using} or embedded feature selection methods \cite{xiao2015feature}. A more general approach was introduced by Mu\~noz-Gonz\'alez \MakeLowercase{\textit{et al.}} \cite{munoz2017towards}, formulating different optimal attack strategies for targeting multiclass classifiers. The authors also proposed an algorithm to estimate the hypergradients in the corresponding bilevel optimization problem through Reverse-Mode Differentiation (RMD), which significantly improves the scalability of optimal attacks, allowing to poison a broader range of learning algorithms, including neural networks. Koh \MakeLowercase{\textit{et al.}} \cite{koh2018stronger} proposed an algorithm for solving bilevel problems with detectability constraints, allowing to craft poisoning points that can bypass outlier detectors. However, the algorithm is computationally demanding, which limits its applicability in practical scenarios. None of the previous approaches consider the effect and influence of the hyperparameters on the learning algorithm when the training dataset is poisoned. 

Other approaches have also been proposed for crafting poisoning attacks: Koh \MakeLowercase{\textit{et al.}} \cite{koh2017understanding} created adversarial training examples by exploiting influence functions. This approach allows to craft successful targeted attacks by injecting small perturbations to genuine data points in the training set. Shafahi \MakeLowercase{\textit{et al.}} \cite{shafahi2018poison}, Zhu \MakeLowercase{\textit{et al.}} \cite{zhu2019transferable}, Huang \MakeLowercase{\textit{et al.}} \cite{huang2020metapoison}, and Geiping \MakeLowercase{\textit{et al.}} \cite{geiping2021witches} proposed targeted attacks for situations where the adversary does not control the labels of the poisoning points. A Generative Adversarial Net-based model to craft indiscriminate and targeted poisoning attacks at scale against deep networks was proposed in \cite{munoz2019poisoning}. This approach allows to naturally model detectability constraints for the attacker, enabling attacks with different levels of ``aggressiveness" to bypass different types of defenses. 

On the defender's side, it is possible to mitigate poisoning attacks by analyzing the samples that have a negative impact on the target algorithms \cite{nelson2008exploiting}. However, this approach can be impractical in many applications, as it scales poorly. Following a similar approach, Koh \MakeLowercase{\textit{et al.}} \cite{koh2017understanding} propose to use influence functions as a mechanism to detect poisoning points. Different outlier detection schemes have proved to be effective to mitigate poisoning attacks in those cases where the attacker does not consider appropriate detectability constraints \cite{steinhardt2017certified,paudice2018detection}. Label sanitization has also been proposed as a mechanism to identify and relabel suspicious training points \cite{paudice2018label, zhang2018training}. However, this strategy can fail when the poisoning points ``collude" \cite{munoz2019poisoning}. Finally, Diakonikolas \MakeLowercase{\textit{et al.}} \cite{diakonikolas2019sever} proposed a robust meta-algorithm, based on Singular Value Decomposition, capable of mitigating some attacks.

Koh \MakeLowercase{\textit{et al.}} \cite{koh2018stronger} reported some results on poisoning attacks against a linear SVM using $L_2$ regularization. Their results suggest that, in some cases, increasing the value of the regularization hyperparameter can make outlier detectors less effective. However, a direct comparison with our results is not possible as we consider a different threat model. Compared to \cite{koh2018stronger}, we provide a more general formulation to model the effect of hyperparameters, including the case of $L_2$ regularization, in the presence of data poisoning. Furthermore, we provide a more complete and systematic evaluation of the benefits of using regularization to mitigate the effect of poisoning attacks, under reasonable assumptions.

\section{General Optimal Poisoning Attacks}

\label{sec:generalAttacks}

\begin{table*}[!t]

	\centering
	\caption{List of notations.}{ 
		\begin{tabular}{|l|l|}
			\hline
			Symbol &  Description \\
			\hline
				 $\nabla_{(\cdot)}(\cdot)$,  $\nabla_{(\cdot)}\nabla_{(\cdot)}(\cdot)$, $\nabla_{(\cdot)}^2(\cdot)$ & Gradient, Hessian matrix, Hessian matrix\\
			 $\frac{\partial(\cdot)}{\partial(\cdot)}$, $\frac{d(\cdot)}{d(\cdot)}$ & Partial derivative, Total derivative \\

							$(\cdot)_\text{tr}$	$(\cdot)_\text{val}$, $(\cdot)_\text{p}$, $(\cdot)_\text{target}$ & Reference to   training, validation, poisoning, and target datasets \\
							
								$n \in \mathbb{N}$, $m \in \mathbb{R}$, $c \in \mathbb{N}$  &  Number of datapoints, Number of input features, Number of classes  \\
								
			$d  \in \mathbb{N}$, 		$d_s \in \mathbb{N}$, $h \in \mathbb{N}$ & Number of model's parameters,   		  Dimension of training state, Number of model's hyperparameters\\

			$\mathcal{D} = \{ ({\bf x}_i, y_i) \}_{i = 1}^n \in \mathbb{R}^{n \times m} \times \mathbb{Z}_c^n$	 &    Dataset \\

								$\mathcal{D}_\text{p}' \in \Phi(\mathcal{D}_\text{p})$	& Constrained poisoning dataset \\
						
												$\mathcal{D}_\text{tr}'= \mathcal{D}_\text{tr} \cup \mathcal{D}_\text{p}' \in \mathbb{R}^{n_\text{tr} \times m} \times \mathbb{Z}_c^{n_\text{tr}}$	& Poisoned training dataset \\

 			  			 ${\bf x} \in \mathbb{R}^m$, ${y} \in \mathbb{Z}_c$ & Input sample vector, Output sample scalar \\
 	  			  			  			 ${\bf X}  \in \mathbb{R}^{n \times m}$, ${\bf y} \in \mathbb{Z}_c^n$ & Input samples matrix, Output samples vector \\
 	  			  			  			  			 $\mathcal{X}\subseteq \mathbb{R}^m$, $\mathcal{Y}\subseteq \mathbb{Z}_c$  & Input space,  Output space \\

 			  			  			  			$p(\mathcal{X}, \mathcal{Y}): \mathcal{X} \times \mathcal{Y} \rightarrow \mathbb{R}_{\geq 0}$  & Underlying probability distribution \\
 			  			  			  			$f:\mathcal{X} \rightarrow \mathcal{Y}$  & Mapping between input space and output space \\
			 ${\bf w}\in \mathbb{R}^d$, ${\bf w}^\star \in \mathbb{R}^d$, ${\bf s} \in \mathbb{R}^{d_s}$ & Model's parameters, Optimal model's parameters,  Training state\\

			 $\mathcal{W} \subseteq \mathbb{R}^d$ & Feasible domain for the model's parameters \\

			 $\boldsymbol{\Lambda} \in \mathbb{R}^h$, $\lambda \subseteq \mathbb{R}^h$  & Model's hyperparameters, Regularization hyperparameters    \\
			 		
			 			 			 			 $\lambda_\text{RMD} \subseteq \mathbb{R}^h$ & Regularization hyperparameters learned through reverse-mode differentiation  \\
			 			 			 			 
			 			 			 			 $\lambda_\text{CLEAN} \subseteq \mathbb{R}^h$ & Regularization hyperparameters learned through $5$-fold cross-validation on the clean dataset \\

			 			 $\Phi(\mathcal{D}_\text{p}) \subseteq \mathbb{R}^{m} \times \mathbb{Z}_c$ & Feasible domain set of the poisoning datapoints  \\
			 			 			 			 $\Pi_{\Phi(\mathcal{D}_\text{p})}: \mathbb{R}^{n \times m} \times \mathbb{Z}_c^n \rightarrow \mathbb{R}^{n \times m} \times \mathbb{Z}_c^n$ & Projection operator for the feasible domain of the poisoning datapoints \\
			 			 			 			 
			 			 			 			 $\Phi(\boldsymbol{\Lambda}) \subseteq \mathbb{R}^{h}$ & Feasible domain set of the hyperparameters  \\
			 			 			 			 $\Pi_{\Phi(\boldsymbol{\Lambda})}: \mathbb{R}^{h}  \rightarrow \mathbb{R}^{h}$ & Projection operator for the feasible domain of the hyperparameters \\
			 			 			 			 
			 			${\bf v} \in \mathbb{R}^d$ & Generic vector \\ 
			 		 $\mathcal{M}: \mathbb{R}^{n \times m} \rightarrow \mathbb{R}^{n \times c}$ & Target model \\
		 $\mathcal{A}: \mathbb{R}^{n \times m} \times \mathbb{Z}_c^n  \rightarrow \mathbb{R}$, $\mathcal{L}: \mathbb{R}^{n \times m} \times {Z}_c^n \rightarrow \mathbb{R}$ & Attacker's objective function, Loss function \\

		 		 $\mathcal{P}=\left\{p_k\right\}_{k=1}^{n_\text{p}} \in \mathbb{N}^{n_\text{p}}: p_k \leq n_\text{tr}$	 & Indices of the $n_\text{p}$ datapoints of $\mathcal{D}_\text{tr}$ to be replaced by $\mathcal{D}_\text{p}$ \\ 
		 		 $p \in \mathbb{N}: p \leq n_\text{tr}$ & Index of  $\mathcal{D}_\text{tr}$ to be replaced by a poisoning datapoint \\
		 			$T \in \mathbb{N}$,  $T_\text{mul} \in \mathbb{N}$	 & Number of training iterations for the inner problem, number of hyperiterations for the outer problem \\
		 			
		 			$T_\text{KKT} \in \mathbb{N}$ &  Number of training iterations for the inner problem that satisfy the stationarity conditions\\
			$\eta  \in (0, 1]$, $\eta_\text{tr}  \in (0, 1]$	 &   Learning rate for the inner problem, learning rate for the inner problem when testing the attack \\
		$\alpha \in (0, 1]$ & Learning rate for the outer problem  \\

						$\kappa  \in \mathbb{R}_{\geq 0}$ & Condition number of the Hessian matrix\\
												$I_c(A, B)  \in \mathbb{R}$ & Kuncheva's consistency index between two feature subsets
$A, B \subseteq \mathcal{X}$\\

$\mathcal{N}(\boldsymbol{\mu}, \boldsymbol{\Sigma}) : \mathcal{X} \times \mathcal{Y} \rightarrow \mathbb{R}_{> 0}$ & Gaussian distribution with mean vector $\boldsymbol{\mu}$ and covariance matrix $\boldsymbol{\Sigma}$\\

			\hline
	\end{tabular}}
	\label{tabNotat}

\end{table*}

In data poisoning attacks the attacker can tamper with a fraction of the training set to manipulate the behavior of the learning algorithm \cite{barreno2006can,barreno2010security}. We assume that the attacker can arbitrarily manipulate all the features and the label of the injected poisoning points, provided that the resulting points are within a feasible domain of valid data points. We consider white-box attacks with perfect knowledge, i.e., the attacker knows everything about the target system, including the training data, the feature representation, the loss function, the ML model, and the defense (if applicable) used by the victim. Although unrealistic in most cases, these assumptions are needed to analyze the robustness of the ML algorithms in worst-case scenarios for attacks of different strengths.  

\subsection{Problem Formulation}

In line with most literature on poisoning attacks we consider ML classifiers. Then, in a classification task, given the input space $\mathcal{X}\subseteq \mathbb{R}^m$ and the discrete label space, $\mathcal{Y} \subseteq \mathbb{Z}_c$, where $c$ is the number of classes, the learner aims to estimate the mapping $f: \mathcal{X} \rightarrow \mathcal{Y}$. Given a training set $\mathcal{D}_\text{tr}= \{({\bf x}_{\text{tr}_i} , y_{\text{tr}_i})\}^{n_\text{tr}}_{i=1}$ with $n_\text{tr}$ IID samples drawn from the underlying
probability distribution $p(\mathcal{X}, \mathcal{Y})$, we can estimate $f$ with a model $\mathcal{M} {: \mathbb{R}^{n_\text{tr} \times m} \rightarrow \mathbb{R}^{n_\text{tr} \times c}}$ trained by minimizing an objective function $\mathcal{L}(\mathcal{D}_\text{tr}, \boldsymbol{\Lambda}, {\bf w}) : \mathbb{R}^{n_\text{tr} \times m} \times {Z}_c^{n_\text{tr}} \rightarrow \mathbb{R}$ w.r.t. its parameters,\footnote{As in \cite{maclaurin2015gradient} we use parameters
to denote ``parameters that are just parameters and not hyperparameters''.} ${\bf w}\in \mathbb{R}^d$, given a set of hyperparameters $\boldsymbol{\Lambda}\in \mathbb{R}^{ h}$.

In  this  paper, we use gradient-based algorithms to optimize the performance of the model on a clean validation set with respect to the hyperparameters \cite{larsen1996design, bengio2000gradient, foo2008efficient, franceschi2017forward, franceschi2018bilevel} and poisoning points \cite{biggio2012poisoning, xiao2015feature, munoz2017towards}. Thus, we assume that the defender has access to a small validation dataset $\mathcal{D}_\text{val}= \{({\bf x}_{\text{val}_j} , y_{\text{val}_j})\}^{n_\text{val}}_{j=1}$ with $n_\text{val}$ trusted data points, representative of the ground-truth underlying data distribution. In practice, it is not uncommon to have access to a limited clean set, for example, because the integrity of a small set of data sources can be ascertained.\footnote{Note that if the quality of the trusted data is limited, the model's performance can be limited as well.} This small clean dataset is held out for the optimization of the hyperparameters (and the poisoning points, as we describe later). Then, as proposed in \cite{foo2008efficient}, the model's hyperparameters can be learned by solving the following bilevel optimization problem: 
\begin{equation}
\begin{aligned}
\min_{\boldsymbol{\Lambda}' \in \Phi(\boldsymbol{\Lambda})}  \quad & \mathcal{L}(\mathcal{D}_\text{val},  {\bf w}^\star)  \\
\text{s.t.} \quad & {\bf w}^\star\in\argmin_{{\bf w} \in \mathcal{W}}  & \mathcal{L}\left(\mathcal{D}_\text{tr}, \boldsymbol{\Lambda}', {\bf w}\right),
\end{aligned}
\label{eqHyperparams}
\end{equation}
where $\Phi(\boldsymbol{\Lambda})$ represents the feasible domain set for the hyperparameters $\boldsymbol{\Lambda}$. The use of this approach to select the model's hyperparameters has some advantages compared to other selection methods. Cross-validation-based approaches require to re-train the model multiple times over different training and validation set splits, making it computationally very demanding when the number of hyperparameters is large and training the learning algorithm is expensive. Grid search techniques also rely on a separate validation set to select the hyperparameters. However, the exploration of all the hyperparameters values considered in the grid also requires to train the learning algorithm many times, which can be computationally demanding, especially as the number of hyperparameters in the model grows. This can be alleviated using more guided search techniques, such as Bayesian optimization, but still, the exploration of each combination of hyperparameters requires training the algorithms from scratch multiple times and the performance and scalability with the number of hyperparameters is reduced. In contrast, solving the bilevel optimization problem in Eq.~(\ref{eqHyperparams}), with gradient-based techniques, is computationally more efficient than previous approaches when using approximate techniques to estimate the hypergradients in the outer objective \cite{domke2012generic,pedregosa2016hyperparameter,franceschi2017forward,franceschi2018bilevel,munoz2017towards}. In this case, the computation of these hypergradients does not require to train the learning algorithm (in the inner objective) completely, but just for a reduced number of epochs. This approach is more scalable, especially when the number of hyperparameters is large. On the downside, gradient-based techniques to solve Eq.~(\ref{eqHyperparams}) do not guarantee to find the global optimum for the outer objective but possibly a local one. However, this problem can be mitigated with multiple re-starts. 

In a poisoning attack, the adversary aims to inject a set of $n_\text{p}$ poisoning data points, $\mathcal{D}_\text{p}= \{({\bf x}_{\text{p}_k} ,y_{\text{p}_k})\}^{n_\text{p}}_{k=1}$, in the training set to maximize some arbitrary objective, $\mathcal{A}:  \mathbb{R}^{n_\text{target} \times m} \times \mathbb{Z}_c^{n_\text{target}}  \rightarrow \mathbb{R}$, evaluated on a set of target data points $\mathcal{D}_\text{target}$. As described in \cite{munoz2017towards} different attack scenarios can be considered depending on both the set of target data points and the attacker's objective, including indiscriminate and targeted attacks. To allow for the learning of the hyperparameters we therefore propose to formulate the attacker's problem as a \emph{multiobjective} bilevel optimization problem:
\begin{equation}
\begin{aligned}
\min_{\boldsymbol{\Lambda}' \in \Phi(\boldsymbol{\Lambda})}  \mathcal{L}(\mathcal{D}_\text{val}, &   {\bf w}^\star),  \max_{\mathcal{D}_\text{p}' \in \Phi(\mathcal{D}_\text{p})}    \mathcal{A}(\mathcal{D}_\text{target},  {\bf w}^\star) \\
\text{s.t.} & \quad {\bf w}^\star\in\argmin_{{\bf w} \in \mathcal{W}}  \mathcal{L}\left(\mathcal{D}_\text{tr}',  \boldsymbol{\Lambda}', {\bf w}\right),
\end{aligned}
\label{eqAttacker}
\end{equation} 
where $\mathcal{D}_\text{tr}' = \mathcal{D}_\text{tr} \cup \mathcal{D}_\text{p}'$ is the poisoned dataset and $\Phi(\mathcal{D}_\text{p})$ is the feasible domain for the attacker.

From the general formulation in Eq.~(\ref{eqAttacker}) it is clear that the poisoning points in $\mathcal{D}_\text{tr}'$ have an effect not only on the parameters of the classifier (in the inner problem), but also on its hyperparameters (in the outer objective for the defender). Previous studies have neglected the effect of the hyperparameters in the problem for the attacker, e.g., the regularization hyperparameter for the loss function for SVMs \cite{biggio2012poisoning} or for embedded feature selection methods \cite{xiao2015feature}. This can overestimate the adversary's capabilities to influence the learning algorithm, as we show in the synthetic experiment in Fig.~\ref{fig:synthetic}.

Our novel attack formulation in Eq.~(\ref{eqAttacker}) allows to model a wide variety of attack scenarios, depending on the attacker's objective and the combinations between the target, validation and training sets. However, for the sake of clarity, in the remainder of the paper we focus on analyzing \emph{worst-case scenarios for indiscriminate poisoning attacks}, where the attacker, having perfect knowledge, aims to increase the overall classification error in the target algorithm. These settings have been commonly used in most of the related work on poisoning attacks using bilevel optimization \cite{biggio2012poisoning,mei2015using,xiao2015feature,munoz2017towards, carnerero2020regularisation}. To achieve such a goal, the attacker aims to maximize the loss evaluated on a separate validation set, i.e., $\mathcal{A}(\mathcal{D}_\text{target},  {\bf w}^{\star}) = \mathcal{L}(\mathcal{D}_\text{val},  \  {\bf w}^{\star}){:  \mathbb{R}^{n_\text{val} \times m} \times \mathbb{Z}_c^{n_\text{val}}  \rightarrow \mathbb{R}}$. In our case, where the attacker is also aware of the effect of the hyperparameters in the performance of the algorithm, $\mathcal{D}_\text{val}$ is the same as the validation dataset used by the defender, to maximize the overall error not only compromising the learning of the model's parameters, but also the selection (or learning) of its hyperparameters. Then, the attacker's problem can be formulated a bilevel optimization problem where the outer objective is a \emph{minimax} problem:
\begin{equation}
\begin{aligned}
\min_{\boldsymbol{\Lambda}' \in \Phi(\boldsymbol{\Lambda})} &  \max_{\mathcal{D}_\text{p}' \in \Phi(\mathcal{D}_\text{p})} \mathcal{L}(\mathcal{D}_\text{val}, {\bf w}^\star) \\
\text{s.t.} &  \quad {\bf w}^\star\in\argmin_{{\bf w} \in \mathcal{W}}  \mathcal{L}\left(\mathcal{D}_\text{tr}', \boldsymbol{\Lambda}', {\bf w}\right).
\end{aligned}
\label{eqAttacker2}
\end{equation}

In this formulation, in the outer problem, there is an implicit dependency of both the hyperparameters, $\boldsymbol{\Lambda}$, and the poisoning points, $\mathcal{D}_\text{p}$, on the parameters of the model learned in the inner optimization problem, ${\bf w}^\star$. We can also observe that the value of the poisoning points has an effect on the learning of both ${\bf w}$ and $\boldsymbol{\Lambda}$ in the inner and outer objectives respectively.

This formulation is compatible with grid-search-based approaches, which select the hyperparameters using a separate validation dataset. However, it is computationally infeasible to solve the problem for the attacker using these techniques, as the number of { variables} to be learned in the outer objective, {i.e.,} the model's hyperparameters and the value of the features for all the poisoning points, is very large. On the other hand, cross-validation uses the same dataset for creating the different training and validation splits. Thus, the learner can not benefit from the trusted dataset and, both the training and validation datasets would contain poisoning points across all splits. It is important to note that the availability of the small trusted dataset gives a chance to the learner to defend against poisoning attacks. In our case, the learner uses the trusted set for validation aiming to mitigate the effect of the poisoning attack by the selection of appropriate hyperparameters. Our experiments show that this can be a good approach in some cases, for example, when using regularization to increase the stability of the learning algorithm, and helps mitigate the attack. Of course, more specialized algorithms can be devised to make a different use of the trusted set of data points (e.g., data hypercleaning \cite{franceschi2017forward}). However, it is not our intention here to develop a specific algorithm for defending against data poisoning, but rather to show that the existence of a trusted dataset can be helpful to reduce the impact of poisoning attacks just by using standard techniques to increase the stability of the algorithm, such as regularization, and learning the model's hyperparameters appropriately. Our attack formulation allows us to characterize the worst-case performance under such assumptions. Thus, our findings provide ML practitioners a methodology to better use their trusted data points to mitigate poisoning attacks without requiring specialized knowledge or algorithms, but using techniques commonly used for training ML algorithms, as is the case of regularization.

\subsection{Solving General Optimal Poisoning Attacks}
\label{subsec:genpois}
Solving the multiobjective bilevel optimization problems in Eq.~(\ref{eqAttacker}) and Eq.~(\ref{eqAttacker2}) is strongly NP-Hard \cite{bard2013practical} and, even if the inner problem is convex, the bilevel problem is, in general, non-convex. However, it is possible to use gradient-based approaches to obtain (possibly) suboptimal solutions, {i.e.,} finding local optima for the problem in Eq.~(\ref{eqAttacker}) and saddle points for the minimax problem in Eq.~(\ref{eqAttacker2}). For clarity, in the rest of this paper we focus on the solution to Eq.~(\ref{eqAttacker2}), which we use in our experiments to show the robustness of $L_2$ regularization to indiscriminate poisoning attacks. The solution of Eq.~(\ref{eqAttacker}) follows a similar procedure.

Similar to \cite{biggio2012poisoning,mei2015using,xiao2015feature,munoz2017towards}, we assume that the label of the poisoning points is set a priori, so the attacker just needs to learn the features for the poisoning points, ${\bf X}_\text{p}$. For clarity, in the following description we use $\mathcal{A}$ (which does not explicitly depend on the poisoning points or the hyperparameters, but implicitly through the parameters) to denote the loss function evaluated on $\mathcal{D}_\text{val}$ in the outer objective, {i.e.,} $\mathcal{L}(\mathcal{D}_\text{val}, {\bf w}^{\star})$, and $\mathcal{L}$ to refer to the loss function evaluated on $\mathcal{D}_\text{tr}'$ in the inner objective, $\mathcal{L}(\mathcal{D}_\text{tr}', \boldsymbol{\Lambda}, {\bf w}^{\star})$. Both are evaluated on ${\bf w}^{\star}$, the parameters obtained when solving { the} inner optimization problem. 

To compute the hypergradients for the outer objective, we assume that the first and second derivatives of the loss function, $\mathcal{L}$, are Lipschitz-continuous functions. We can then compute the hypergradients applying the chain rule, so that $\nabla_{{\bf X}_{\text{p}}}\mathcal{A} = \left( d {\bf w}^\star / d {\bf X}_\text{p} \right)\tran \nabla_{\bf w} \mathcal{L}$.\footnote{\label{note1}The expression for ${\boldsymbol \Lambda}$ is analogous.} To compute the implicit derivative, $d {\bf w}^\star / d  {\bf X}_\text{p}$, we can leverage the stationarity (Karush-Kuhn-Tucker, KKT) {conditions} in the inner problem{, i.e.,} $\nabla_{{\bf w}}\mathcal{L} = {\bf 0}$, and apply the implicit function theorem \cite{mei2015using, pedregosa2016hyperparameter, koh2017understanding}, so that $\nabla_{{\bf X}_\text{p}}\nabla_{{\bf w}}\mathcal{L} + \left( d {\bf w}^\star / d {\bf X}_\text{p} \right)\tran \nabla_{\bf w} \mathcal{L} = {\bf 0}$.\footref{note1} Then, the hypergradients can be computed as
\begin{equation}
  \begin{aligned}
\nabla_{{\bf X}_\text{p}} \mathcal{A} &= -\left( \nabla_{{\bf X}_\text{p}}\nabla_{{\bf w}} \mathcal{L} \right)\tran  \left( \nabla^2_{\bf{w}} \mathcal{L} \right)^{-1} \nabla_{\bf{w}} \mathcal{A},
\label{eqHyperGrads}
  \end{aligned}
\end{equation} 
where we assume that the Hessian $\nabla^2_{\bf{w}} \mathcal{L}$ is not singular. Brute-force computation of Eq.~(\ref{eqHyperGrads}) requires inverting the Hessian, which scales in time as $\mathcal{O}(d^3)$ and in space as $\mathcal{O}(d^2)$---where $d$ is the number of parameters. However, as in \cite{foo2008efficient, pedregosa2016hyperparameter}, we can rearrange the terms in the second part of Eq.~(\ref{eqHyperGrads}), solve the linear system: $\left( \nabla^2_{{\bf w}} \mathcal{L}\right){\bf v} = \nabla_{{\bf w}}\mathcal{A}$, and compute $\nabla_{{\bf X}_\text{p}} \mathcal{A}=-\left(\nabla_{{\bf X}_\text{p}} \nabla_{{\bf w}}\mathcal{L}\right)\tran {\bf v}$.\footref{note1} The linear system can be efficiently solved by using Conjugate Gradient (CG) descent, as described in \cite{foo2008efficient}. For this, let us assume that the inner problem is solved by an iterative algorithm that arrives at a local minima after $T_\text{KKT}$ training iterations. After solving the linear system, the procedure  scales in time $\mathcal{O}\left(\left(T_\text{KKT} + \sqrt{\kappa}\right)d\right)$ and in space $\mathcal{O}(d)$ \cite{shewchuk1994introduction}, where $\kappa$ is the condition number of the Hessian $\nabla^2_{\bf{w}} \mathcal{L}$. Moreover, the Hessian-vector products $\left( \nabla^2_{{\bf w}} \mathcal{L}\right){\bf v}$ and $\left(\nabla_{{\bf X}_\text{p}} \nabla_{{\bf w}}\mathcal{L}\right)\tran {\bf v}$ can be computed exactly and efficiently with the technique proposed in \cite{pearlmutter1994fast}, { thus} avoiding the computation and storage of the Hessian, as follows:\footref{note1}
\begin{gather}
  \begin{aligned}
\begin{split}
\left(\nabla_{\bf w}^2 \mathcal{L} \right){\bf v} &= \nabla_{\bf w}\left({\bf v}\tran\nabla_{\bf w} \mathcal{L} \right),\\
\left(\nabla_{{\bf X}_\text{p}}\nabla_{\bf w} \mathcal{L} \right)\tran{\bf v} &= \nabla_{{\bf X}_\text{p}}\left({\bf v}\tran\nabla_{\bf w} \mathcal{L} \right).
\end{split}
  \end{aligned}
\end{gather} 

The computation of the first and second expression above scales as $\mathcal{O}(d)$ and $\mathcal{O}(\max(d, n_\text{p} m))$\footnote{$\mathcal{O}(\max(d, { h}))$ for  $\boldsymbol{\Lambda}$, where ${ h}$ is the number of hyperparameters.} respectively---where $n_\text{p}$ denotes the number of poisoning points, each one containing $m$ features---both in time and in space. An elegant aspect of this technique is that, for ML models optimized with gradient-based methods, the equations for evaluating
the Hessian-vector products emulate closely those for standard forward and backward propagation. Hence, the application of existing automatic differentiation frameworks to compute this product is typically straightforward \cite{pearlmutter1994fast, bishop2006pattern}.

However, approaches based on the implicit function theorem require training the whole learning algorithm to compute the hypergradient, i.e., until the stationarity conditions are met. This can be intractable for some learning algorithms such as deep networks, where the number of parameters is huge. To sidestep this problem, different techniques have been proposed to estimate the value of the hypergradients \cite{domke2012generic,maclaurin2015gradient,pedregosa2016hyperparameter,franceschi2017forward,munoz2017towards,franceschi2018bilevel, grazzi2020iteration}. These techniques do not require to re-train the learning algorithm each time the hypergradient is computed. Instead, they estimate the hypergradient by truncating the learning in the inner problem to a reduced number of training iterations.

As described in \cite{franceschi2017forward}, we can think of the training algorithm (inner problem) as a discrete-time dynamical system, described by a sequence of states ${\bf s}^{(t)}\left({\bf X}_{\text{p}}, {\boldsymbol \Lambda}\right) \in\mathbb{R}^{d_\text{s}}$, with $t = 1, \ldots, T$, where each state depends on model's parameters, the accumulated gradients and/or the velocities, and the training data and hyperparameters. In this paper, we focus on Stochastic Gradient Descent (SGD), {i.e.,} ${\bf s}^{(t)}\left({\bf X}_{\text{p}}, {\boldsymbol \Lambda}\right) = {\bf w}^{(t)}\left({\bf X}_{\text{p}}, {\boldsymbol \Lambda}\right)$, so that each state of the sequence depends \emph{only} on the previous state. We can therefore reformulate the bilevel problem in (\ref{eqAttacker2}) as the constrained single-level optimization problem:
\begin{equation}
\begin{aligned}
\min_{\boldsymbol{\Lambda}' \in \Phi(\boldsymbol{\Lambda})} &  \max_{{\bf X}_\text{p}' \in \Phi(\mathcal{D}_\text{p})} \mathcal{L}\left(\mathcal{D}_\text{val}, {\bf w}^{(T)} \left({\bf X}_\text{p}, {\boldsymbol \Lambda}\right)\right) \\
\text{s.t.} &  \quad  {\bf w}^{(t)}\left({\bf X}_\text{p}, {\boldsymbol \Lambda}'\right) = {\bf w}^{(t - 1)}\left({\bf X}_\text{p}, {\boldsymbol \Lambda}\right) \\
& \qquad \qquad \qquad \quad \ \, - \eta \nabla_{{\bf w}} \mathcal{L}\left(\mathcal{D}_\text{tr}', \boldsymbol{\Lambda}', {\bf w}^{(t - 1)}\right),\\
& \quad t = 1, \dots, T,
\end{aligned}
\label{eqAttacker3}
\end{equation} 
where $\eta$ is the learning rate for SGD.

Then, we estimate the hypergradients from the values of the parameters collected in the set of training states as\footref{note1}
\begin{equation}
\begin{split}
\nabla_{{\bf X}_\text{p}} \mathcal{A} &=  \left(\frac{{d} {\bf w}^{(T)}\left({\bf X}_\text{p}, {\boldsymbol \Lambda}\right)}{{d} {\bf X}_\text{p}}  \right)\tran \nabla_{\bf{w}} \mathcal{A}, \\
\end{split}
\end{equation}
where the bottleneck is, again, the computation of the implicit derivatives. Given the constraints in Eq.~(\ref{eqAttacker3}), it is obvious that the state ${\bf w}^{(t)}\left({\bf X}_\text{p}, {\boldsymbol \Lambda}\right)$ depends on the poisoning points and hyperparameters both, directly by its expression, and indirectly through the previous state ${\bf w}^{(t-1)}\left({\bf X}_\text{p}, {\boldsymbol \Lambda}\right)$. Then, by applying the chain rule we obtain\footref{note1}
\begin{equation}
\frac{d {\bf w}^{(t)} \left({\bf X}_\text{p}, {\boldsymbol \Lambda}\right)}{d {\bf X}_\text{p}} = \frac{\partial {\bf w}^{(t)} }{\partial {\bf X}_\text{p}} + \frac{\partial {\bf w}^{\left(t\right)}}{\partial {\bf w}^{\left(t-1\right)}}  \frac{d {\bf w}^{(t-1)} \left({\bf X}_\text{p}, {\boldsymbol \Lambda}\right)}{d {\bf X}_\text{p}}
\end{equation} Then, from a reduced number of training iterations, $T \leq T_\text{KKT}$ (which does not necessarily satisfy the stationarity conditions \cite{franceschi2017forward, munoz2017towards, franceschi2018bilevel, grazzi2020iteration}),  these expressions can be expanded, according to the updates of SGD \cite{domke2012generic, franceschi2017forward}, as follows:
\begin{equation}
  \begin{aligned}
\nabla_{{\bf X}_\text{p}}\mathcal{A}  &= \left( \frac{\partial {\bf w}^{(T)}}{\partial {\bf X}_\text{p}} + \sum_{t=1}^{T-1} \left(\prod_{t'=t+1}^T \frac{\partial {\bf w}^{\left(t'\right)}}{\partial {\bf w}^{\left(t'-1\right)}} \right) \frac{\partial {\bf w}^{(t)}}{\partial {\bf X}_\text{p}} \right)\nabla_{\bf{w}} \mathcal{A},\\
 \nabla_{{\boldsymbol \Lambda}}\mathcal{A}  &=   \left( \frac{\partial {\bf w}^{(T)}}{\partial {\boldsymbol \Lambda}} + \sum_{t=1}^{T - 1} \left(\prod_{t'=t+1}^T \frac{\partial {\bf w}^{\left(t'\right)}}{\partial {\bf w}^{\left(t'-1\right)}} \right) \frac{\partial {\bf w}^{(t)}}{\partial {\boldsymbol \Lambda}}  \right)\nabla_{\bf{w}} \mathcal{A},
\label{eqHyperGrads3}
  \end{aligned}
\end{equation} 
where $ \partial {\bf w}^{\left(t'\right)} / \partial {\bf w}^{\left(t'-1\right)}  = {\bf I} - \eta \nabla_{\bf w}^2 \mathcal{L}$, $ \partial {\bf w}^{(t)} / \partial {\bf X}_\text{p}  = -\eta \nabla_{{\bf X}_\text{p}} \nabla_{\bf w} \mathcal{L}$, and $ \partial {\bf w}^{(t)} / \partial {\boldsymbol \Lambda}  = -\eta \nabla_{{\boldsymbol \Lambda}} \nabla_{\bf w} \mathcal{L}$.

Depending on the order to compute the different terms in Eq.~(\ref{eqHyperGrads3}), we can use two approaches to estimate the hypergradients: Reverse-Mode (RMD) and Forward-Mode Differentiation (FMD) \cite{griewank2008evaluating,franceschi2017forward}. In the first case, RMD requires first to train the learning algorithm for $T$ training iterations, {i.e.,} to compute ${\bf w}^{(1)}$ to ${\bf w}^{(T)}$. Then, the hypergradients estimate is computed by reversing the steps followed by the learning algorithm from ${\bf w}^{(T)}$ down to ${\bf w}^{(1)}$. On the other hand, FMD computes the estimate of the hypergradients as the algorithm is trained, {i.e.,} from ${\bf w}^{(1)}$ to ${\bf w}^{(T)}$ (i.e. the estimates can be computed in parallel with the training procedure).

 To estimate the hypergradients, RMD requires to compute a forward and a backward pass through the set of states. In some cases, as in \cite{domke2012generic, franceschi2017forward, franceschi2018bilevel}, RMD requires to store all the information collected in the states in the forward pass.\footnote{However, other RMD methods proposed in the literature do not require to store this information \cite{maclaurin2015gradient,munoz2017towards}.}  In contrast, FMD just needs to do the forward computation. However, compared to RMD, the scalability of FMD depends heavily on the number of hyperparameters. As a practical example, consider training a neural network (including LR as a  special case) with $d$ weights, using classic iterative optimization algorithms such as SGD. According to Eq.~(\ref{eqHyperGrads3}), RMD scales in time as $\mathcal{O}(Td)$ and in space as $\mathcal{O}(n_\text{p} m  + { h} + Td)$, while FMD scales as $\mathcal{O}((n_\text{p} m  + { h})Td)$ and $\mathcal{O}((n_\text{p} m  + { h})d)$ in time and space respectively. Thus, the time complexity of RMD does not depend on the size of the poisoning points or hyperparameters. Then, for problems where the number of hyperparameters is large, as is the case for the poisoning attacks we introduced in the paper, RMD is computationally more efficient to estimate the hypergradients. As mentioned before, it is also clear that RMD is more efficient compared to grid search, where the learning algorithms need to be trained from scratch for each combination of the hyperparameters' values explored in the grid. 

\begin{table}[t]

\centering
	\caption{Complexity of state-of-the-art methods to compute the hypergradient for the poisoning points and hyperparameters, assuming that SGD is the solver for the inner problem.}{ 

\begin{tabular}{|l|c|c|c|}
\hline
Method & Time Complexity                                                      & Space Complexity          &  \begin{tabular}[c]{@{}c@{}}Requires \\ Stationarity\end{tabular} \\ \hline
Exact     & $\mathcal{O}\left( d^3 \right)$ & $\mathcal{O}\left(d^2\right)$          & Yes                   \\
CG     & $\mathcal{O}\left(\left(T_\text{KKT} + \sqrt{\kappa}\right)d\right)$ & $\mathcal{O}(n_\text{p} m  + { h} + d)$          & Yes                   \\
RMD    & $\mathcal{O}(Td)$                                                    & $\mathcal{O}(n_\text{p} m  + { h} + Td)$ & No                    \\
FMD    & $\mathcal{O}((n_\text{p} m  + { h})Td)$                                             & $\mathcal{O}((n_\text{p} m  + { h})d)$   & No                    \\ \hline
\end{tabular}}
\label{table:complex}
\end{table}

Table~\ref{table:complex} summarizes the computational trade-offs between different state-of-the-art methods to compute the hypergradients. For the analysis of the convergence properties of the hypergradients, we refer the reader to \cite{grazzi2020iteration}, which studies and compares the convergence rate of techniques such as CG and RMD. From a practical perspective, the number of training iterations for the inner problem plays a crucial role in the convergence rate \cite{pedregosa2016hyperparameter, franceschi2018bilevel, grazzi2020iteration}, but can also cause overfitting in the outer objective \cite{franceschi2018bilevel}.

\begin{algorithm}[!t]
	\caption{{Reverse-Mode Differentiation}}
	\label{alg:bg}

	\begin{flushleft}

    {\bfseries Input:}  $\mathcal{M}$, $\mathcal{A}$, $\mathcal{L}$,  $\mathcal{D}_\text{val}$, $\mathcal{D}_\text{tr}'$, $\boldsymbol{\Lambda}$, ${\bf w}^{(0)}$,  $T$, $\eta$ \\
		{\bfseries Output:} { 	$ 	\nabla_{{\bf X}_\text{p}}\mathcal{A}\left({ \mathcal{D}_\text{val},} {\bf w}^{(T)}\right)$, $\nabla_{\boldsymbol{\Lambda}}\mathcal{A}\left({ \mathcal{D}_\text{val},} {\bf w}^{(T)}\right)$}

	\end{flushleft}
	
	\begin{algorithmic}[1]

		\FOR{$t=0$ \textbf{to} $T-1$}  \label{lin:forrev}

		\STATE  { ${\bf w}^{(t+1)}\leftarrow {\bf w}^{(t)} - \eta \nabla_{{\bf w}_{}}\mathcal{L}\left(	{ \mathcal{D}_\text{tr}', \boldsymbol{\Lambda},} {\bf w}^{(t)}\right)$} 
		\ENDFOR \label{lin:endforrev}
		
				\STATE ${ { \partial}{\bf w}^{(T)} \leftarrow \nabla_{{\bf w}_{}}\mathcal{A}\left({ \mathcal{D}_\text{val},} {\bf w}^{(T)}\right)}$ 
		
		\STATE 	${ \partial}{\bf X}_\text{p}^{(T)} \leftarrow {\bf 0}$
		
		\STATE ${ \partial}\boldsymbol{\Lambda}^{(T)} \leftarrow {\bf 0}$

		\FOR{$t=T-1$ \textbf{down to} $0$}   \label{lin:forrev2}

		\STATE 	${ \partial}{\bf g}_{\text{X}_\text{p}} \gets \left(\nabla_{{\bf X}_\text{p}}\nabla_{{\bf w}_{}}\mathcal{L}\left(	{ \mathcal{D}_\text{tr}', \boldsymbol{\Lambda},} {\bf w}^{(t)}\right)\right)\tran { \partial}{\bf w}^{(t+1)} $ \label{lin:hvp1} 
		
				\STATE ${ \partial}{\bf g}_{{\Lambda}} \gets \left(\nabla_{\boldsymbol{\Lambda}}\nabla_{{\bf w}_{}}\mathcal{L}\left(	{ \mathcal{D}_\text{tr}', \boldsymbol{\Lambda},} {\bf w}^{(t)}\right)\right)\tran { \partial}{\bf w}^{(t+1)} $ \label{lin:hvp2} 
		
				\STATE	${ \partial}{\bf w}^{(t)}  \leftarrow { \left({\bf I} - \eta \nabla^2_{{\bf w}_{}}\mathcal{L}\left(	{ \mathcal{D}_\text{tr}', \boldsymbol{\Lambda},} {\bf w}^{(t)}\right)\right){ \partial}{\bf w}^{(t+1)} }
		$ \label{lin:hvp3} 
		
		\STATE $	{ \partial}{\bf X}_\text{p}^{(t)}  \leftarrow { \partial}{\bf X}_\text{p}^{(t+1)}  - \eta { \partial}{\bf g}_{\text{X}_\text{p}}
		$ \label{lin:uphg1}
		
				\STATE	$
		{ \partial}\boldsymbol{\Lambda}^{(t)}  \leftarrow { \partial}\boldsymbol{\Lambda}^{(t+1)}  - \eta { \partial}{\bf g}_{{\Lambda}}
		$ \label{lin:uphg2}

		\ENDFOR  \label{lin:endforrev2}

				\STATE 
		$ 	\nabla_{{\bf X}_\text{p}}\mathcal{A}\left({ \mathcal{D}_\text{val},} {\bf w}^{(T)}\right)
		\leftarrow { \partial}{\bf X}_\text{p}^{(0)}$ \label{lin:finhg1}
				\STATE $\nabla_{\boldsymbol{\Lambda}}\mathcal{A}\left({ \mathcal{D}_\text{val},} {\bf w}^{(T)}\right)
		\leftarrow { \partial}{\boldsymbol{\Lambda}}^{(0)}$ \label{lin:finhg2}

	\end{algorithmic}
	
\end{algorithm}

Here we include the RMD algorithm (Alg.~\ref{alg:bg}), which we use to compute the {hypergradients estimate} at the outer level problem (both for the features of the poisoning points (Line \ref{lin:finhg1}), and the hyperparameters (Line \ref{lin:finhg2})). { RMD requires first to train the learning algorithm for $T$ training iterations (Lines \ref{lin:forrev}-\ref{lin:endforrev}). Then, the hypergradients estimate is computed by differentiating the updates of the learning algorithm and reversing its sequence of parameters (Lines \ref{lin:forrev2}-\ref{lin:endforrev2}), i.e., expanding the terms in Eq.~(\ref{eqHyperGrads3}) in reverse order. This approach can be derived by leveraging a Lagrangian formulation associated with
the parameter optimization dynamics \cite{franceschi2017forward}. Lines \ref{lin:hvp1}-\ref{lin:hvp3} compute the corresponding Hessian-vector products, whereas Lines \ref{lin:uphg1}-\ref{lin:uphg2} update the value of the hypergradients.} We use a notation similar to \cite{domke2012generic, maclaurin2015gradient, munoz2017towards}{, where more details on the derivation of this algorithm can be found.

\subsection{Projected Hypergradient Descent/Ascent}

After computing the hypergradients, at each \emph{hyperiteration} we use projected hypergradient descent/ascent to update the poisoning points and the hyperparameters: 
\begin{equation}
  \begin{aligned}
{\bf X}_\text{p} & \leftarrow \Pi_{\Phi(\mathcal{D}_\text{p})} \left( {\bf X}_\text{p} + \alpha \ \nabla_{{\bf X}_\text{p}} \mathcal{A} \right),\\
{\boldsymbol \Lambda} & \leftarrow \Pi_{\Phi({\boldsymbol \Lambda})} \left( {\boldsymbol \Lambda} - \alpha \ \nabla_{{\boldsymbol \Lambda}} \mathcal{A} \right),
  \end{aligned}
\label{eqUpdates}
\end{equation} where $\alpha$ is the learning rate for the outer problem and $\Pi_{\Phi(\mathcal{D}_\text{p})}$ and $\Pi_{\Phi({\boldsymbol \Lambda})}$ are the projection operators for the features of the poisoning points, ${\bf X}_\text{p}$, and the hyperparameters, ${\boldsymbol \Lambda}${,  defined as $\Pi_{\Phi{(\cdot)}} (input) 	\triangleq \text{clip}(input, \inf \Phi(\cdot), \sup \Phi(\cdot))$}, so that their updated values are within the corresponding feasible domains, $\Phi(\cdot)$. In our case we used standard gradient descent/ascent to solve Eq.~(\ref{eqAttacker2}). The analysis of other alternatives to solve minimax games, such as \cite{daskalakis2018limit}, is left for future work. 

\begin{algorithm*}[t]
	\caption{Projected Hypergradient Descent/Ascent}
	\label{alg:adreg}
	
    \begin{flushleft}
	{\bfseries Input:} $\mathcal{M}$, $\mathcal{A}$, $\mathcal{L}$, $\mathcal{D}_\text{val}$, $\mathcal{D}_\text{tr}$,  $n_\text{p}$, $\mathcal{P}$, $T_\text{mul}$, $T$, $\alpha$, $\eta$ \\
		{\bfseries Output:} $\mathcal{D}_\text{p}^{(T_\text{mul})}$, $\boldsymbol{\Lambda}^{(T_\text{mul})}$
	\end{flushleft}
	
	\begin{algorithmic}[1]
		\STATE $\mathcal{D}_\text{p}^{(0)} \leftarrow \texttt{initDp}\left(\mathcal{D}_\text{tr}, n_\text{p}\right)$ \COMMENT{Generate $n_\text{p}$ initial samples for $\mathcal{D}_\text{p}^{(0)}$ }
		
		\label{lin:initdp}

		\STATE $\mathcal{D}_\text{tr}'^{(0)}  \leftarrow \left( \mathcal{D}_\text{tr} \setminus  \{({\bf x}_{\text{tr}_k} ,y_{\text{tr}_k})\}_{k\in \mathcal{P}} \right) \cup \mathcal{D}_\text{p}^{(0)}$ \COMMENT{Replace $n_\text{p}$ samples of $\mathcal{D}_\text{\text{tr}}$ by  $\mathcal{D}_\text{p}^{(0)}$ }
		
				\label{lin:upddtr1}

		\STATE $\boldsymbol{\Lambda}^{(0)}\leftarrow \texttt{initL}(\mathcal{M})$ \COMMENT{Initialize $\boldsymbol{\Lambda}$}
		
		\label{lin:initl}

		\FOR{$\tau=0$ \textbf{to} $T_\text{mul}-1$}  	\label{lin:formul}
		
		\STATE 	${\bf w}^{(0)} \leftarrow \texttt{initW}(\mathcal{M})$ \COMMENT{Initialize {\bf w}}
		\label{lin:initw1} 
		\STATE $  \nabla_{{\bf X}_\text{p}}\mathcal{A}\left({ \mathcal{D}_\text{val},} {\bf w}^{(T)}\right), \nabla_{\boldsymbol{\Lambda}}\mathcal{A}\left({ \mathcal{D}_\text{val},} {\bf w}^{(T)}\right)
		\leftarrow \texttt{hypGrad}\left(\mathcal{M},\mathcal{A}, {\mathcal{L}, \mathcal{D}_\text{val}, \mathcal{D}_\text{tr}'^{(\tau)},  \boldsymbol{\Lambda}^{(\tau)}}, {\bf w}^{(0)}, T, \eta\right)$
		\label{lin:hypgrad} \COMMENT{ RMD (Alg.~\ref{alg:bg})}	
		\STATE ${\bf X}_\text{p}^{(\tau+1)}\leftarrow \Pi_{\Phi \left(\mathcal{D}_\text{p}\right)} \left({\bf X}_\text{p}^{(\tau)} + \alpha\nabla_{{\bf X}_\text{p}}\mathcal{A}\left({ \mathcal{D}_\text{val},} {\bf w}^{(T)}\right)\right)$ \COMMENT{Projected Hypergradient Ascent}
		\label{lin:phga}

		\STATE ${\bf X}_\text{tr}'^{(\tau+1)}  \leftarrow \left( {\bf X}_\text{tr}'^{(\tau)} \setminus {\bf X}_\text{p}^{(\tau)} \right) \cup {\bf X}_\text{p}^{(\tau+1)}$ \COMMENT{Update ${\bf X}_\text{\text{tr}}'^{(\tau)}$ with ${\bf X}_\text{p}^{(\tau+1)}$ }
		
		\label{lin:upddtr2}

		\STATE $\boldsymbol{\Lambda}^{(\tau+1)}\leftarrow \Pi_{\Phi\left(\boldsymbol{\Lambda}\right)}\left(\boldsymbol{\Lambda}^{(\tau)} - \alpha\nabla_{\boldsymbol{\Lambda}}\mathcal{A}\left({ \mathcal{D}_\text{val},} {\bf w}^{(T)}\right)\right)$ \COMMENT{Projected Hypergradient Descent}
				\label{lin:phgd}
		
		\ENDFOR \label{lin:endformul}
		
	\end{algorithmic}
	
\end{algorithm*}

Alg.~\ref{alg:adreg} describes the procedure to solve the multiobjective bilevel problem proposed in the paper. Essentially, this algorithm implements projected hypergradient descent/ascent for $T_\text{mul}$ iterations (Lines \ref{lin:formul}-\ref{lin:endformul}) to optimize, in a coordinated manner, the poisoning points (Line \ref{lin:phga})---replaced into the training set (Line \ref{lin:upddtr2})---and the set of hyperparameters (Line \ref{lin:phgd}).

To reduce the computational burden, we consider the simultaneous optimization of a batch of $n_\text{p}$ poisoning points, $\mathcal{D}_\text{p} = \{({\bf x}_{\text{p}_k} ,y_{\text{p}_k})\}^{n_\text{p}}_{k=1}$. We generate the initial values of $\mathcal{D}_\text{p}$ by cloning $n_\text{p}$ samples---uniformly sampled without duplicates---of $\mathcal{D}_\text{tr}$. Their labels are initially flipped and
kept fixed during the optimization. This process is carried out in the function \texttt{initDp} (Line \ref{lin:initdp}).
Then, these $n_\text{p}$ poisoning samples replace the $n_\text{p}$ clean samples of $\mathcal{D}_\text{tr}$ whose indices are in the set $\mathcal{P}$ (Line \ref{lin:upddtr1}).
On the other hand, the hyperparameters are initialized in \texttt{initL} (Line \ref{lin:initl}).

To solve the bilevel problem, every time the variables in the outer problem are updated, the model's parameters need to be previously initialized and optimized. Thus, let \texttt{initW} (Line \ref{lin:initw1}) 
be a particular initialization for the model's parameters. $\texttt{hypGrad}$ (Line \ref{lin:hypgrad}) refers to the particular optimization algorithm used to train the model's parameters and compute the corresponding hypergradients. In this work, this algorithm is Reverse-Mode Differentiation (RMD) (Alg.~\ref{alg:bg}).

\section{Regularization to Partially Mitigate Poisoning Attacks} \label{sec:L2}
Poisoning attacks are intrinsically related to the stability of ML algorithms. Attackers aim to produce large changes in the target algorithm by influencing a reduced set of training points. Xu \MakeLowercase{\textit{et al.}} \cite{xu2011sparse} introduced the following definition of stability: \emph{``an ML algorithm is stable if its output is nearly identical on two datasets, differing on only one sample."} This concept of stability has also been studied in the field of robust statistics, in which ``robustness'' formally denotes this definition of stability \cite{rubinstein2009antidote}. It is not our intention here to provide a formal analysis of the stability of ML algorithms, but to show that stability is an important property in the design of ML algorithms { that are} robust to data poisoning.

$L_2$ (or Tikhonov) regularization is a well-known mechanism to increase the stability of ML algorithms \cite{bousquet2002stability, xu2011sparse}. {In $L_2$ regularization,} a penalty term is added to the original loss function, which shrinks the norm of the model's parameters, so that $\mathcal{L}( \mathcal{D}_\text{tr},{\bf w}, \lambda)=\mathcal{L}(\mathcal{D}_\text{tr}, {\bf w})+\frac{e^{\lambda}}{2}\left|\left|{\bf w}\right|\right|_2^2$,
where $\lambda$ is the hyperparameter that controls the strength of the regularization term. The exponential form is used to ensure a positive contribution of the regularization term to the loss function and to help learning $\lambda$, for example by using Eq.~(\ref{eqHyperparams}), as this hyperparameter is usually searched over a log-spaced grid \cite{pedregosa2016hyperparameter}. In principle, different $L_2$ regularization schemes can be considered: e.g., in neural networks, we could have { a different regularization term for each layer or even for each parameter \cite{foo2008efficient}}.

Xiao \MakeLowercase{\textit{et al.}} \cite{xiao2015feature} analyzed the robustness of embedded feature selection, including $L_2$ and $L_1$ regularization, for linear classifiers against optimal poisoning attacks. Although their experimental results showed that $L_2$ was slightly more robust compared to $L_1$ regularization and \emph{elastic-net}, all the classifiers tested where very vulnerable to indiscriminate optimal poisoning attacks. However, these results relied on the assumption that the regularization hyperparameter was constant regardless of the fraction of poisoning data, which as we show in our experiments provides a limited perspective on the robustness of the learning algorithms.

\begin{figure*}[!t]
\centering
{\includegraphics[width=1.8in]{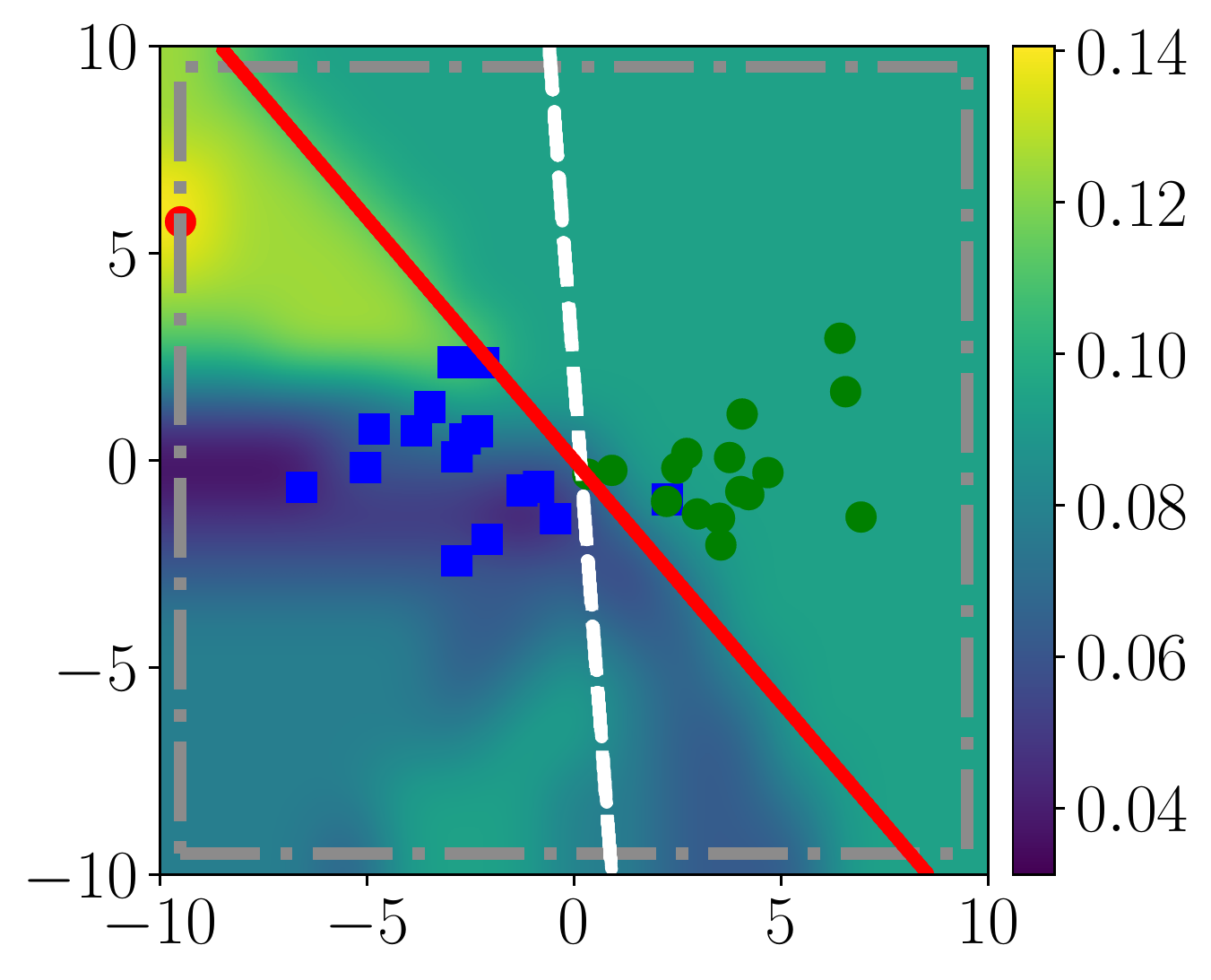}%
\label{fig:synthetic_a}}
{\includegraphics[width=1.8in]{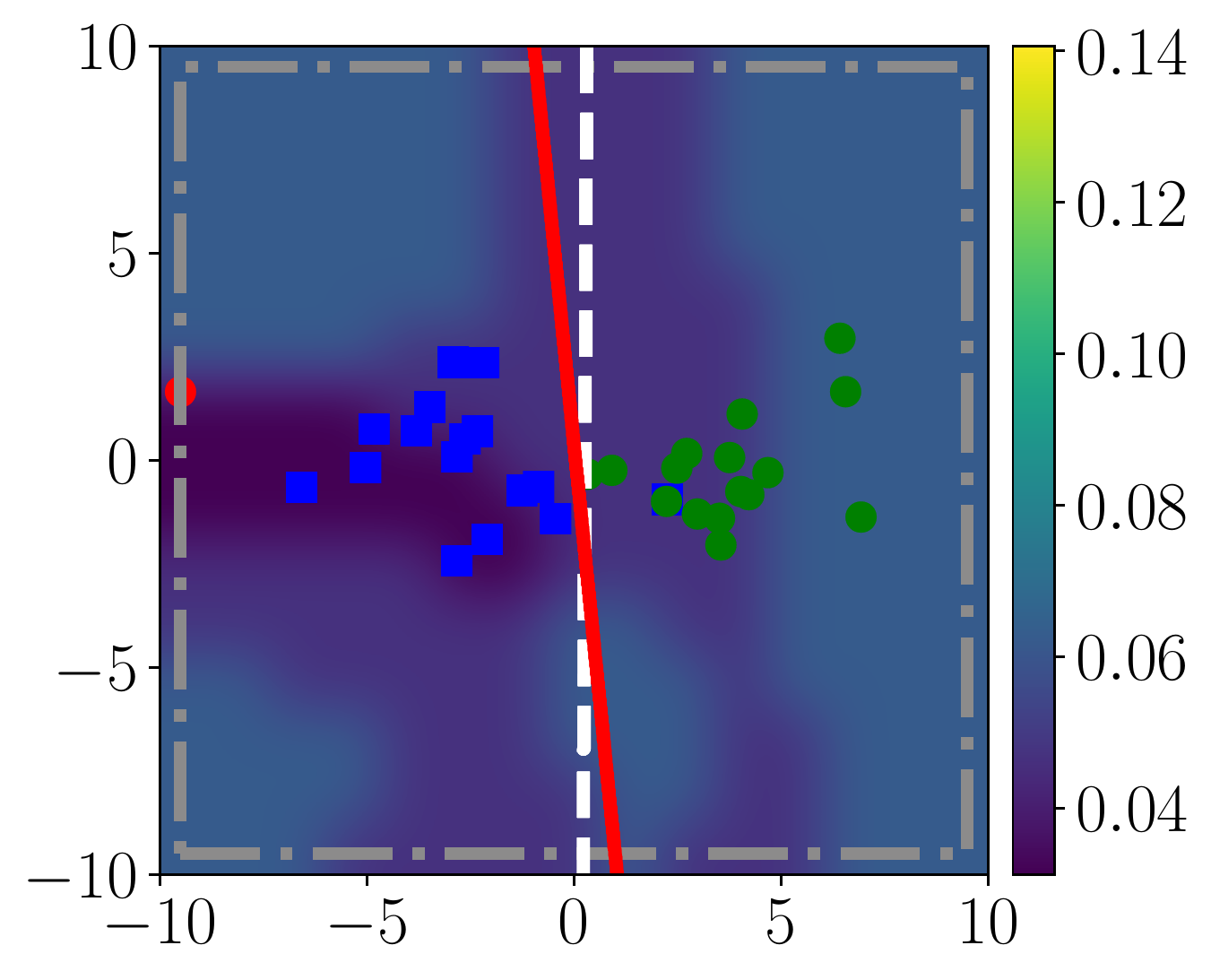}%
\label{fig:synthetic_b}}
{\includegraphics[width=1.8in]{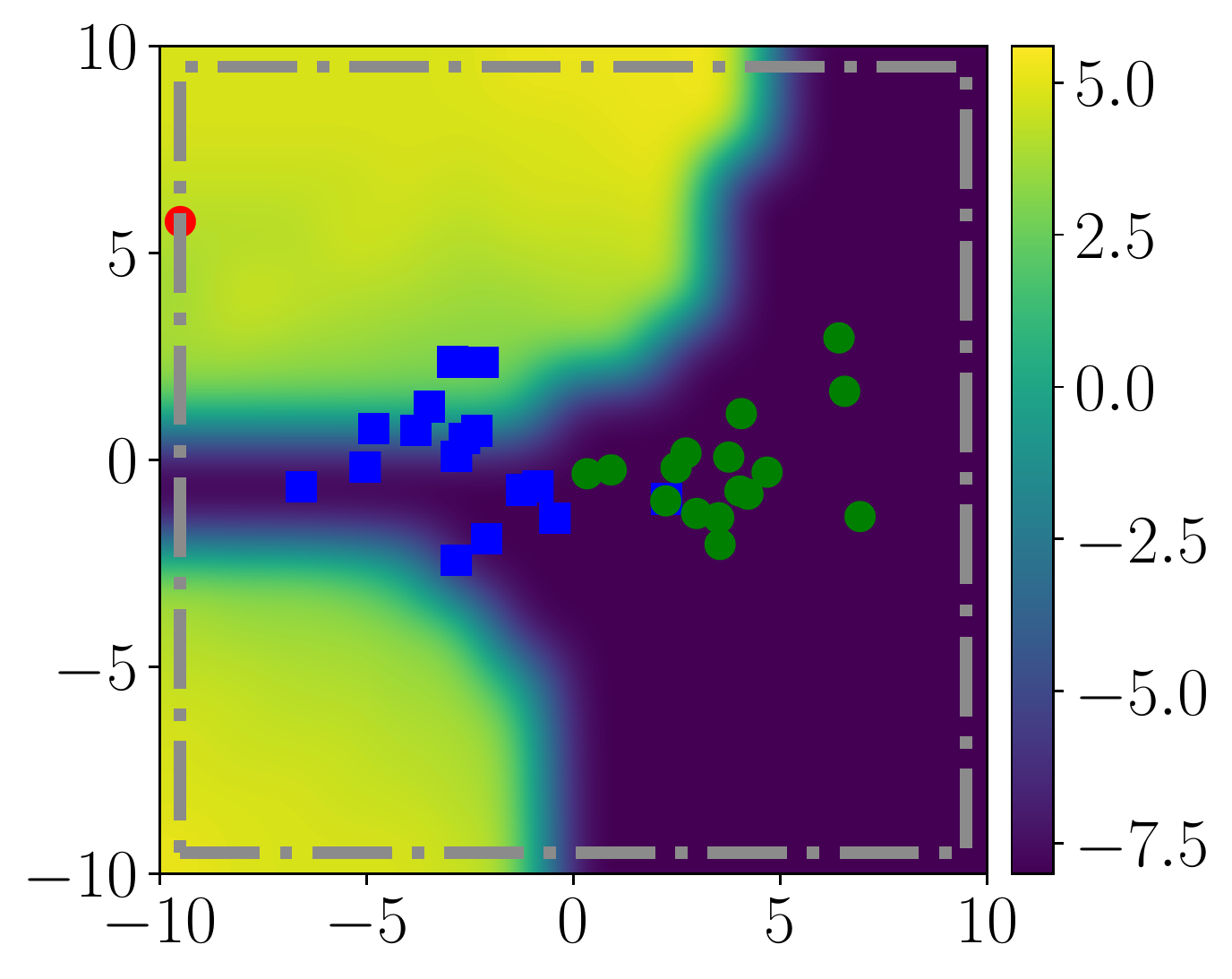}%
\label{fig:synthetic_c}}
\caption{Effect of regularization on a synthetic example. The blue and green points represent the training data points for each class, and the red point is the poisoning point (labeled as green). The dashed-dotted grey box represent the attacker's constraints. Dashed-white lines and solid-red lines depict the decision boundaries for LR classifiers trained on clean and poisoned datasets respectively. (Left) Standard LR with no regularization. (Center) LR with $L_2$ regularization. The colormaps in the two plots represent the validation error as a function of the poisoning point. (Right) Value of $\lambda$ learned by solving Eq.~(\ref{eqHyperparams}) as a function of the injected poisoning point.}
\label{fig:synthetic}
\end{figure*}

The synthetic example with a binary classifier in Fig.~\ref{fig:synthetic} illustrates the limitations of the approach in \cite{xiao2015feature}. Here, $16$ points per class were drawn from two different bivariate Gaussian distributions and we trained an LR classifier. Fig.~\ref{fig:synthetic}(left) shows the effect of injecting a single poisoning point (red point, labeled as green) to maximize the error (measured on a separate validation set with $32$ points per class) against a non-regularized LR classifier.\footnote{ The details of the experiment can be found in {Appx. \ref{sec:expset}.}} The dashed-white line represents the decision boundary learned when training on the clean dataset, and the red line depicts the decision boundary when training on the poisoned dataset. We observe that a single poisoning point can significantly alter the decision boundary. Fig.~\ref{fig:synthetic}(center), shows a similar scenario, but training an LR classifier with $L_2$ regularization, setting $\lambda=\log(20)\approx3$. Here, we observe that the effect of the poisoning point is much reduced and the decision boundary shifts only slightly. In the background of these two figures we represent the validation error of the LR trained on a poisoned dataset as a function of the location of the poisoning point. We observe that, when there is no regularization (left) the error can significantly increase when we inject the poisoning point in certain regions. On the contrary, when regularization is applied (center), the colormap is more uniform, i.e., the algorithm is quite stable regardless of the position of the poisoning point. Note that, when the model is regularized, the increase in the validation error after the attack is small.  In the next section, we also experiment with $L_1$ regularization against data poisoning. In this case, $\mathcal{L}( \mathcal{D}_\text{tr},{\bf w}, \lambda)=\mathcal{L}(\mathcal{D}_\text{tr}, {\bf w})+e^{\lambda}\left|\left|{\bf w}\right|\right|_1${.} 

Fig.~\ref{fig:synthetic}(right) shows how the optimal value of $\lambda$ that minimizes the loss in the trusted validation set changes significantly as a function of the location of the poisoning point. The colormap in the background represents the value of $\lambda$. 
We observe that $\lambda$ is much bigger for the regions where the poisoning point can influence the classifier more  (Fig.~\ref{fig:synthetic}(left)). So, when the poisoning attack has a negative impact on the classifier's performance, the importance of the regularization term, controlled by $\lambda$, increases. It is clear that selecting the value of $\lambda$ appropriately, using a small trusted validation set, can have a significant impact on the classifier's robustness. Furthermore, when testing the robustness of regularized classifiers we must consider the interplay between the attack strength and the value of  $\lambda$.

\section{Experiments}
\label{sec:experiment}

\begin{table*}[t]

	\centering
	
	\caption{Characteristics of the datasets used in the experiments.}{ 
		\begin{tabular}{|l|c|c|c|c|}
			\hline
			Dataset & \makecell{\# Training Samples} & \makecell{\# Validation Samples} & \makecell{\# Test Samples} & \# Features \\
			\hline
			MNIST (`0' vs. `8') & $5,000$ & $500$ & $3,000$ & $784$ \\
			FMNIST (\emph{trouser} vs. \emph{pullover}) & $5,000$ & $500$ & $3,000$ & $784$ \\
			CIFAR-10 (\emph{airplane} vs. \emph{frog}) & $5,000$ & $1,000$ & $2,500$ & $3,072$ \\
			\hline
	\end{tabular}}
	\label{tabDatasets}

\end{table*}

\begin{samepage}

\begin{table*}[!t]

	\centering
	
	\caption{Experimental settings for the poisoning attack.}{ 
	
			\begin{tabular}{|l|c|c|c|c|c|c|}
			\hline
			Dataset (Model) &  $T_\text{mul}$ &  $\alpha$ & $\Phi(\mathcal{D}_\text{p})$ & $\Phi(\lambda)$ & $\eta$ & $T$  \\
			\hline
			MNIST (`0' vs. `8') (LR) & $140$ & $0.300$ & $[0.0, 1.0]^{784}$ & $\left(-\infty,\log\left(5\cdot 10^3\right)\right]$ & $0.10$ & $140$ \\
			FMNIST (\emph{trouser} vs. \emph{pullover}) (LR) & $150$ & $0.300$ & $[0.0, 1.0]^{784}$ & $\left(-\infty, \log\left(5\cdot 10^3\right)\right]$ & $0.04$ & $160$\\
			CIFAR-10 (\emph{airplane} vs. \emph{frog}) (LR) & $120$  & $0.300$ & $[-1, 1]^{3,072}$ & $\left(-\infty,\log\left(10^5\right)\right]$ & $0.01$ & $500$ \\
			MNIST (`0' vs. `8') (DNN) & $180$ & $0.075$ & $[0.0, 1.0]^{784}$ & $\left(-\infty,\log\left(5\cdot 10^3\right)\right]$ & $0.04$ & $700$ \\
			FMNIST (\emph{trouser} vs. \emph{pullover}) (DNN) & $150$ & $0.100$ & $[0.0, 1.0]^{784}$ & $\left(-\infty, \log\left(5\cdot 10^3\right)\right]$ & $0.03$ & $800$\\
			CIFAR-10 (\emph{airplane} vs. \emph{frog}) (DNN) & $120$ & $0.100$ & $[-1, 1]^{3,072}$ & $\left(-\infty,\log\left(10^5\right)\right]$ & $0.03$ & $800$ \\
			\hline
	\end{tabular}}
	\label{tabAttack}

\end{table*}

\begin{table}[!t]

	\centering
	{
	\caption{Experimental settings for testing the attacks.}{ 
		\begin{tabular}{|l|c|c|}
			\hline
			Dataset (Model) &  $\eta_\text{tr}$ &  {\#} Epochs \\
			\hline
			MNIST (`0' vs. `8') (LR) & $0.10$ & $140$ \\
			FMNIST (\emph{trouser} vs. \emph{pullover}) (LR) & $0.04$ & $160$ \\
			CIFAR-10 (\emph{airplane} vs. \emph{frog}) (LR) & $0.01$ & $500$ \\
			MNIST (`0' vs. `8') (DNN) & $0.04$ & $700$ \\
				FMNIST (\emph{trouser} vs. \emph{pullover}) (DNN) & $0.03$ & $800$ \\
			CIFAR-10 (\emph{airplane} vs. \emph{frog}) (DNN) & $0.03$ &  $800$  \\
			\hline
	\end{tabular}}
	\label{tabTrain}
	}
\end{table}

\end{samepage}

We evaluate the effectiveness of the attack strategy in Eq.~(\ref{eqAttacker2}) against LR and feed-forward DNNs. We study the influence of $L_2$ and $L_1$ regularization on the attack, providing an analysis of the robustness of the learning algorithms to worst-case scenarios for attacks with different strengths. Note that the analysis of optimal indiscriminate poisoning attacks against non-convex models is substantially more computationally difficult. Most previous work in optimal poisoning attacks focuses on linear classifiers and, to our knowledge, our study is the first to analyze the effect of regularization against data poisoning on DNNs.

\subsection{Experimental Settings}

\label{subsec:expset}

For both LR and DNNs, we use three different binary classification problems: MNIST (`0' vs. `8') \cite{lecun1998gradient}, FMNIST (\emph{trouser} vs. \emph{pullover}) \cite{xiao2017fashion}, and CIFAR-10 (\emph{airplane} vs. \emph{frog}) \cite{krizhevsky2009learning}. All datasets are balanced and drawn at random from the original joint pool of training and test points.  The details for each dataset are included in Table~\ref{tabDatasets}.

All our results are the average of $10$ repetitions with different random data splits for training, validation and test sets. Moreover, both MNIST and FMNIST sets are normalized to be in the range $[0, 1]^{784}$, whereas CIFAR-10 sets are normalized to be in the range $[-1, 1]^{3,072}$. For all the attacks, we measure the average test error for different attack strengths, where the number of poisoning points ranges from $0$ ($0\%$) to $1,750$ ($35\%$).  The size of the batch of poisoning points that are simultaneously optimized is $350$ for all the datasets. For MNIST and FMNIST, this leads to $274,400$ features to be optimized simultaneously, and to $1,075,200$ features for CIFAR-10. In this way, we simulate six different ratios of poisoning ranging from $0\%$ to $35\%$.

We simulate different ratios of poisoning points in a cumulative manner: Once the optimization of the current batch of poisoning points and hyperparameters is finished,\footnote{The criterion to finish the loop that optimizes the variables of the outer level problem is given by the number of hyperiterations.} this batch of poisoning points is fixed and the next batch of poisoning points is replaced into the remaining clean training set, whereas the hyperparameters are re-initialized, to carry out their corresponding optimization.  To accelerate their optimization, the hypergradients for the poisoning points are normalized with respect to their $L_2$ norm, and the hypergradients for each $\boldsymbol{\Lambda}$ are also normalized with respect to their corresponding value.\footnote{The analysis of other techniques to accelerate the optimization, such as adaptive learning rates, is left for future work.}

The LR classifier's parameters are always initialized with zeros, for all the datasets. The DNN models have two hidden layers with Leaky ReLU activation functions as follows: $784\times32\times8\times1$, i.e., $25,393$ parameters, for MNIST and FMNIST; and $3,072\times64\times32\times1$, i.e., $198,785$ parameters, for CIFAR-10. In DNN models, these parameters are initially filled with values according to Xavier Initialization method \cite{glorot2010understanding}, using a uniform distribution for all the parameters except the bias terms, which are initialized with a value of $10^{-2}$. 

For all the experiments, we make use of SGD both to update the parameters in the forward pass of RMD, and to train the model when testing the attack (full batch training). The choice of the number of iterations for the inner problem, $T$, depends on the model and the training dataset. Low values of $T$ could lead to low-quality approximations for the hypergradient. As $T$ increases, the solution of RMD approaches the exact (true) hypergradient, but at the risk of overfitting the outer objective in the bilevel optimization problem \cite{franceschi2018bilevel}. The details of the attack settings are shown in Table~\ref{tabAttack}, whereas the ones for testing the attacks are in Table~\ref{tabTrain}.

All the experiments have been run on $2 \times 11$~GB NVIDIA GeForce\textregistered \hspace{0cm}  GTX 1080 Ti GPUs. The RAM memory is $64$~GB ($4\times16$~GB) Corsair VENGEANCE DDR4 $3000~\text{MHz}$. The processor (CPU) is Intel\textregistered \hspace{0cm} Core\texttrademark \hspace{0cm} i7 Quad Core Processor i7-7700k ($4.2$~GHz) $8$~MB Cache.

\subsection{Logistic Regression}

 \subsubsection{Test Error and Value of \texorpdfstring{$\lambda$}~~Learned}

\begin{figure*}[!t]
\centering
\subfloat[]{\includegraphics[width=1.9in]{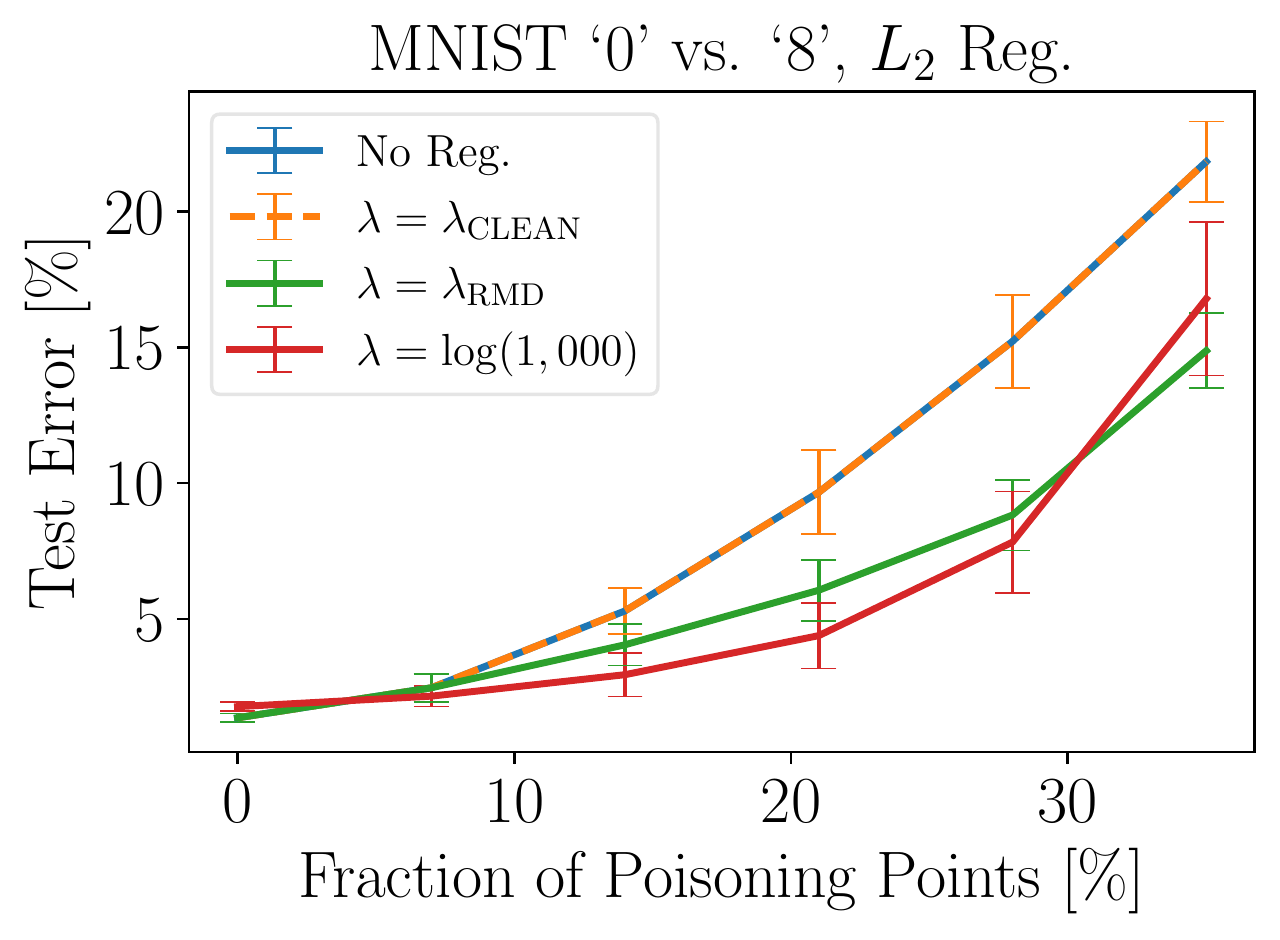}%
\label{fig:lropt_a}}
\subfloat[]{\includegraphics[width=1.8in]{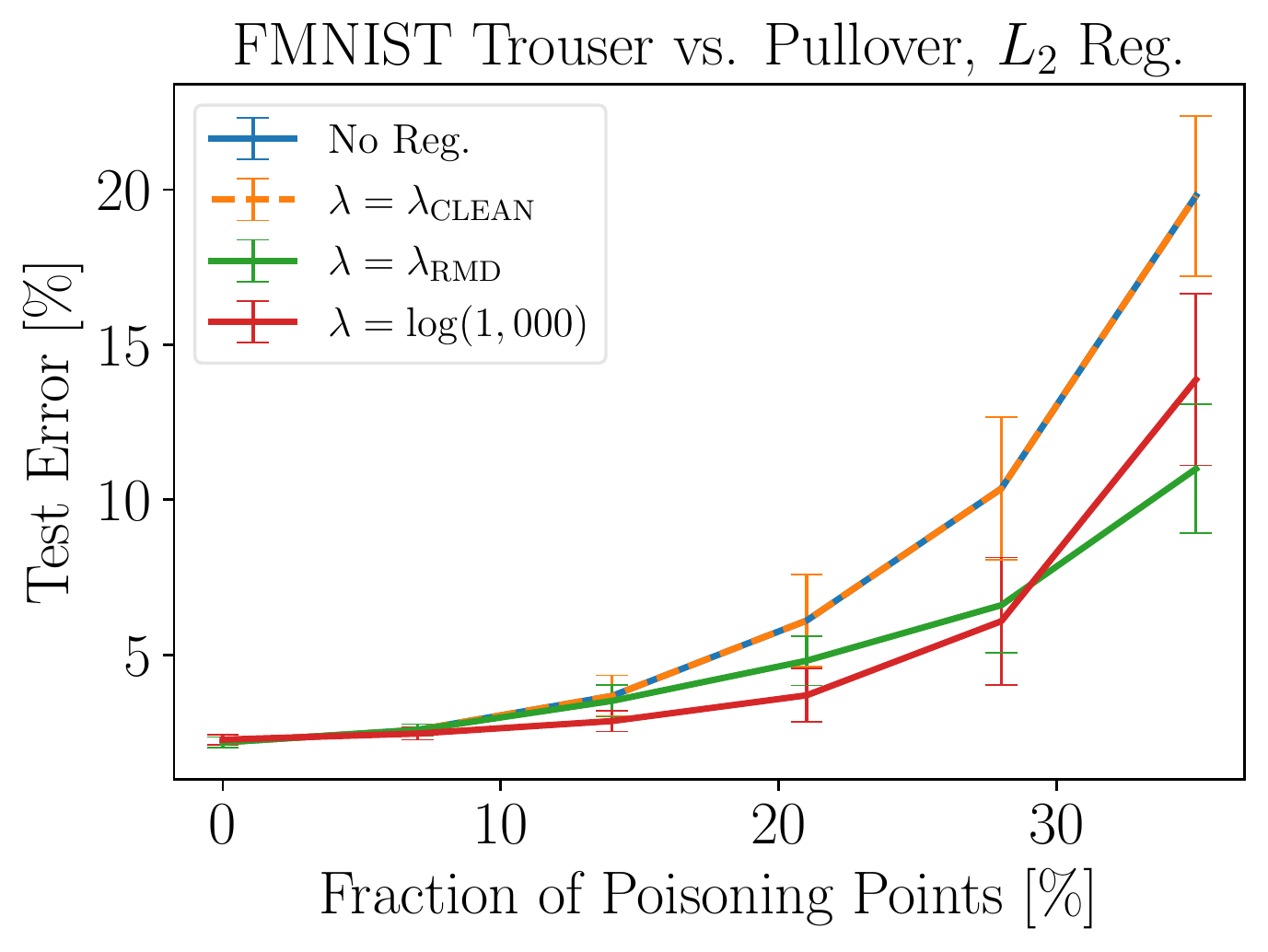}%
\label{fig:lropt_b}}
\subfloat[]{\includegraphics[width=1.9in]{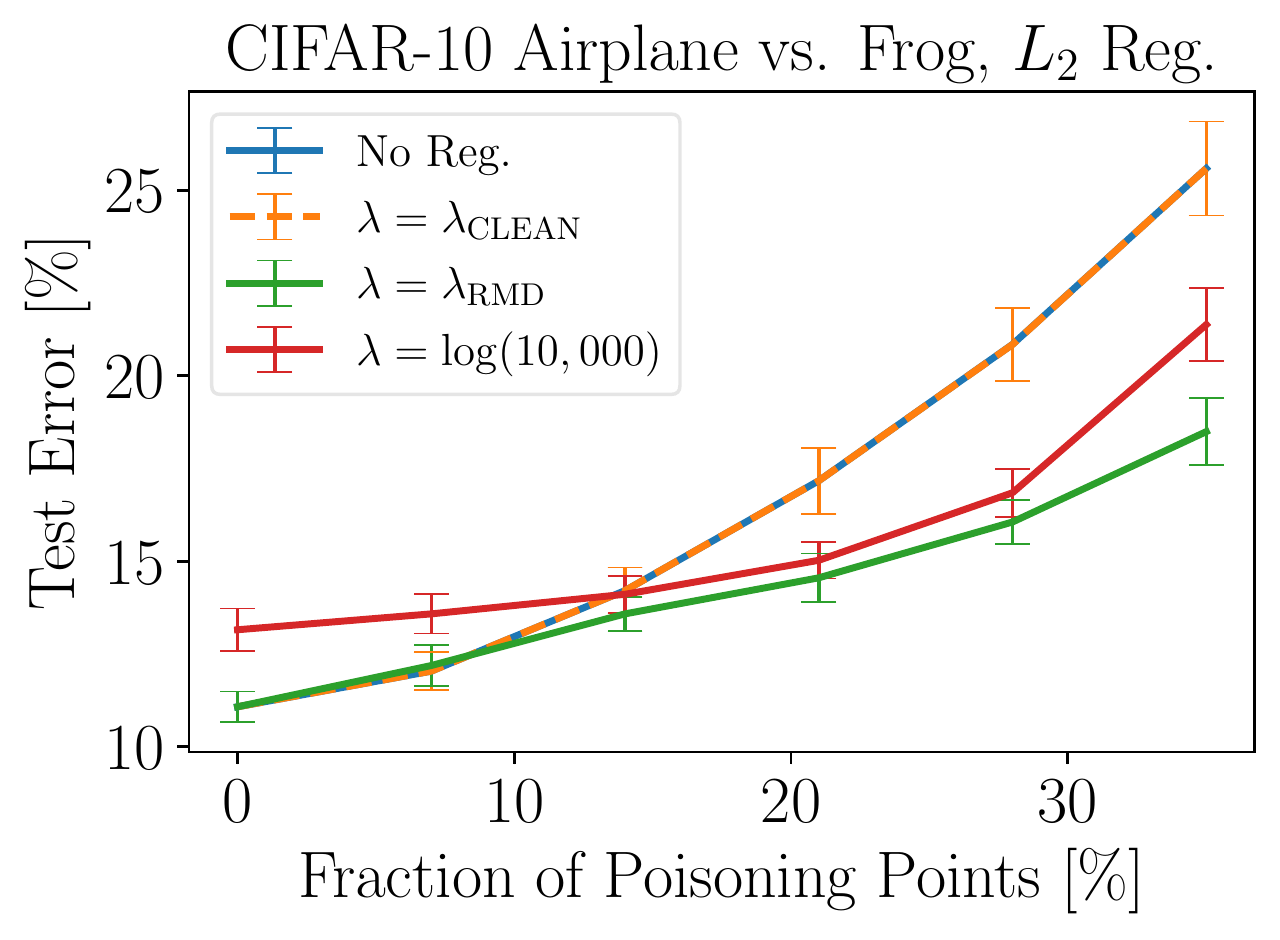}%
\label{fig:lropt_c}}
\\
\subfloat[]{\includegraphics[width=1.9in]{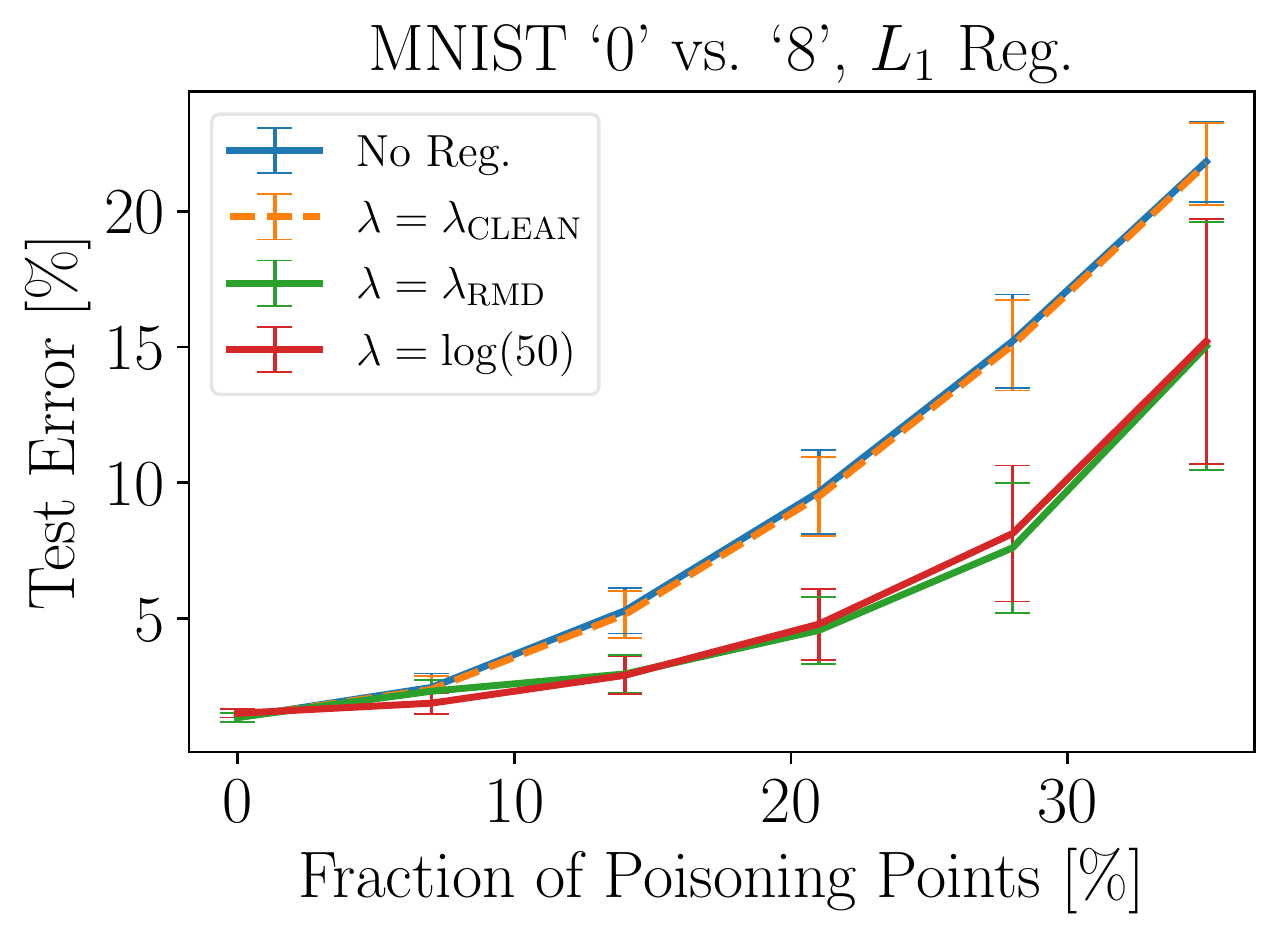}%
\label{fig:lropt_d}}
\subfloat[]{\includegraphics[width=1.9in]{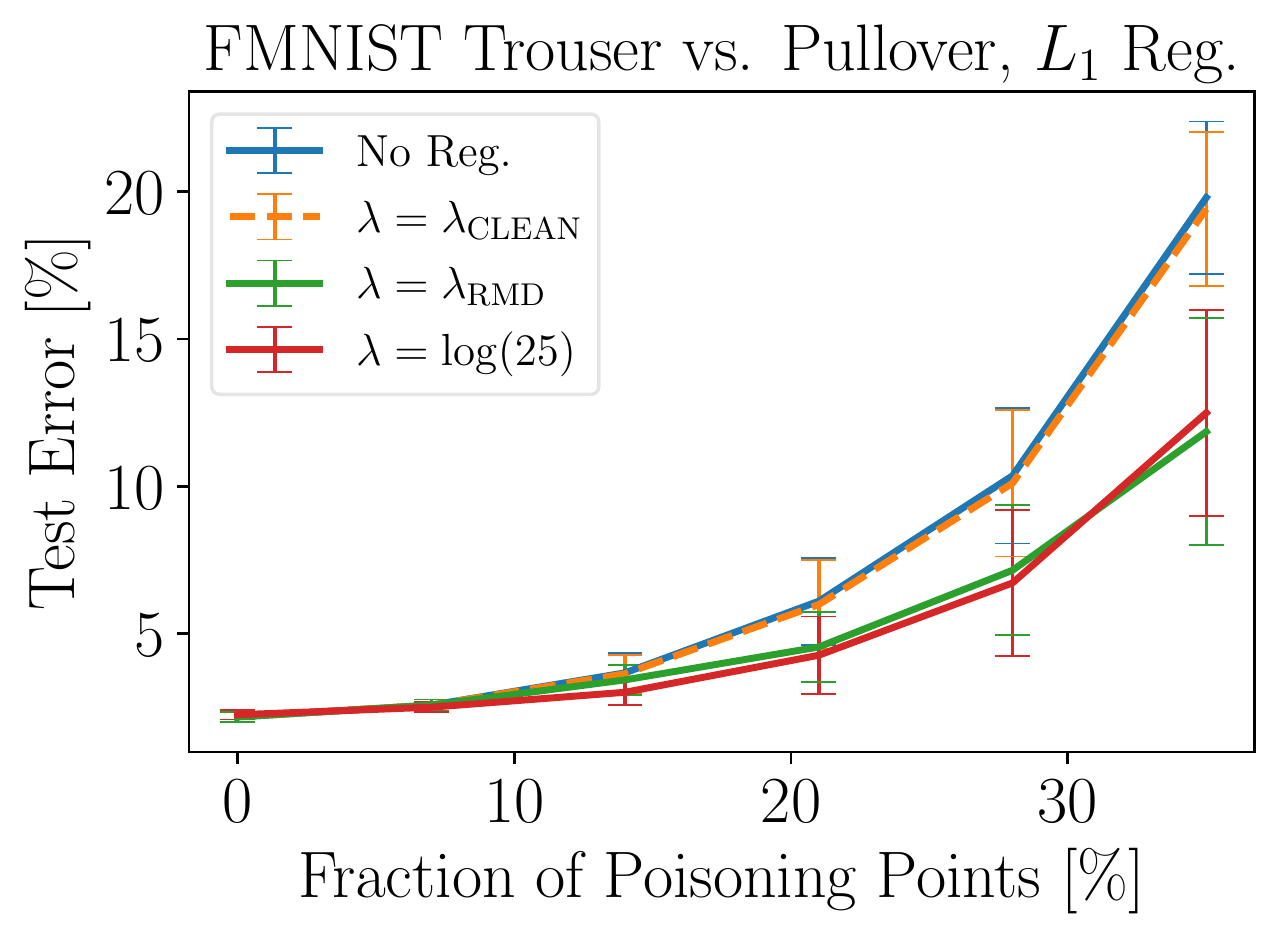}%
\label{fig:lropt_e}}
\subfloat[]{\includegraphics[width=1.9in]{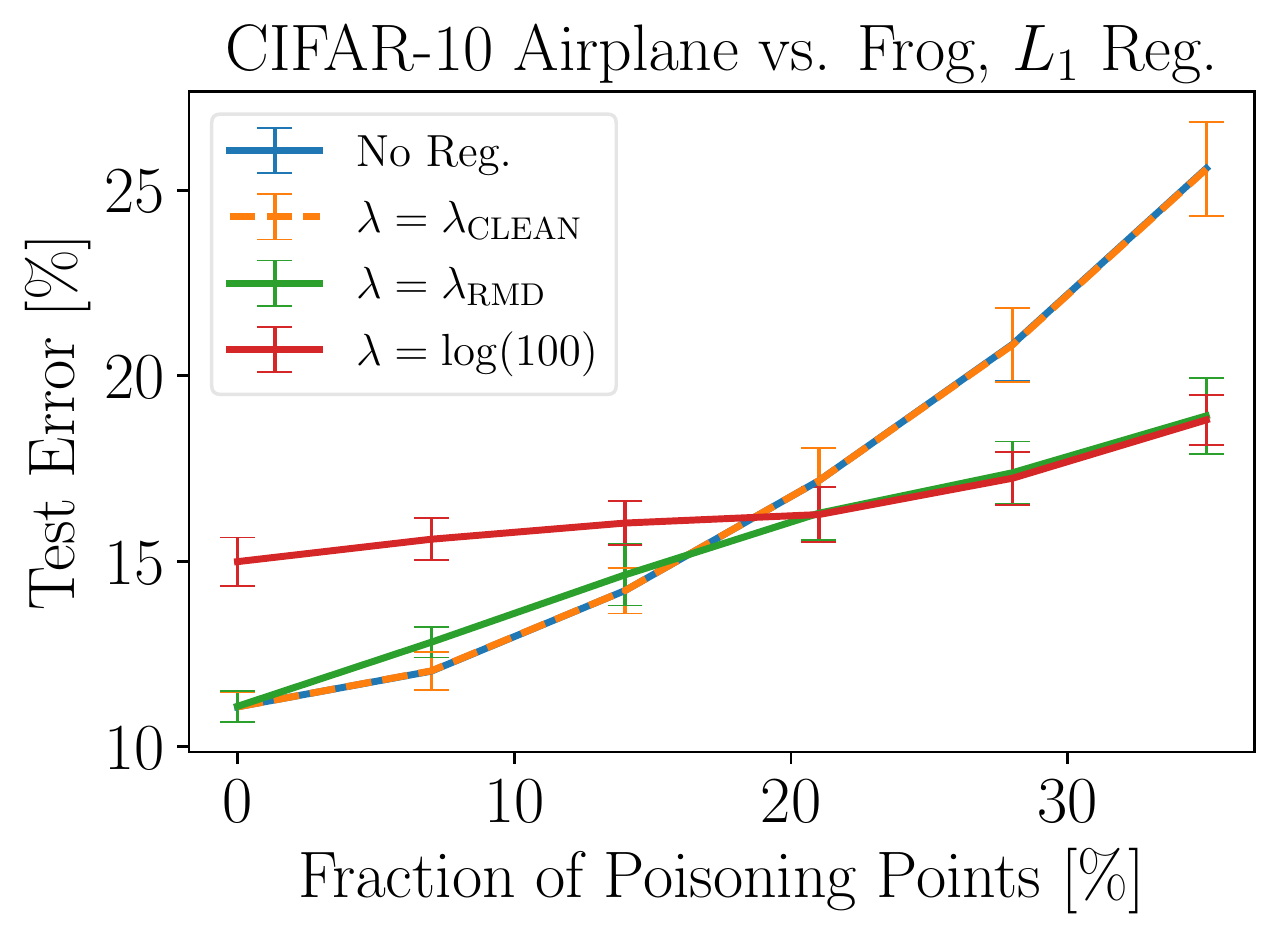}%
\label{fig:lropt_f}}
\caption{Average test error for the optimal attack against LR: The first row represents the case of $L_2$ regularization on (a) MNIST, (b) FMNIST, and (c) CIFAR-10. The second row contains the plots for $L_1$ regularization on (d) MNIST, (e) FMNIST, and (f) CIFAR-10.}
\label{fig:lropt}
\end{figure*}

\begin{figure}[!t]
\centering
\subfloat[]{\includegraphics[width=1.8in]{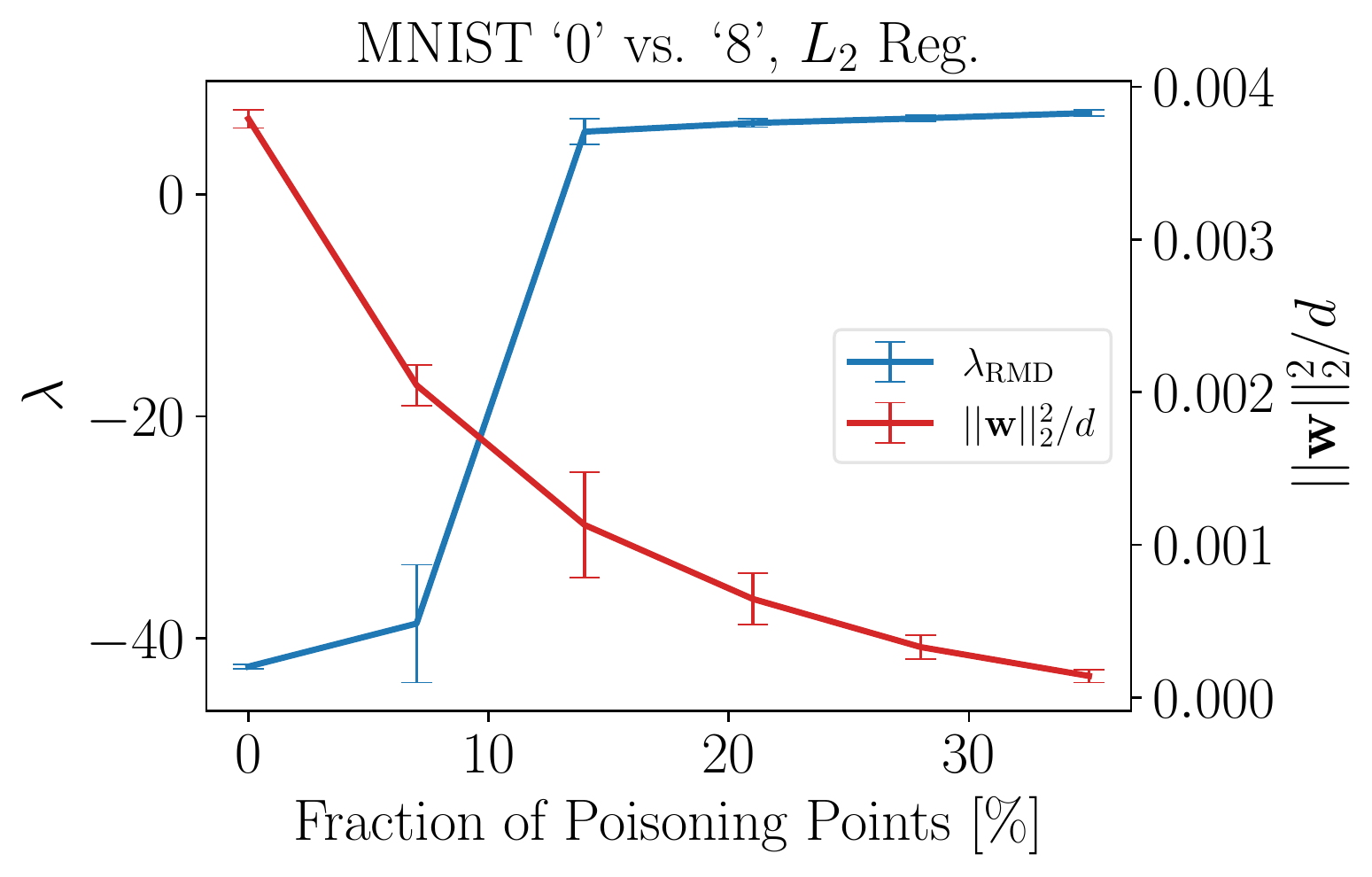}%
\label{fig:lrlambd_a}}
\subfloat[]{\includegraphics[width=1.8in]{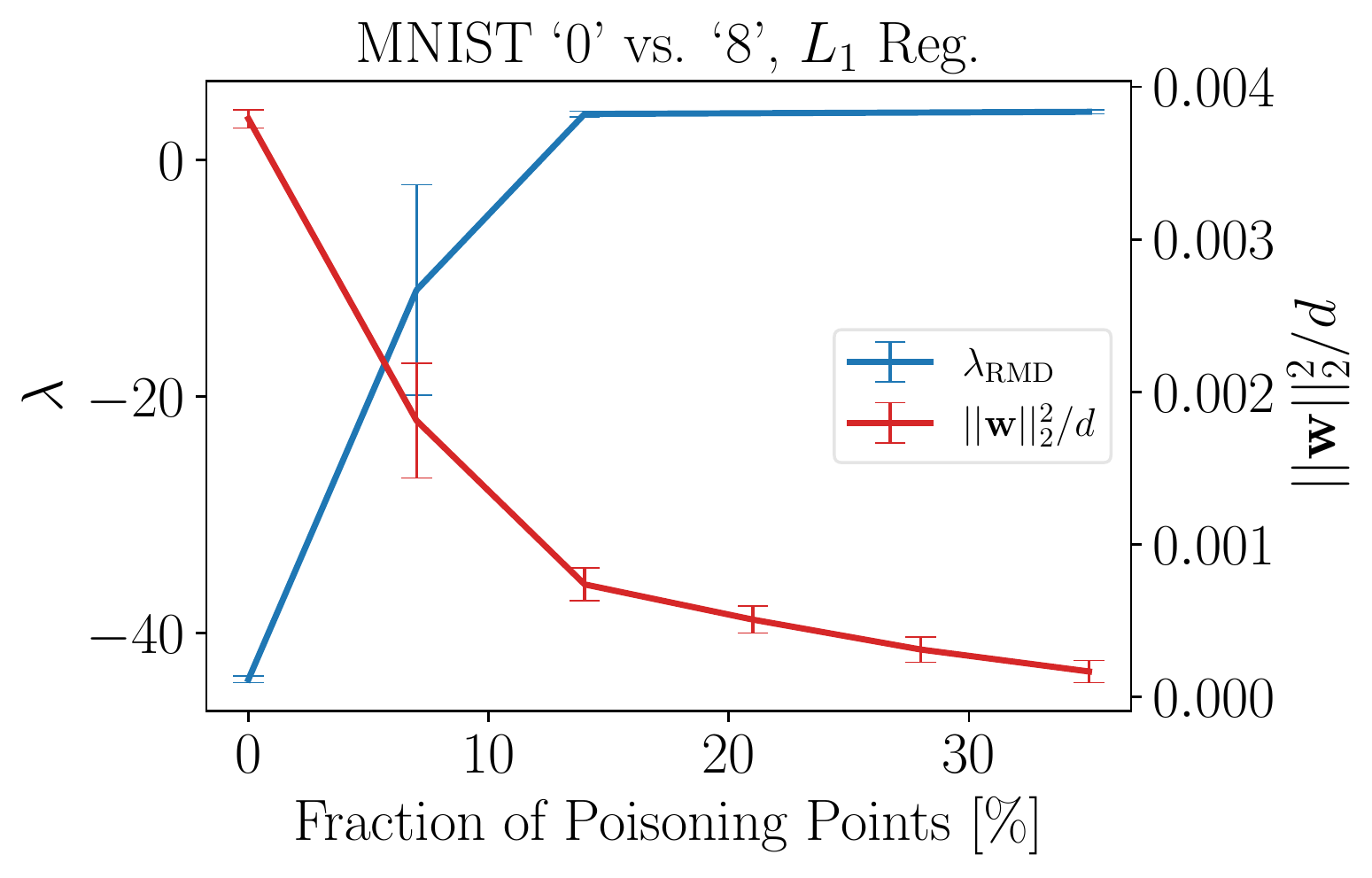}%
\label{fig:lrlambd_d}}
\caption{Average $\lambda$ and $||{\bf w}||_2^2$ for the optimal attack against LR on MNIST: (a) $L_2$ regularization; (b) $L_1$ regularization.}
\label{fig:lrlambd}
\end{figure}

For LR we test the general poisoning attack strategy in Eq.~(\ref{eqAttacker2})---labeled as $\lambda_\text{RMD}$ in the figures---using the following settings for the computation of the hypergradients with RMD. For MNIST we set $T$, the number of iterations for the inner problem, to $140$. For FMNIST and CIFAR-10 we use $T=160$ and $T=500$, {respectively}.
{For comparison purposes, in addition to crafting attacks learning the value of $\lambda$, $\lambda_\text{RMD}$, we also craft optimal poisoning attacks setting the value of $\lambda$ to different constant values: no regularization ($\lambda =-\infty$); a very large one (for $L_2$ regularization: $\lambda = \log (1,000)$ for MNIST and FMNIST, and $\lambda = \log (10,000)$ for CIFAR-10; for $L_1$ regularization: $\lambda = \log (50)$ for MNIST, $\lambda = \log (25)$ for FMNIST, and $\lambda = \log (100)$ for CIFAR-10); and the value of $\lambda$ optimized with $5$-fold cross-validation ($\lambda_\text{CLEAN}$).} By comparing with no regularization and large constant values for $\lambda$, we aim to show the trade-off between accuracy (under clean data) and robustness to different attack strengths. The case of $\lambda_\text{CLEAN}$ is similar to the settings used in \cite{xiao2015feature}, which uses a methodology akin to \cite{friedman2010regularization}, where the authors use $K$-fold cross-validation to select the value of $\lambda$, and the clean data is used both for training and validation in an unbiased way.

The results are shown in Fig.~\ref{fig:lropt}. We observe that when the model is not regularized or uses $\lambda_\text{CLEAN}$, the attacks are very effective and the test error increases significantly when compared to the algorithm's performance on the clean dataset ($0\%$ of poisoning). In contrast, for the largest $\lambda$ the test error increases moderately with the increasing fraction of poisoning points, showing a lower test error compared to the case of no regularization. However, in the absence of an attack, the algorithm \emph{underfits} and the error is higher compared to the other models (especially in the case of CIFAR-10). When the value of $\lambda$ is learned ($\lambda_\text{RMD}$) using the trusted validation dataset, the increase in the test error is moderate and, when the ratio of poisoning points is large, the performance is similar to when $\lambda$ is large. We can also observe that, in this case, when there is no attack, the performance is similar to that of the non-regularized classifier.

The results in Fig.~\ref{fig:lropt} also show that the attack and the methodology presented in \cite{xiao2015feature} provide an overly pessimistic view on the robustness of $L_2$ and $L_1$ regularization to poisoning attacks, and that using the hyperparameter learned when the data is clean can be detrimental under data poisoning. We show that, by appropriately selecting the value of $\lambda$, we can effectively reduce the impact of such attacks. We can also observe that there is a trade-off between accuracy and robustness: over-regularizing (i.e., setting a very large value for $\lambda$) makes the algorithm more robust to the attack, but the performance on clean data is degraded.

In Fig.~\ref{fig:lrlambd} we show the value of $\lambda$ learned and the norm of the model's parameters divided by the number of parameters, $||{\bf w}||^2_2/d$, as a function of the fraction of poisoning points injected. We observe that the regularization hyperparameter increases and then saturates as we increase the fraction of poisoning points. Thus, the regularization term compensates the effect of the poisoning points on the model's parameters up to a point.

Comparing $L_2$ and $L_1$, we observe that both regularization techniques provide similar mitigation effects against the attack. Thus, even if $L_1$ regularization does not necessarily provide stability to the learning algorithm, as is the case of $L_2$ regularization, the use of the trusted validation set for learning the regularization hyperparameter helps to mitigate the impact of the attack in both cases. The presence of the poisoning points increases the norm of the parameters if no regularization is applied. But, when the trusted validation dataset is available for selecting the regularization parameter, both $L_1$ and $L_2$ regularization are capable of mitigating this effect, and thus, of reducing the impact of the poisoning points.

\subsubsection{Sensitivity Analysis of the Size of the Validation Set}

\begin{figure*}[!t]
\centering
\subfloat[]{\includegraphics[width=1.9in]{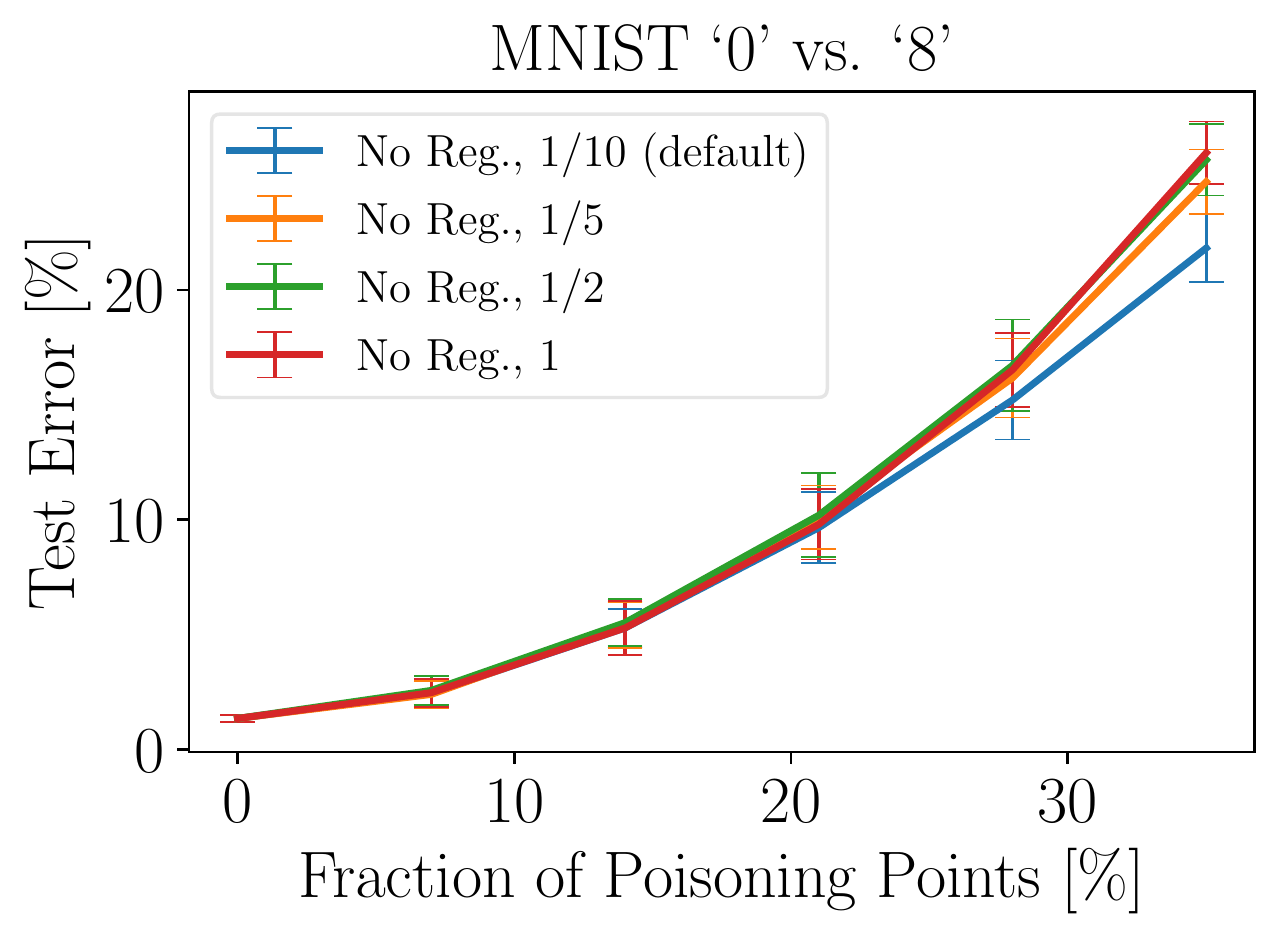}%
\label{fig:lr_val_a}}
\subfloat[]{\includegraphics[width=1.9in]{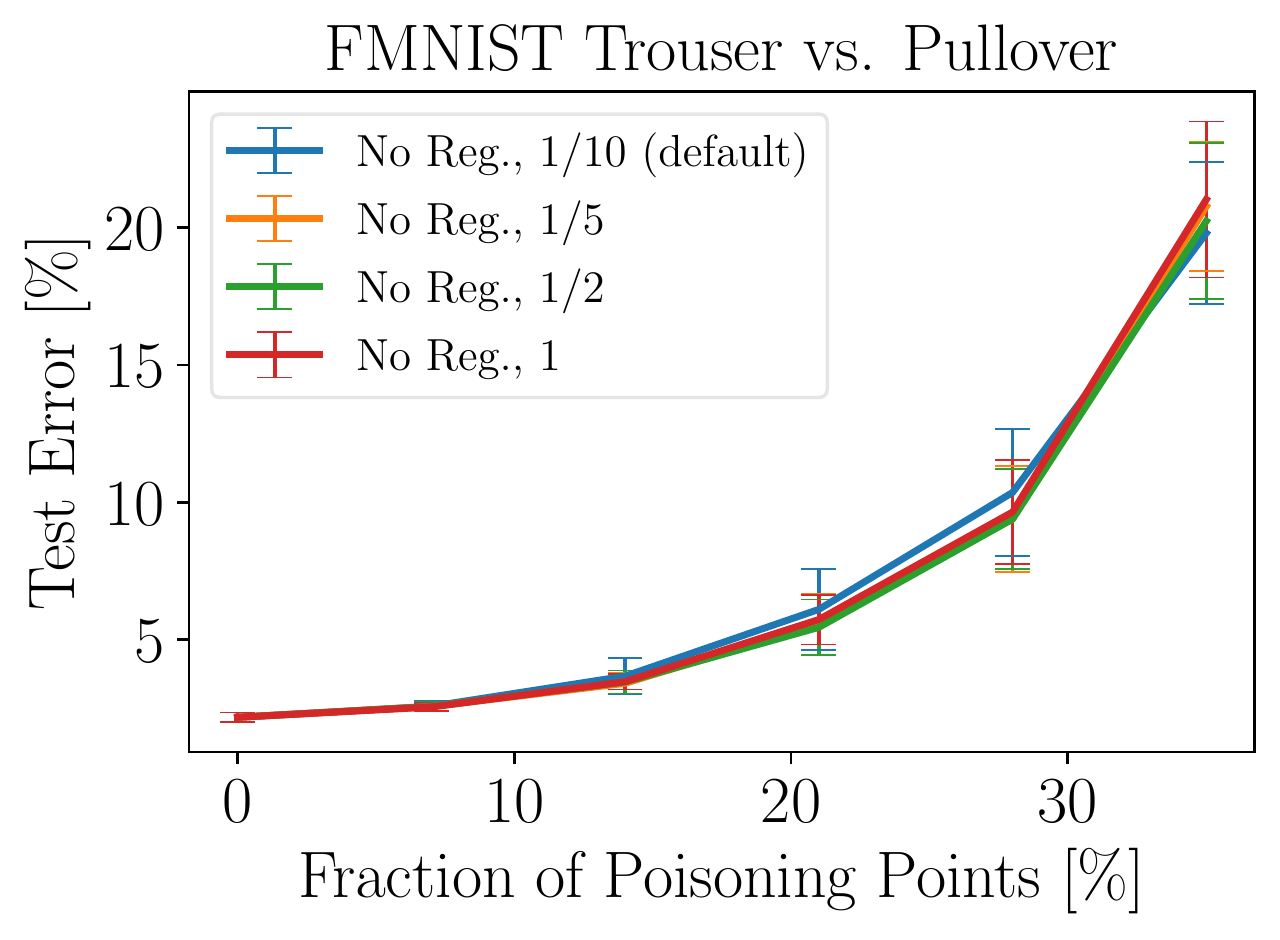}%
\label{fig:lr_val_d}}
\subfloat[]{\includegraphics[width=1.9in]{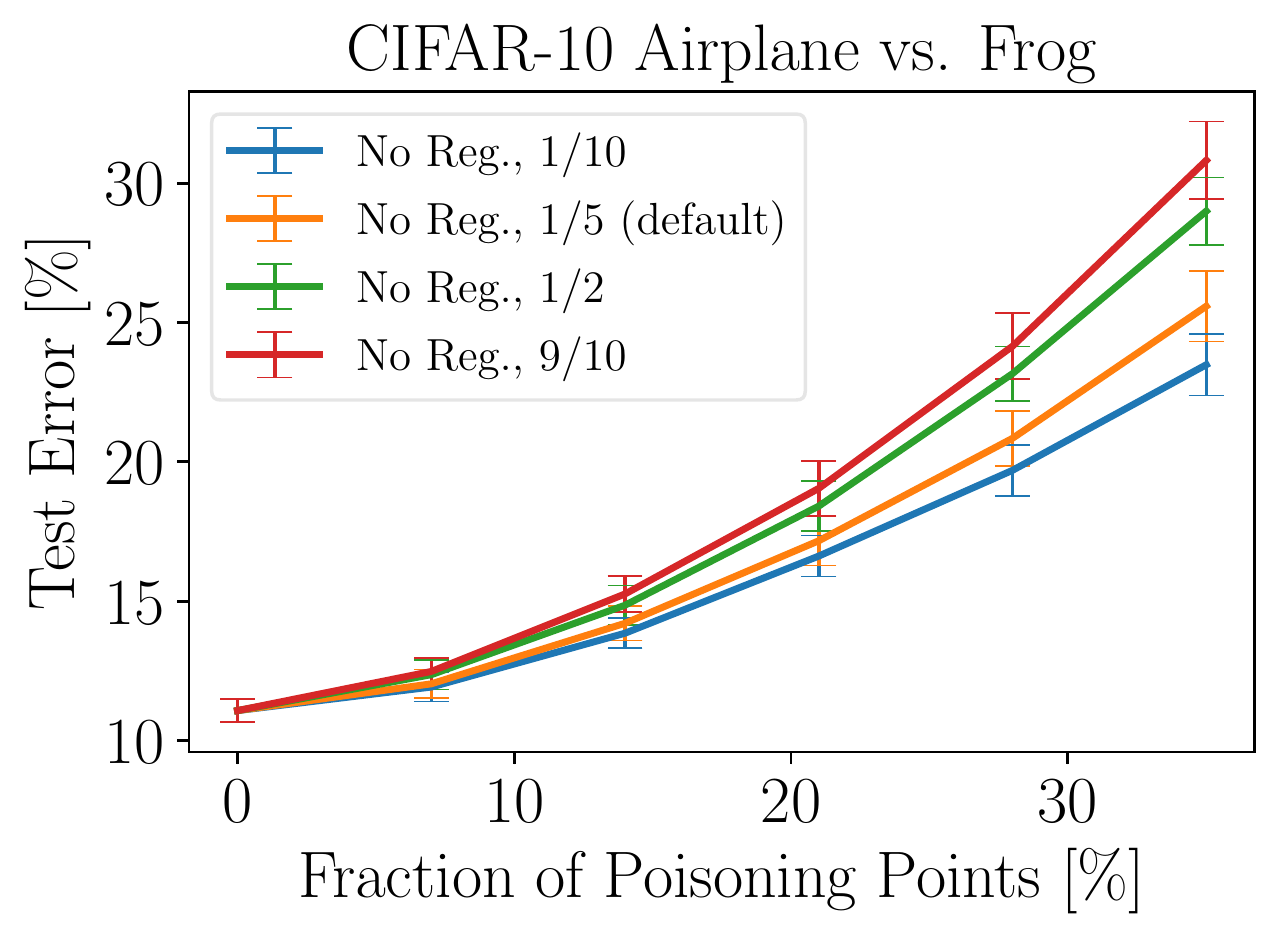}%
\label{fig:lr_val_g}}
\vspace{-0.2cm} \\
\subfloat[]{\includegraphics[width=1.9in]{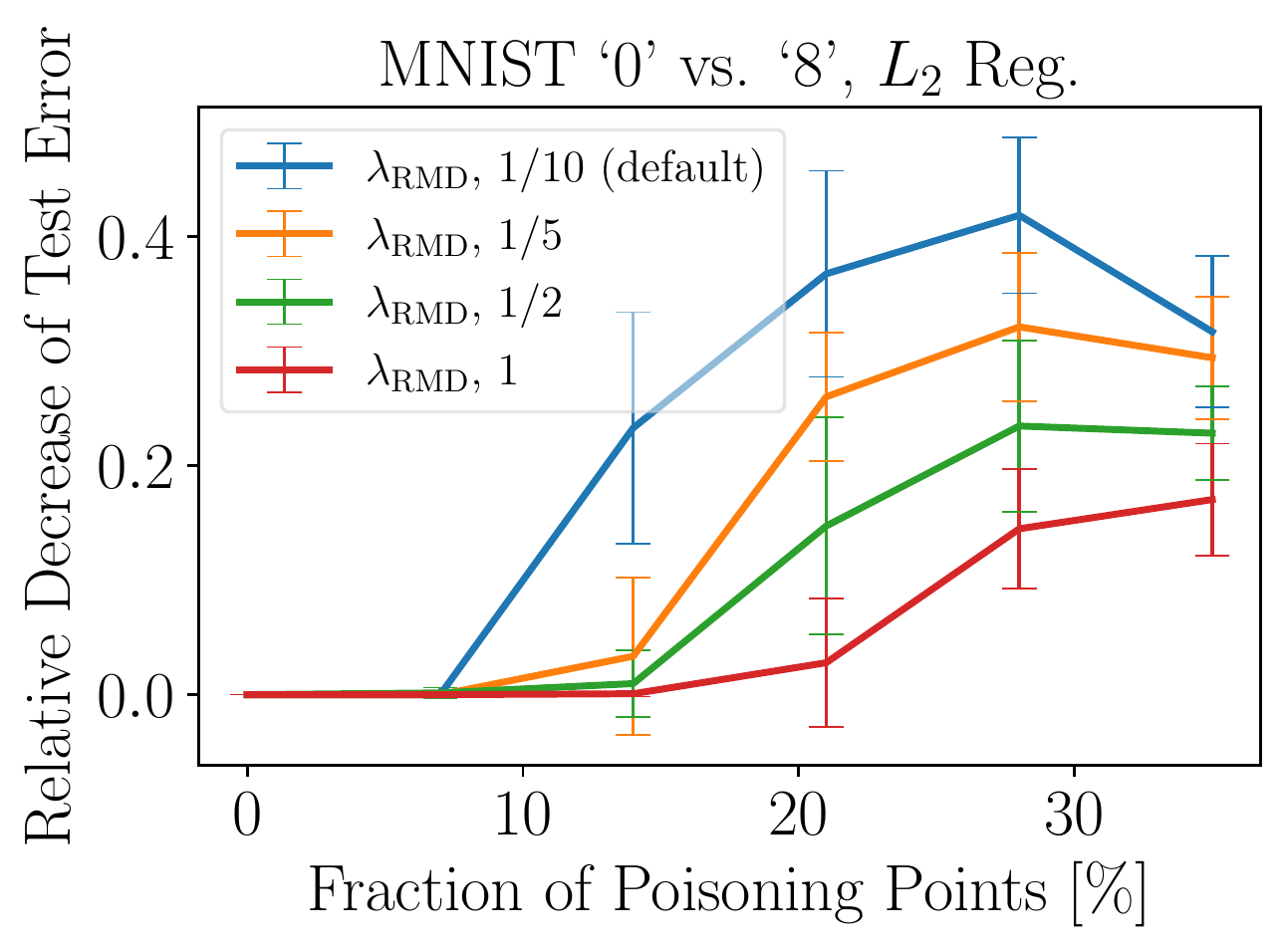}%
\label{fig:lr_valt_b}}
\subfloat[]{\includegraphics[width=1.9in]{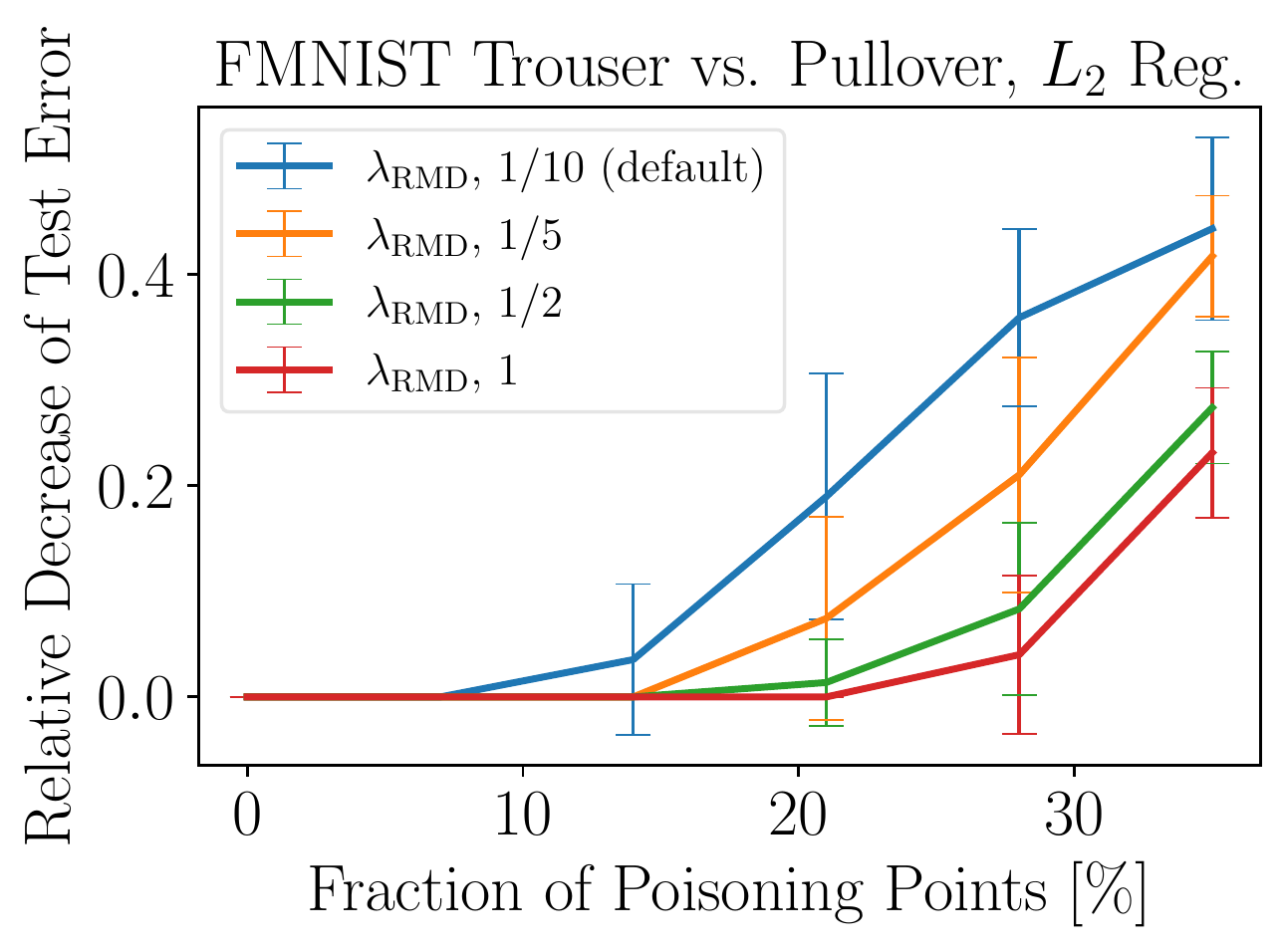}%
\label{fig:lr_valt_e}}
\subfloat[]{\includegraphics[width=1.9in]{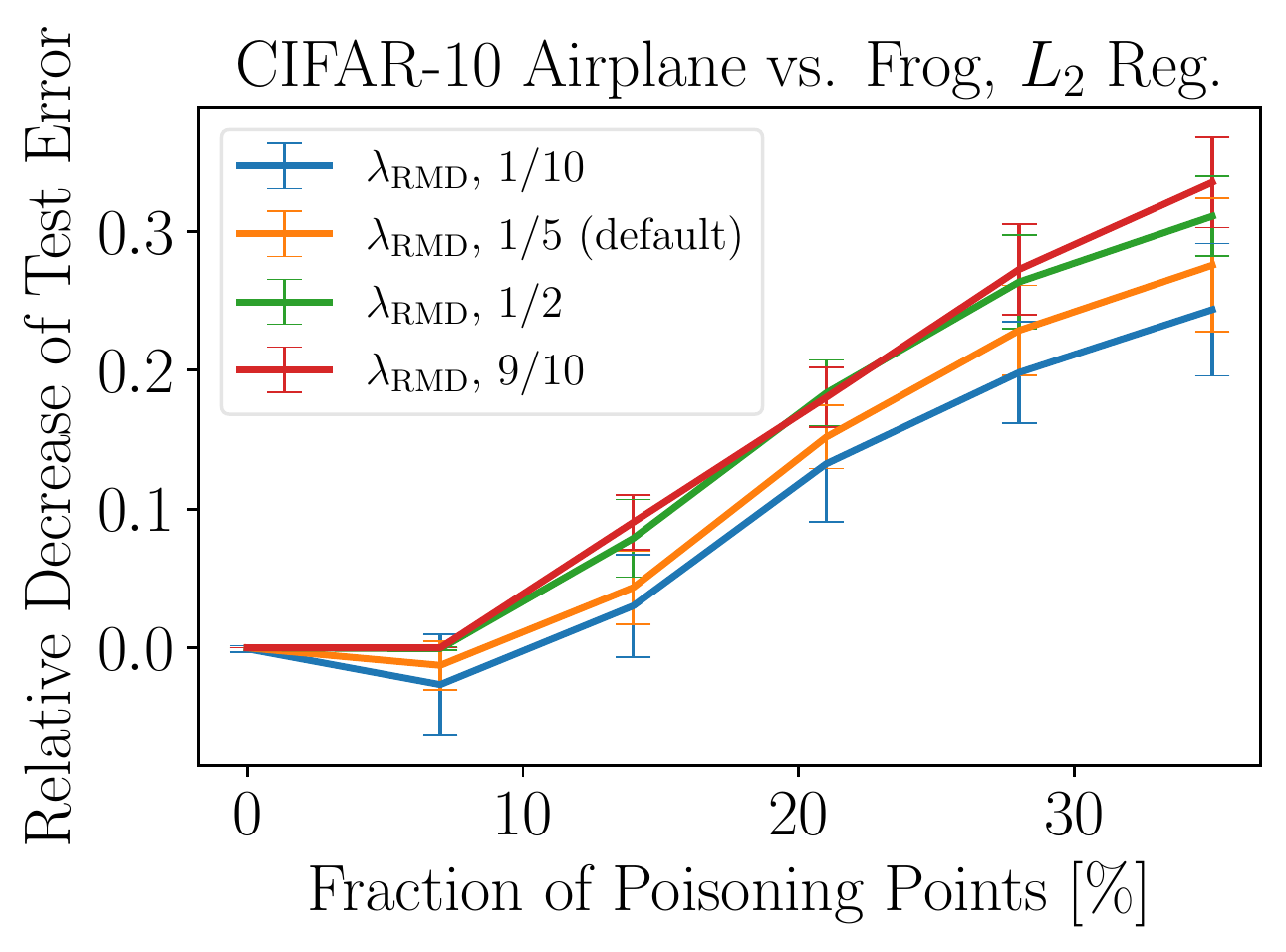}%
\label{fig:lr_valt_h}}
\caption{Sensitivity analysis of the size of the validation set. The first, second and third { column represent the results for MNIST, FMNIST and CIFAR-10, respectively. The first and second row show the results when there is no regularization and for $L_2$ regularization, respectively.}}
\label{fig:lr_val}
\end{figure*}

The size of the trusted validation set has an effect not only on the selection of the hyperparameters, but also on the effectiveness of the poisoning points learned using the attack in Eq.~(\ref{eqAttacker2}) when evaluated on a separate test set. Note that having a larger trusted dataset is not necessarily beneficial only for the learner, but also for the attacker, who, under worst-case scenario assumptions, also has access to the trusted validation set.
To study this effect, we consider an LR classifier and the same datasets (i.e., MNIST, FMNIST and CIFAR-10) and settings as before. Previously, we assumed that the validation set was ten times smaller than the training set for MNIST and FMNIST, and five times smaller for CIFAR-10. Now, the size of the training and test sets is fixed, and we evaluate different sizes for the validation set---compared to the size of the training set. To analyze the influence of the validation set both when there is no regularization and when there is, we define the relative decrease of test error as the relative difference of the test error obtained when there is no regularization and when the value of $\lambda$ is learned using the trusted validation set, i.e., $(\text{Test Error}_{|\text{No Reg.}} - \text{Test Error}_{|\lambda_{\text{RMD}}}) /  \text{Test Error}_{|\text{No Reg.}}$. 

In Fig.~\ref{fig:lr_val} { and Fig.~\ref{fig:lr_val_2}}, we observe that when the model is not regularized, for MNIST and CIFAR-10, the test error is higher when the validation set is larger, as the poisoning points do not overfit the validation set. In contrast, for FMNIST the different-size validation sets result in a similar test error. On the other hand, when $\lambda$ is learned ($L_2$ and $L_1$ regularization), for MNIST and FMNIST the test error decreases when the validation set is smaller, whereas for CIFAR-10, the opposite occurs. This shows that having a larger validation set is not always advantageous. When the poisoning points are learned with no regularization, a larger validation set provides more effectiveness for the attack, reducing the overfitting of the attack points. However, when using regularization and the poisoning points and hyperparameters are jointly learned, the optimal size of the validation set can be task-dependent.  Our results show that, with this interplay between the learner and the attacker, the net benefit for the learner depends on the specific classification task, the size of the validation set and the attack strength. However, it is also important to note that, across all experiments, there is a clear benefit for using regularization to mitigate the impact of the attack in all cases and, especially, for strong attacks.

\subsubsection{Consistency Index}

\begin{figure*}[!t]
\centering
\subfloat[]{\includegraphics[width=1.9in]{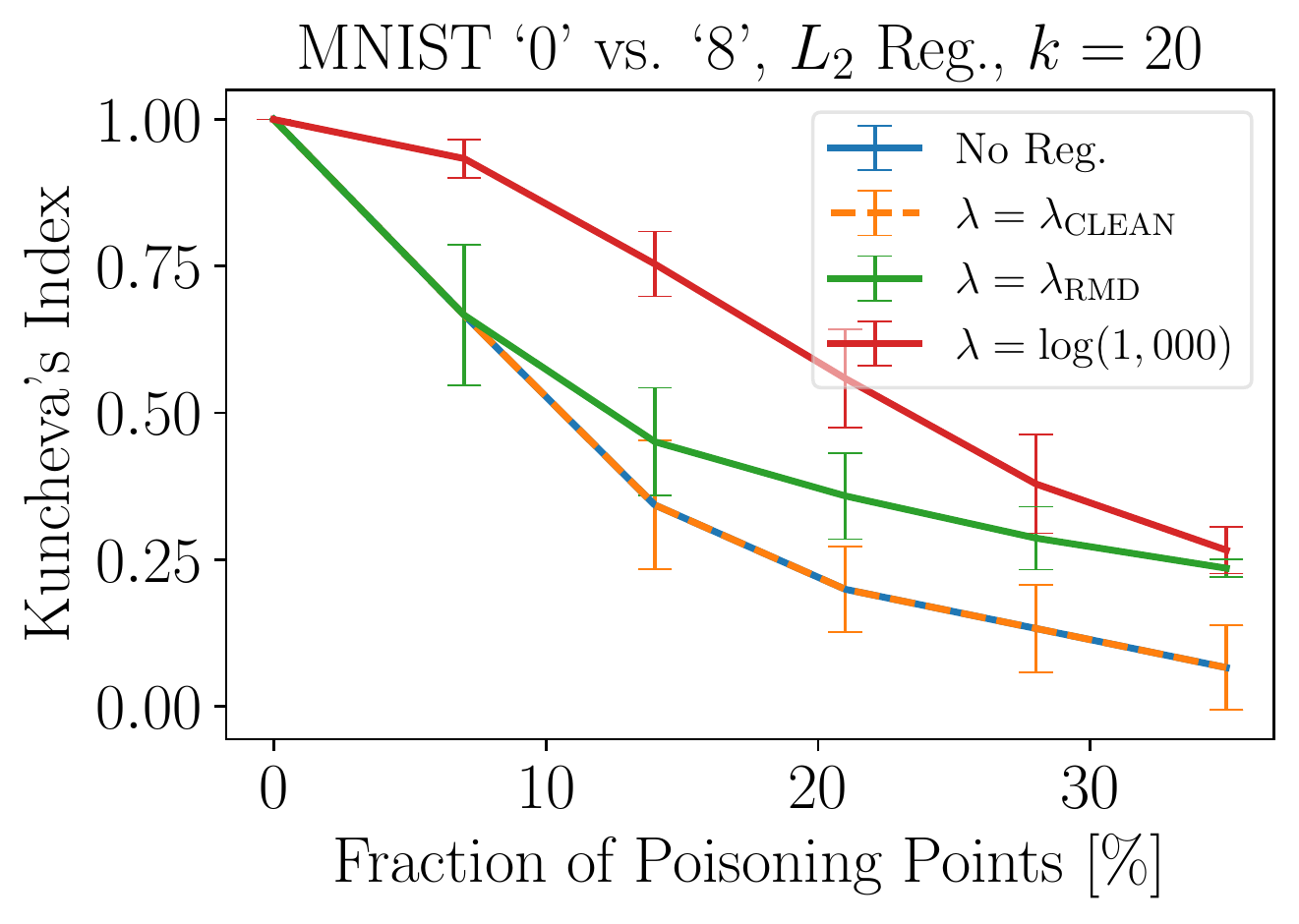}%
\label{fig:kunchevamnist_a}}
\subfloat[]{\includegraphics[width=1.95in]{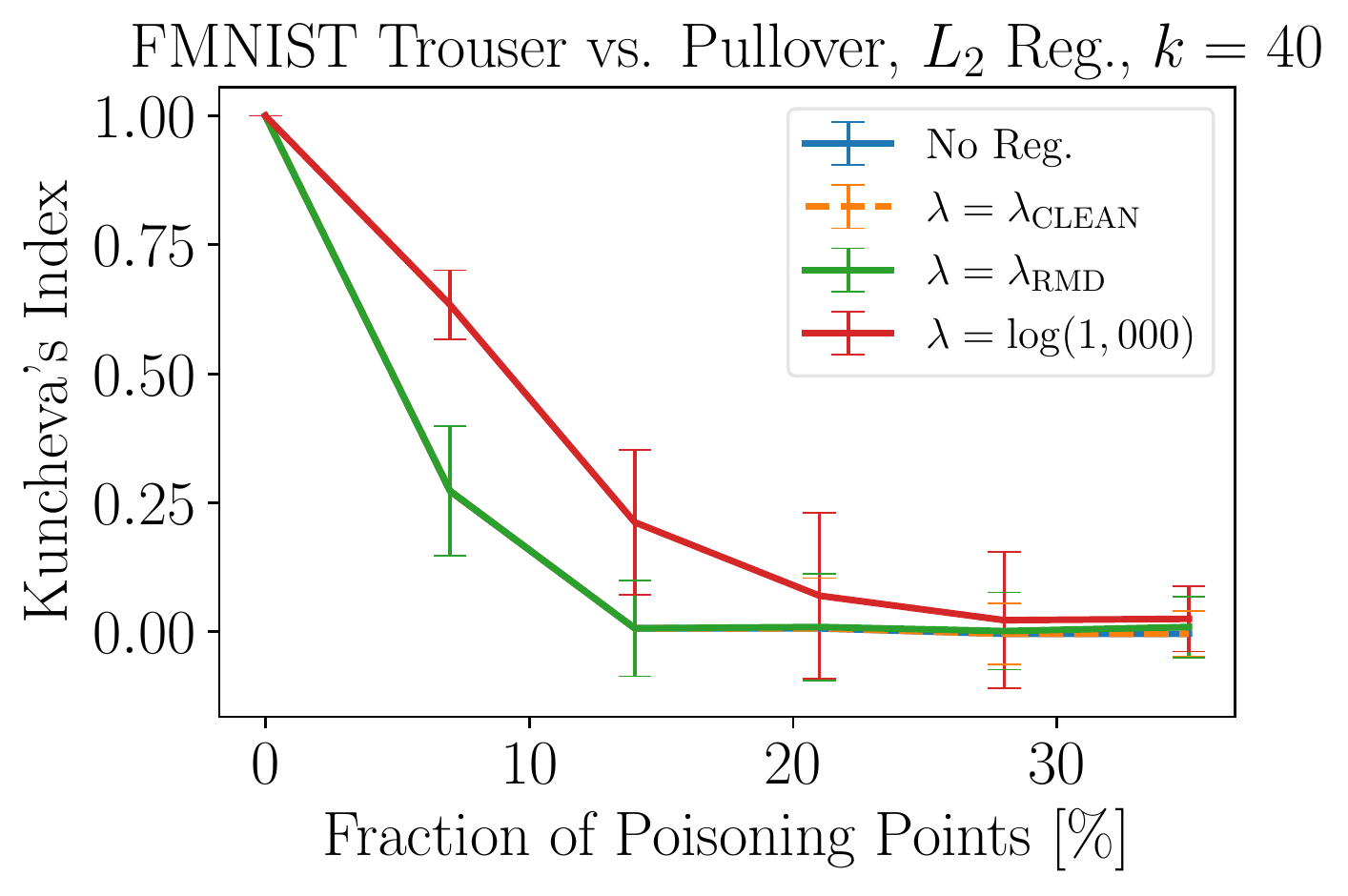}%
\label{fig:kunchevafmnist_a}}
\subfloat[]{\includegraphics[width=1.95in]{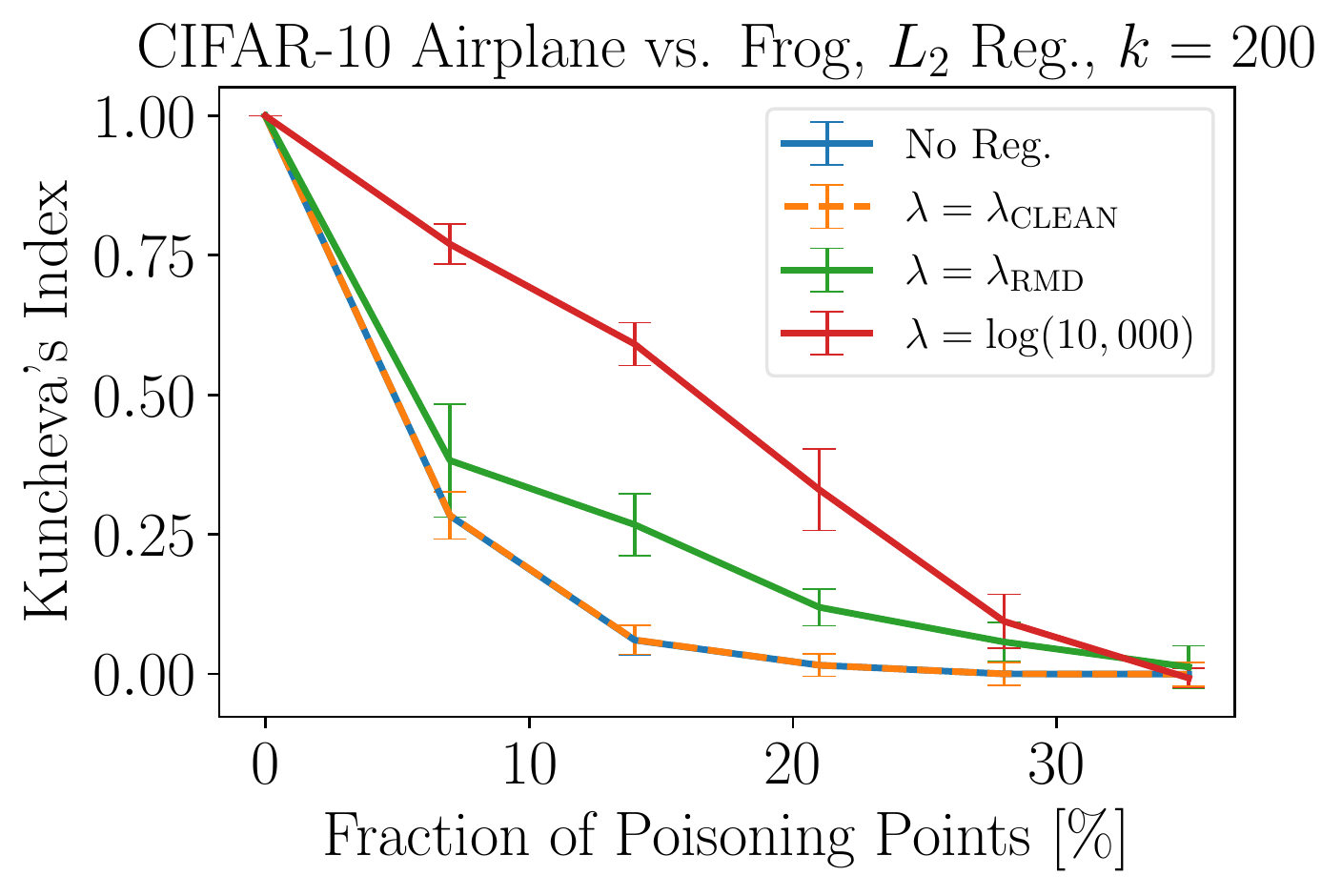}%
\label{fig:kunchevacifar_a}}
\caption{Average Kuncheva's consistency index for the optimal attack against LR using $L_2$ regularization on (a) MNIST, (b) FMNIST, and (c) CIFAR-10.}
\vspace{-.3cm}
\label{fig:kuncheva_l2}
\end{figure*}

To understand how embedded
feature selection methods based on $L_2$  regularization are affected by the attack, we evaluate the stability of feature selection
under poisoning using Kuncheva's consistency index \cite{kuncheva2007stability}. Given two feature subsets
$A, B \subseteq \mathcal{X}$, with $|A| = |B| = k, r = |A \cap B|$, and $0 < k < |\mathcal{X}| = d$, Kuncheva's consistency index is defined as $I_c(A,B) = (rd - k^2)/(k(d - k))$, where positive values indicate similar sets, zero is equivalent
to random selections, and negative values indicate
strong anti-correlation between the feature subsets. The
underlying idea of this consistency index is to normalize the
number of common features in the two sets using a correction for chance
that accounts for the average number of common features
randomly selected out of $k$ trials \cite{xiao2015feature}.

To evaluate how poisoning affects embedded feature selection, we compute this index using for $A$ the feature set selected for the clean training data, and compare it against a set $B$ selected under attack, at different percentages of poisoning. For each scenario,  we consider the first $k$ features exhibiting the highest absolute weight values: for MNIST, given that the most of the features are close to zero, we choose the top $20$, $40$ and $80$ features; for FMNIST, the top $40$, $80$ and $160$ features; and for CIFAR-10, the top $200$, $400$ and $800$ features.

The results for $L_2$ regularization for MNIST, FMNIST and CIFAR-10 are shown in Fig.~\ref{fig:kuncheva_l2}. The corresponding results for $L_1$ regularization are consistent with these and can be found in {Fig.~\ref{fig:kuncheva_l2_2}}. We observe that, in all cases, the consistency index decreases with the ratio of poisoning. This means that, to succeed, the attack naturally modifies the importance of the features of the training set (even if the attack is not specifically designed to do that), so that the poisoned model pays more attention to less relevant features. It is also clear that if the model is not regularized, the features selected are less consistent, and regularization helps to increase the feature stability under poisoning. For $\lambda_\text{RMD}$, it is generally bounded between the cases of no regularization and large value of $\lambda$, showing that the algorithm sacrifices some feature stability to decrease the test error. Compared to $L_1$ {(Fig.~\ref{fig:kuncheva_l2_2})}, $L_2$ regularization provides greater feature stability when using a large regularization hyperparameter. It is important to note that the selection of the regularization hyperparameter, using Eq.~(\ref{eqAttacker2}), aims to minimize the error on the validation set, not to maximize the stability of the features, which would require a different defensive strategy. However, the results in Fig.~\ref{fig:kuncheva_l2} help to understand better the combined effect of the poisoning attack and the use of regularization.

\subsection{Deep Neural Networks}
\label{subsec:expdnn}

\begin{figure*}[!t]
\centering
\subfloat[]{\includegraphics[width=1.9in]{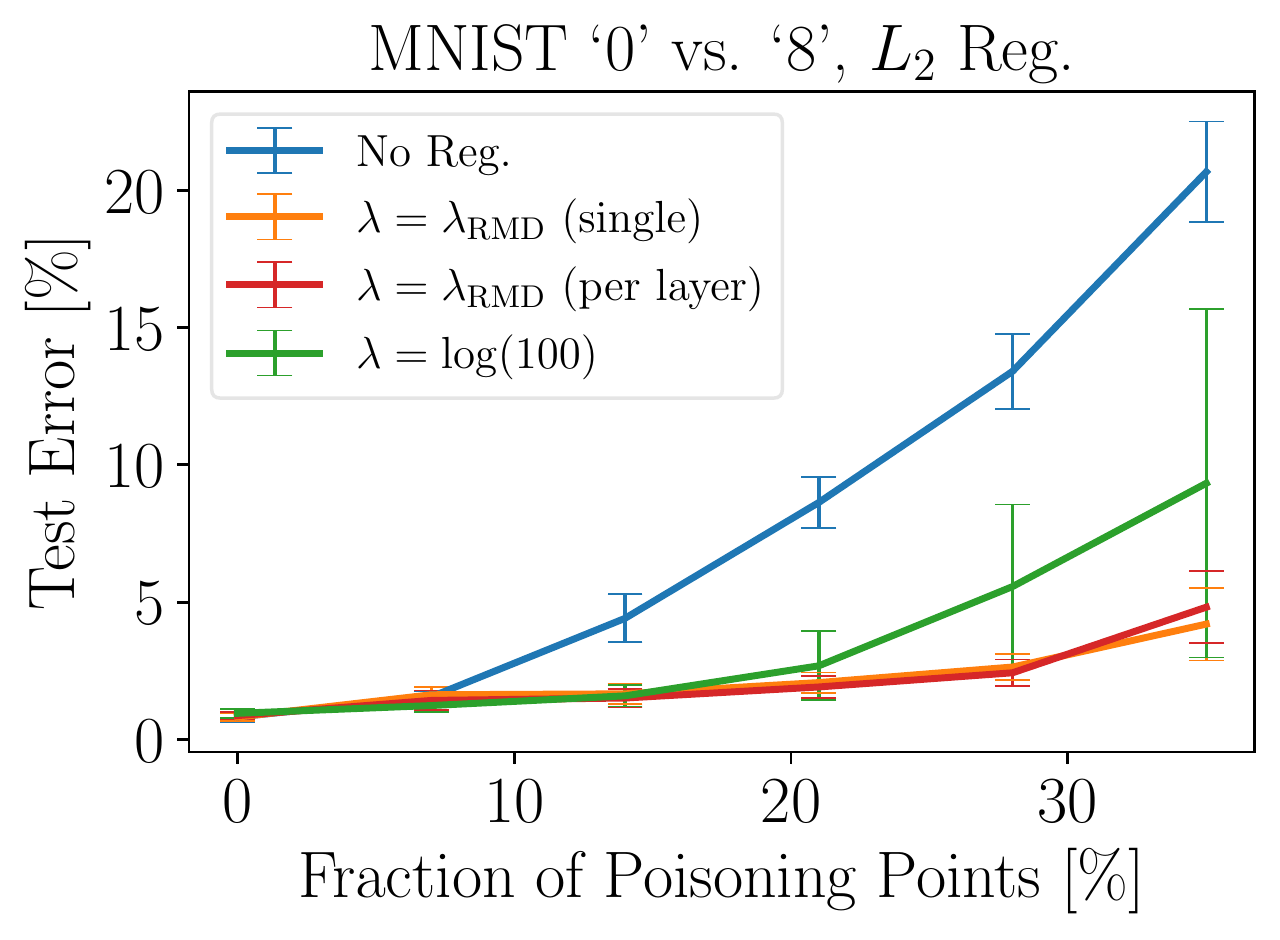}%
\label{fig:dnnopt_a}}
\subfloat[]{\includegraphics[width=1.9in]{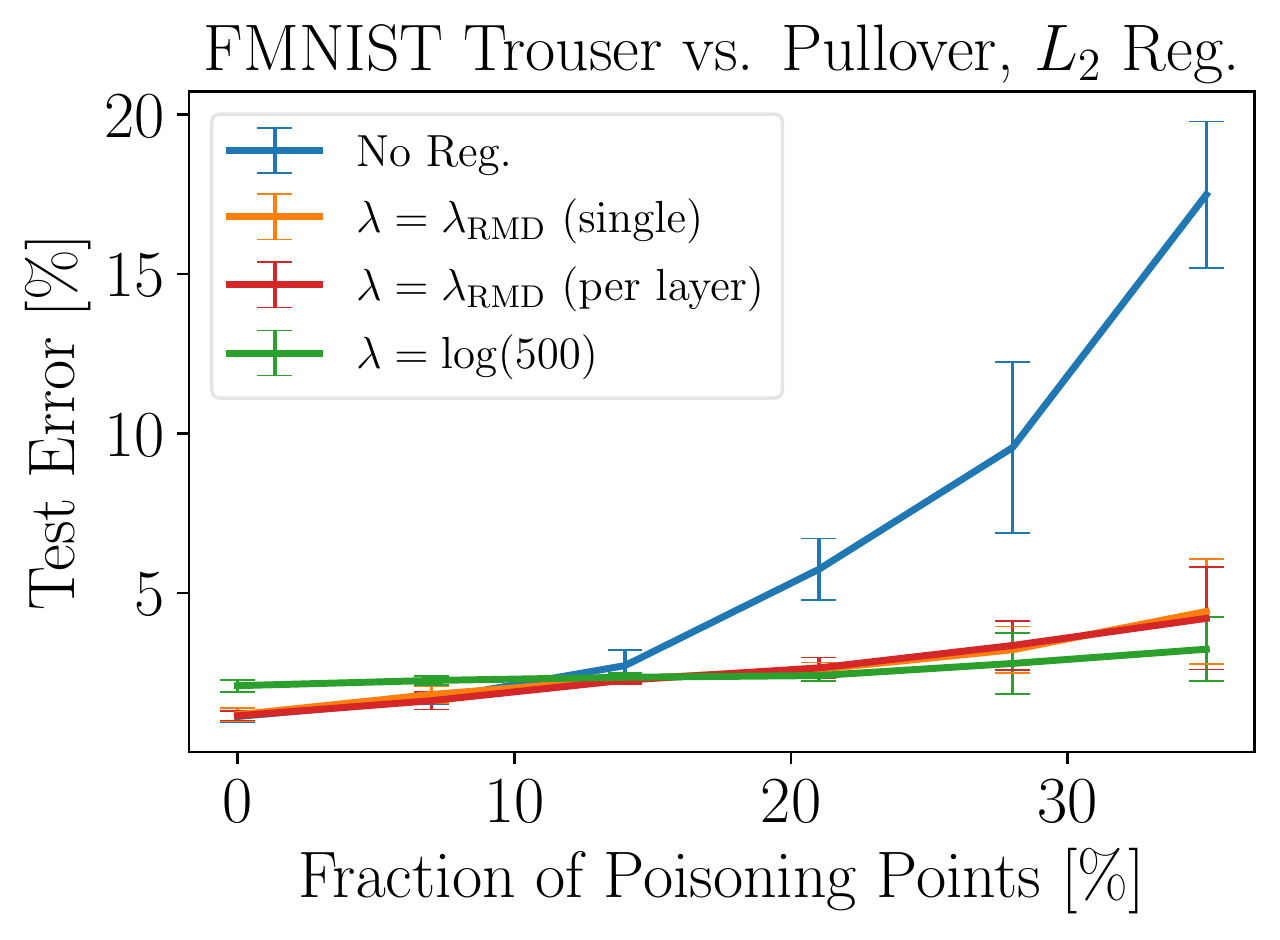}%
\label{fig:dnnopt_b}}
\subfloat[]{\includegraphics[width=1.9in]{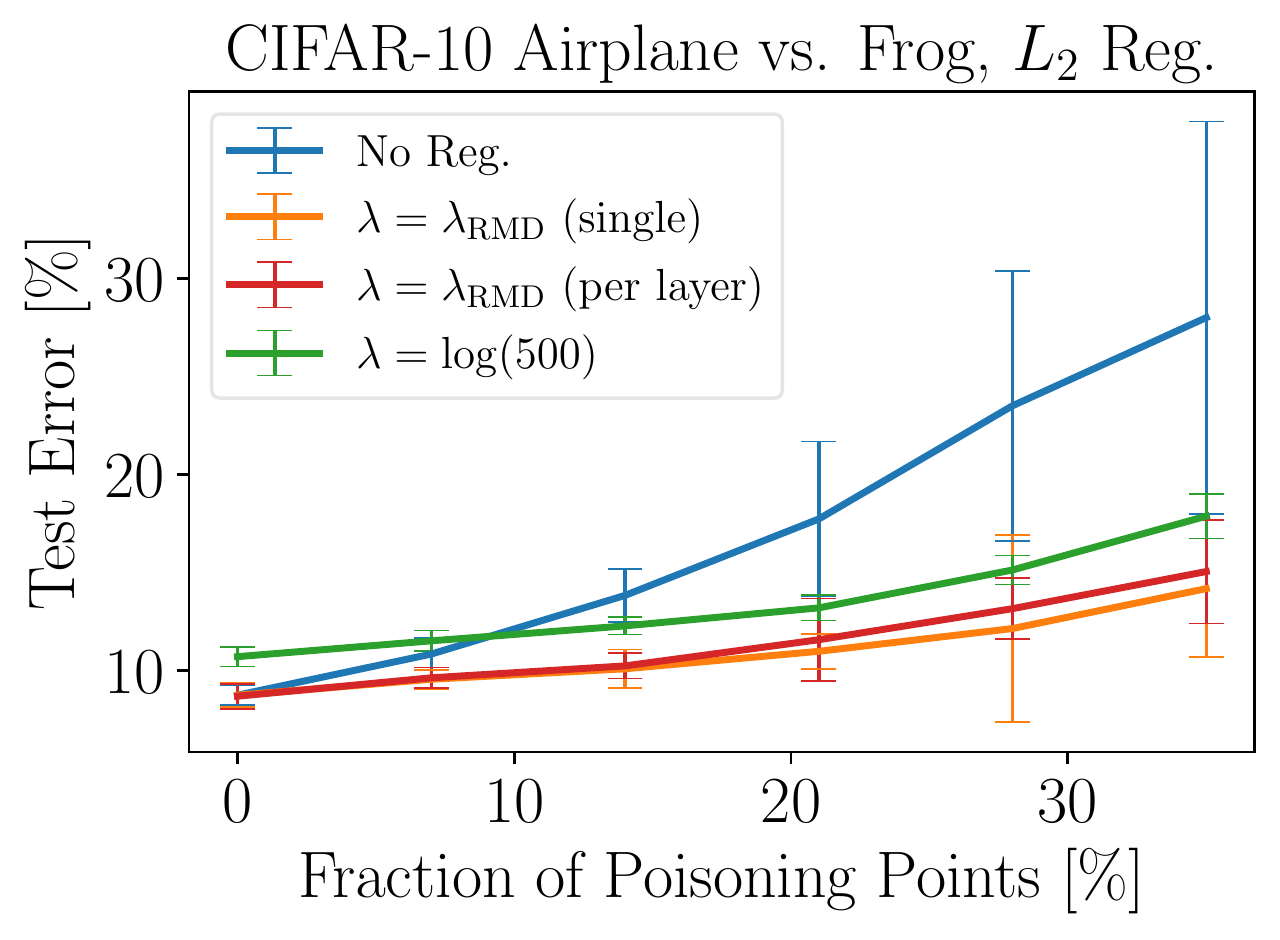}%
\label{fig:dnnopt_c}}
\caption{Average test error for the optimal attack against the DNNs using $L_2$ regularization on (a) MNIST, (b) FMNIST, and (c) CIFAR-10.}
\label{fig:dnnopt_l2}
\end{figure*}

\begin{figure*}[!t]
\centering
\subfloat[]{\includegraphics[width=1.9in]{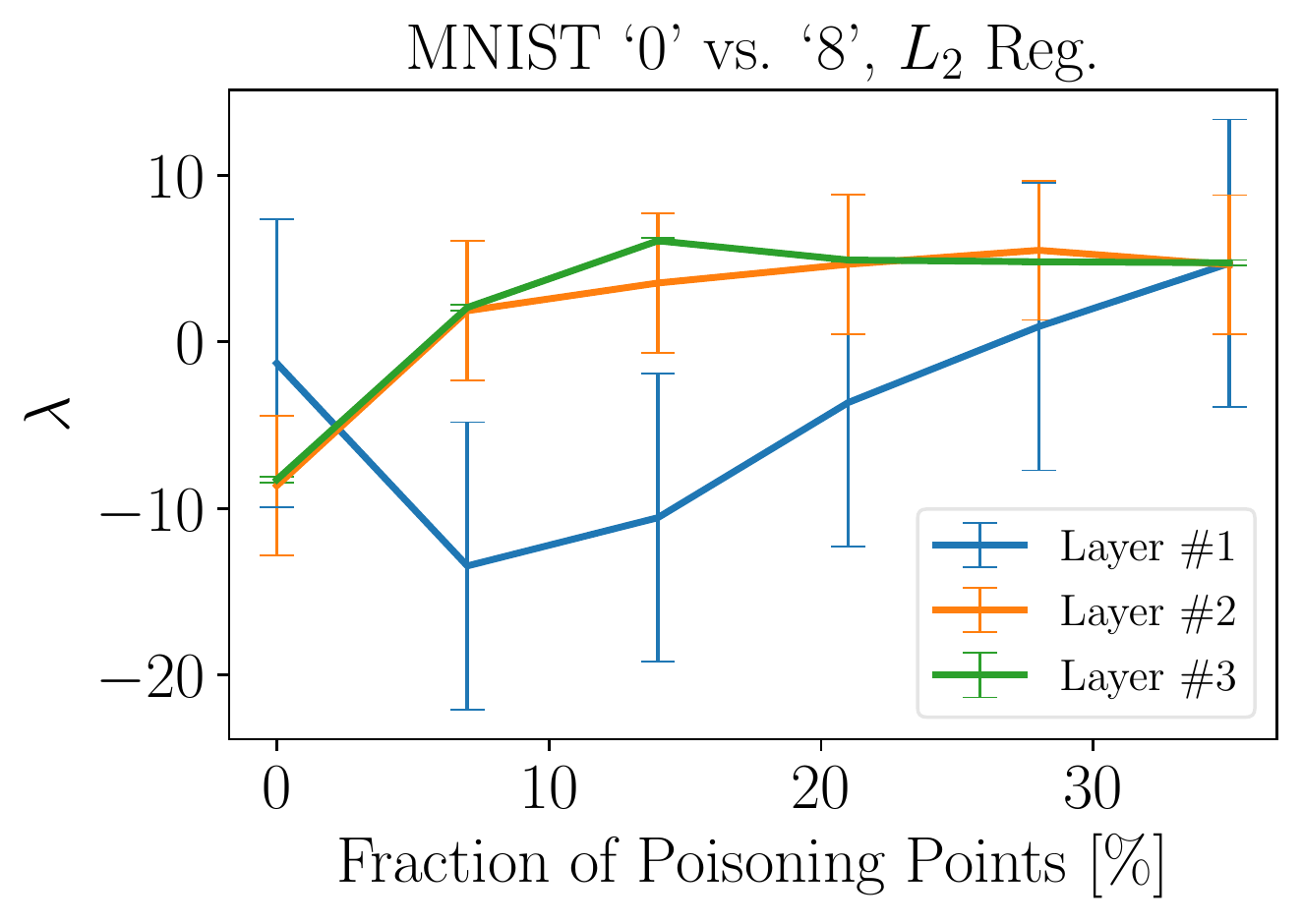}%
\label{fig:dnnlambd_pl_a}}
\subfloat[]{\includegraphics[width=1.9in]{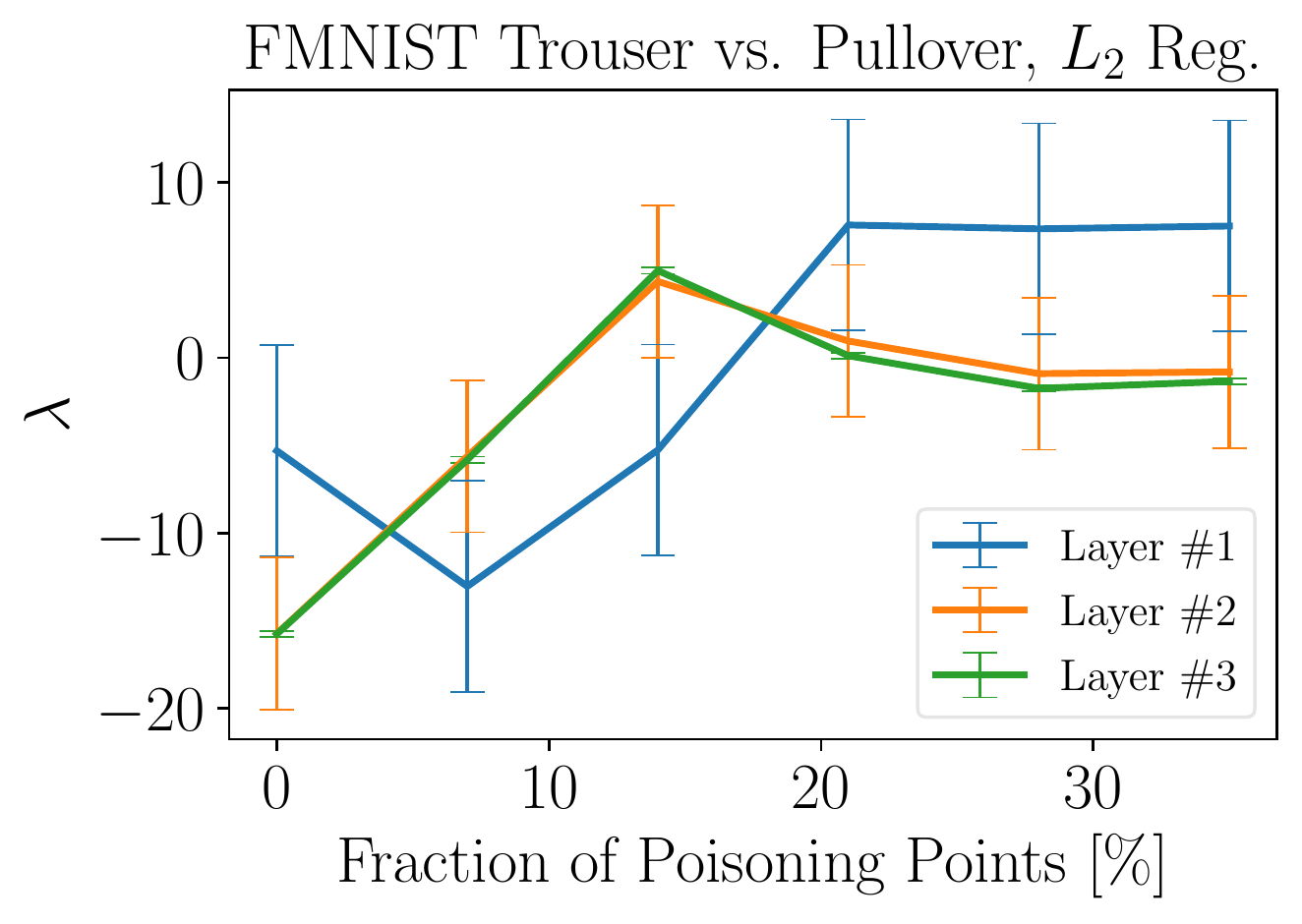}%
\label{fig:dnnlambd_pl_b}}
\subfloat[]{\includegraphics[width=1.9in]{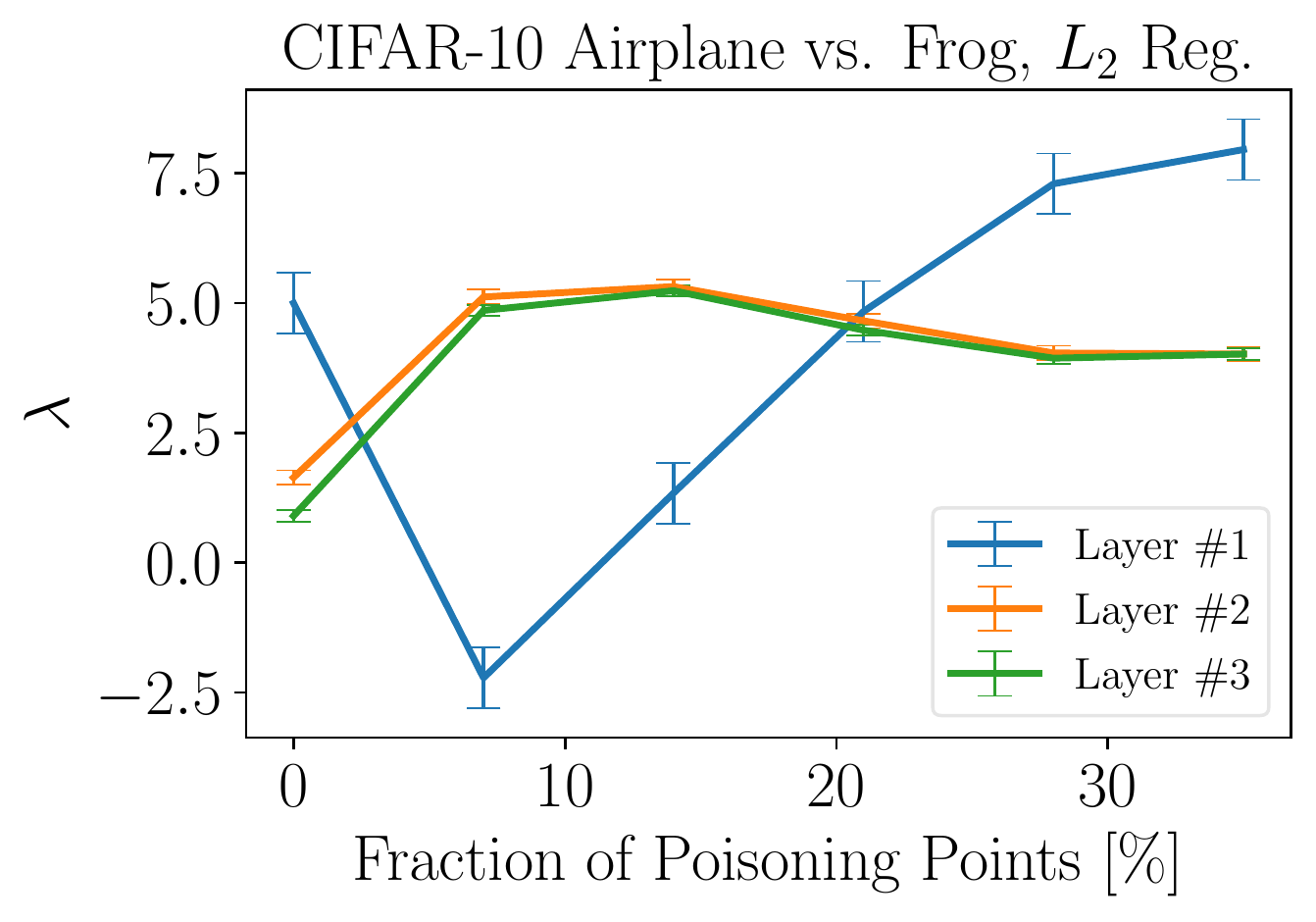}%
\label{fig:dnnlambd_pl_c}}
\caption{Average $\lambda$ learned with RMD at each layer of the DNNs, using $L_2$ regularization.}
\label{fig:dnnlambd_pl_l2}
\end{figure*}

Poisoning attacks can have different effects on the different layers of the target DNNs \cite{chen2018detecting}. This problem has not been sufficiently studied in the research literature and, in this section, we provide useful insights that shed some light in this regard through the lens of regularization. For this, we consider two possibilities: a single regularization hyperparameter, and a vector of regularization hyperparameters---with one hyperparameter for each layer. Intuitively, the amount of scaling needed by each layer's parameters to compensate for a change in the output is not the same, as the activation functions are non-linear. This also gives us an intuition about the layers most vulnerable to the poisoning attack. We also propose an additional modification to the RMD algorithm: we apply different initial random parameters ${\bf w}^{(0)}$ for every update of the poisoning points. This can be interpreted as assembling different randomly initialized DNNs to improve the generalization of the poisoning points across different parameter initializations. We set $T=700$ for MNIST and $T=800$ for FMNIST and CIFAR-10. This scenario is much more challenging for the bilevel problem we aim to solve, as the models have two hidden layers with Leaky ReLU activation functions: $784\times32\times8\times1$, i.e., $25,393$ parameters, for MNIST and FMNIST; and $3,072\times64\times32\times1$, i.e., $198,785$ parameters, for CIFAR-10.

As before, we denote with $\lambda_{\text{RMD}}$ the case where the regularization hyperparameter is learned according to Eq.~(\ref{eqAttacker2}), distinguishing now the cases: (1) when a single regularization hyperparameter is used for the whole DNN, and (2) when a different hyperparameter is used at each layer. We also performed attacks with different strength for the DNN assuming it is trained without regularization ($\lambda=-\infty$) and with a large value for $\lambda$ (for $L_2$ regularization: $\lambda= \log(100)$ for MNIST, and $\lambda= \log(500)$ for FMNIST and CIFAR-10; for $L_1$ regularization: $\lambda= \log(50)$ for MNIST, $\lambda= \log(10)$ for FMNIST, and $\lambda= \log(25)$ for CIFAR-10), constant for all the layers. Fig.~\ref{fig:dnnopt_l2} shows the results for $L_2$ regularization.  The results for $L_1$ regularization are coherent with the ones for $L_2$ and can be found in Fig.~\ref{fig:dnnopt_l1}. In this case, we omitted the case where $\lambda$ is set with $5$-fold cross-validation on the clean dataset as the search space is large, which makes it computationally very expensive.

The results in Fig.~\ref{fig:dnnopt_l2} are consistent with those obtained for the case of LR (Fig.~\ref{fig:lropt}). When there is no regularization, the algorithm is vulnerable to the poisoning attack and its test error increases significantly. For a large value of $\lambda$, the algorithm's performance remains quite stable, but the clean error is higher. For $\lambda_{\text{RMD}}$ the test error increases only moderately, and the results when using a single hyperparameter or a different hyperparameter at each layer are very similar. From Fig.~\ref{fig:dnnopt_l2} and {Fig.~\ref{fig:dnnopt_l1}} we can see that when there is no attack, the test error for $\lambda_{\text{RMD}}$ is smaller than in the other two cases. Although over-regularizing may be appealing to make the algorithm more robust to poisoning, the performance in the absence of attacks may be significantly worse. Learning $\lambda$ evidences this trade-off. For a large fraction of poisoning points, the small discrepancy observed between $\lambda_{\text{RMD}}$ and the large value of $\lambda$ is due to the non-convexity of the bilevel optimization problem, resulting in learning (possibly) suboptimal values for $\lambda_{\text{RMD}}$. On the other hand, comparing the results for the DNNs (Fig.~\ref{fig:dnnopt_l2} and {Fig.~\ref{fig:dnnopt_l1}}) and for LR (Fig.~\ref{fig:lropt}), it is evident that the mitigating effect of regularization is more prominent in the case of DNNs. As the capacity of the DNN (compared to LR) is higher, the attackers can have more flexibility to manipulate the decision boundary. Hence, having regularization in place, in combination with the trusted validation set, is even more important in the case of the DNNs.

Fig.~\ref{fig:dnnlambd_pl_l2} and {Fig.~\ref{fig:dnnlambd_pl_l1}} show the value of $\lambda$ when using a different regularization term at each layer, for $L_2$ and $L_1$ regularization, correspondingly. We observe that the $\lambda$ learned for the second and output layers increases faster than the one for the first layer and, for FMNIST and CIFAR-10, this increase is faster for the first hidden layer from $20\%$ of poisoning. This suggests that the latter layers can be more vulnerable to the poisoning attack. The poisoning attack tries to produce more changes in those layers and, at the same time, the network tries to resist those changes by increasing the value of the corresponding regularization hyperparameters. On the other hand, when the attack is very strong, the impact of the attack appears more uniform across all layers in the DNN, based on the values of $\lambda$ learned for each layer.  

Finally, as in the case of LR, the value of the regularization hyperparameters is also related to the norm of the weights divided by the number of parameters for each layer in the DNN. These results are shown in {Fig.~\ref{fig:dnnlambd_pl_weights_l2}}.

\section{Conclusions}
\label{sec:conclusion}

Existing literature has been ambivalent on the role of regularization in mitigating poisoning attacks. This problem has been insufficiently studied as existing works assume that regularization hyperparameters are constant and chosen ``a priori'' regardless of the number of poisoning points or their effects. We have shown that the value of the hyperparameters depends on the amount of poisoning and that a constant value cannot be chosen a priori: when the value is too low, it provides insufficient robustness; when the value is too high, it damages performance. We have shown that when the value of the hyperparameters is learned as a function of the poisoning incurred, regularization can significantly mitigate the effect of indiscriminate poisoning attacks, whilst at the same time not damaging performance. This, however, requires the use of a small trusted validation set. 

To study the mitigating effect of regularization and choose hyperparameters, we have introduced an novel formulation where the poisoning attack strategy for worst case scenarios is formulated as a \textit{multiobjective bilevel optimization problem}. This formulation allows to learn the most appropriate values for the model's hyperparameters and to calculate the poisoning points simultaneously. Solving this multiobjective bilevel optimisation problem is challenging. However, we have shown how this problem can be solved with gradient-based techniques by extending previous RMD-based approaches. 

With this formulation, we have analyzed the effect of indiscriminate poisoning attacks against LR and DNN classifiers when using both $L_2$ and $L_1$ regularization. Our results confirm that the use of regularization, combined with the presence of the small trusted set to learn the hyperparameters, significantly helps to reduce the error under poisoning attacks. When the regularization hyperparameter is learned appropriately, the algorithm is more robust and, at the same time, the performance of the model is not affected when there is no attack. The trusted validation set required is quite small and task dependent; a larger trusted set is not necessarily advantageous. 

Although $L_2$ regularization typically provides more stability compared to $L_1$, our empirical results show that both types of regularization are useful to reduce the effect of poisoning attacks. Additionally, our results show that the use of regularization plays a more important role in more complex models, such as DNNs. Our empirical evaluation also shows that indiscriminate attacks have a more pronounced effect in the later layers of the network, as the value of the regularization hyperparameters learned for those layers increases significantly (with respect to those learned when there is no attack) compared to the ones learned for the first layers. However, for a large fraction of poisoning points, the effect of the attack is spread across all the different layers.  

In our future work, we plan to investigate these aspects in targeted poisoning attacks and ways to combine and contrast the mitigating effect obtained from regularization with that of other defenses against poisoning attacks, e.g. data sanitization.


%

\appendices

\section*{Acknowledgment}

We gratefully acknowledge funding for this work from the Defence Science and Technology Laboratory (Dstl), under the project ERASE - Evaluating the Robustness of Machine Learning Algorithms in Adversarial Settings.




\bibliographystyle{IEEEtran}
\bibliography{IEEEfull}

\section{Experimental Settings for the Synthetic Example}

\label{sec:expset}

\begin{figure*}[!t]
\centering
\subfloat[]{\includegraphics[width=1.8in]{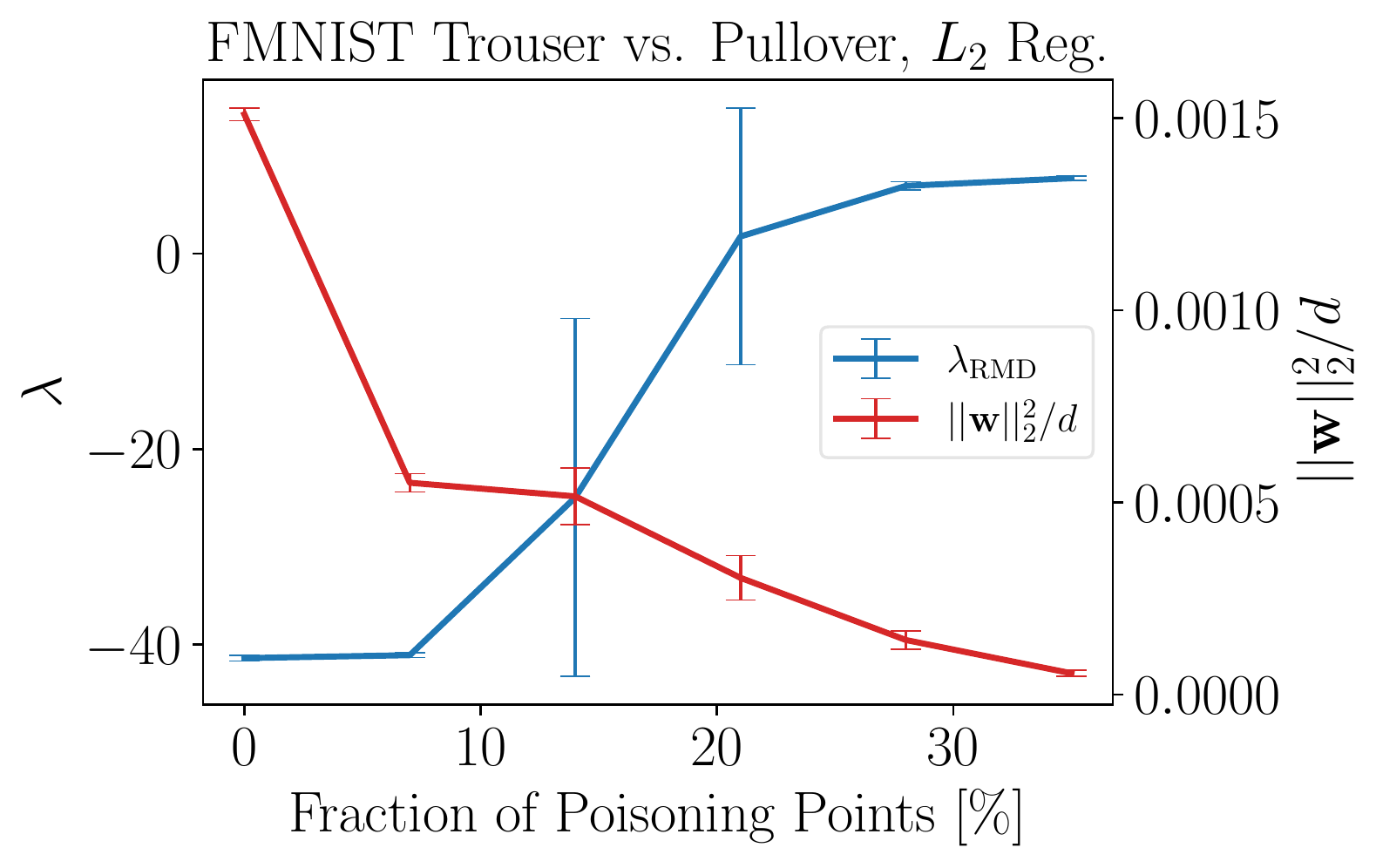}%
\label{fig:lrlambd_b}}
\subfloat[]{\includegraphics[width=1.8in]{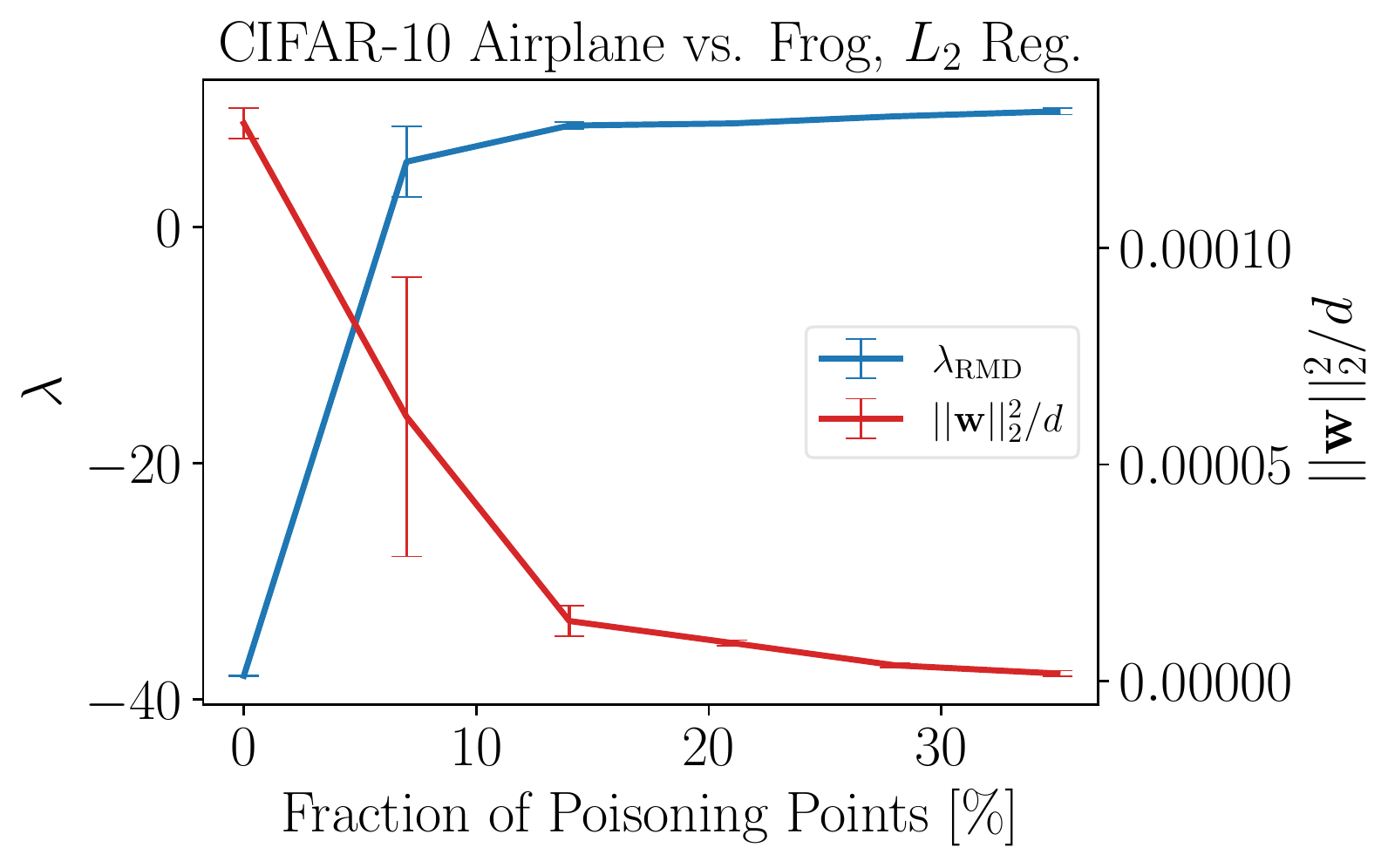}%
\label{fig:lrlambd_c}}
\subfloat[]{\includegraphics[width=1.8in]{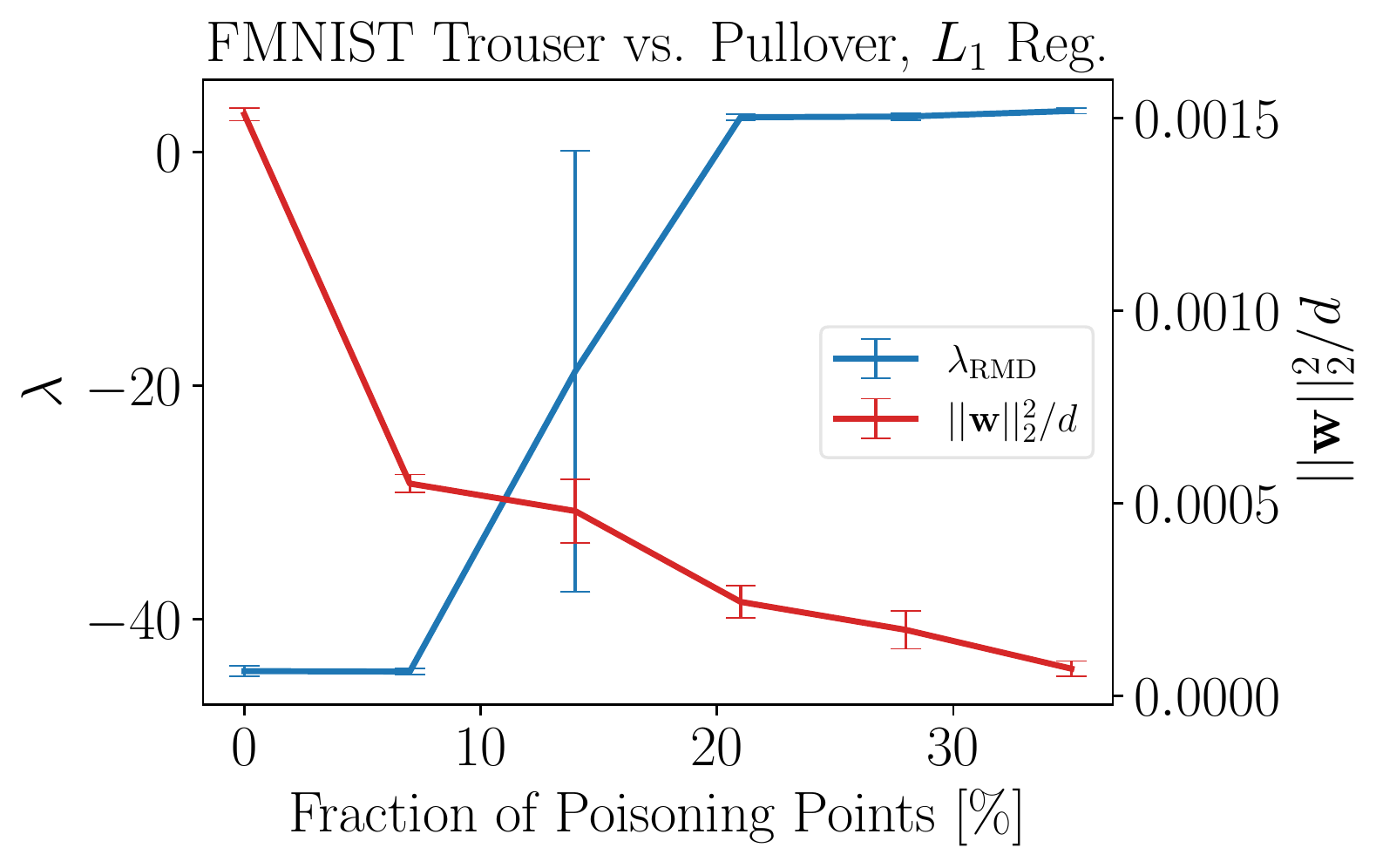}%
\label{fig:lrlambd_e}}
\subfloat[]{\includegraphics[width=1.8in]{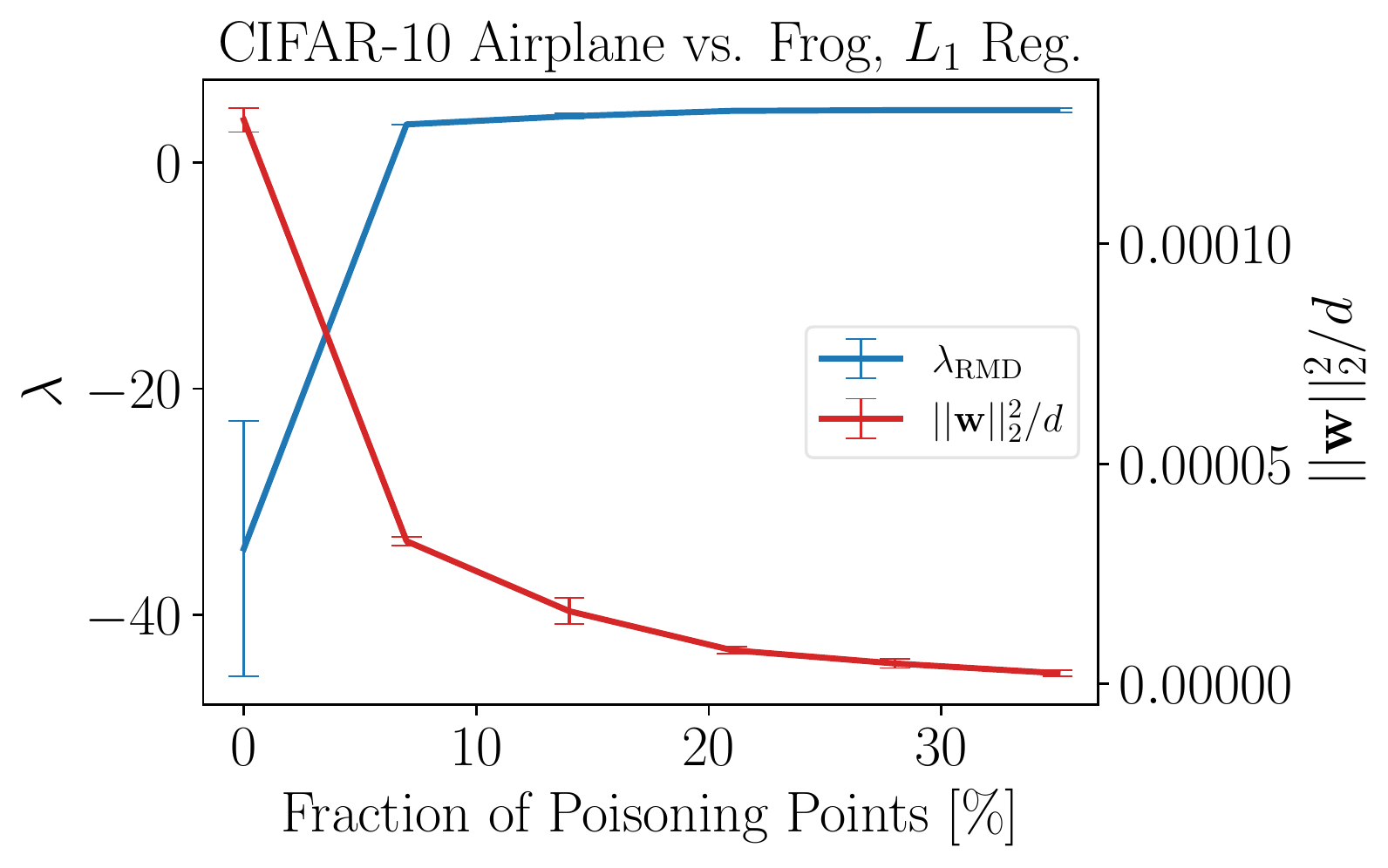}%
\label{fig:lrlambd_f}}
\caption{Average $\lambda$ and { average} $||{\bf w}||_2^2$ for the optimal attack against LR: (a) $L_2$ regularization on FMNIST, (b) $L_2$ regularization on CIFAR-10, (c) $L_1$ regularization on FMNIST, and (d) $L_1$ regularization on CIFAR-10.}
\label{fig:lrlambd_2}
\end{figure*}

\begin{figure*}[!t]
\centering
\subfloat[]{\includegraphics[width=1.8in]{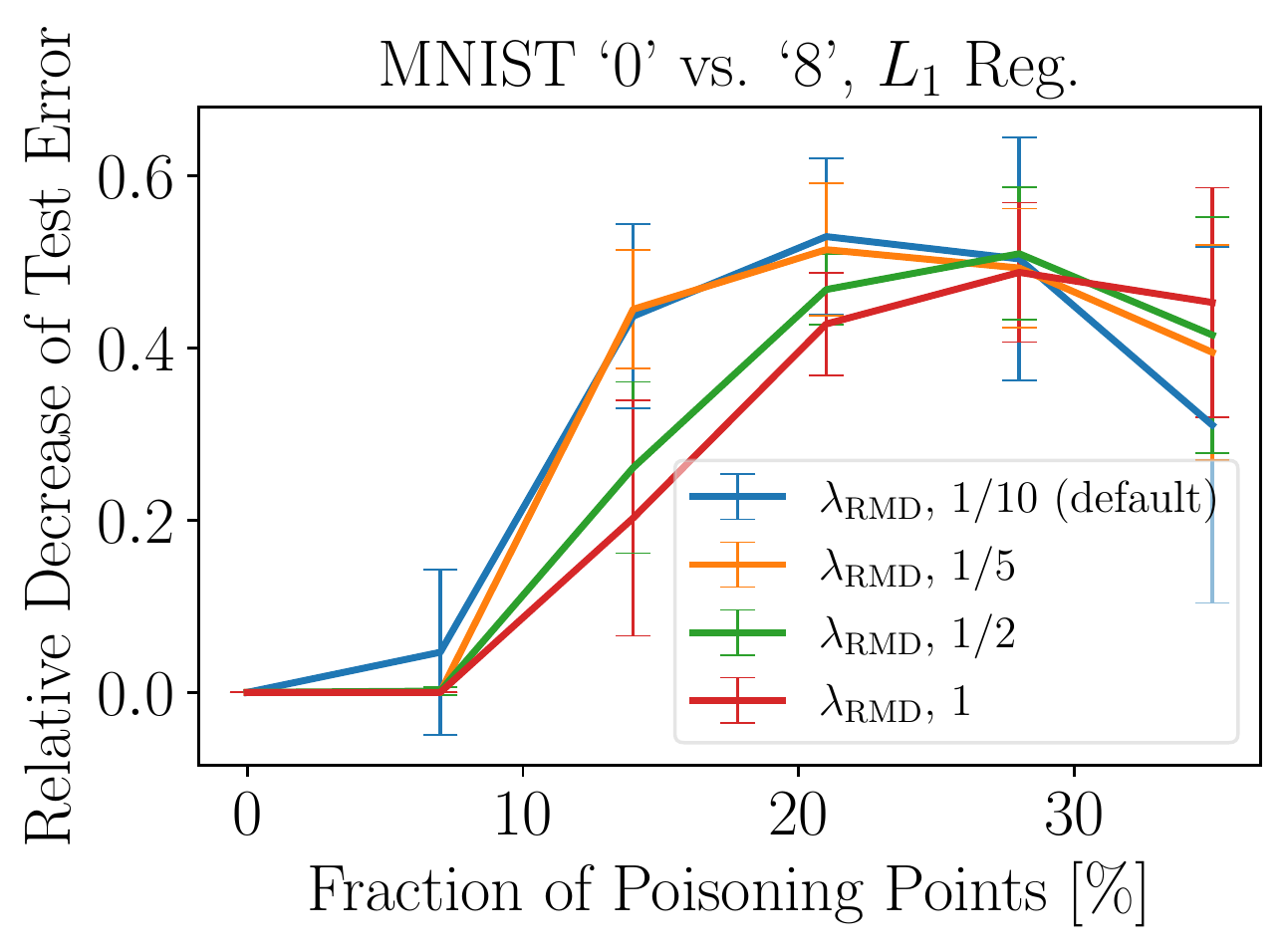}%
\label{fig:lr_val_c}}
\subfloat[]{\includegraphics[width=1.8in]{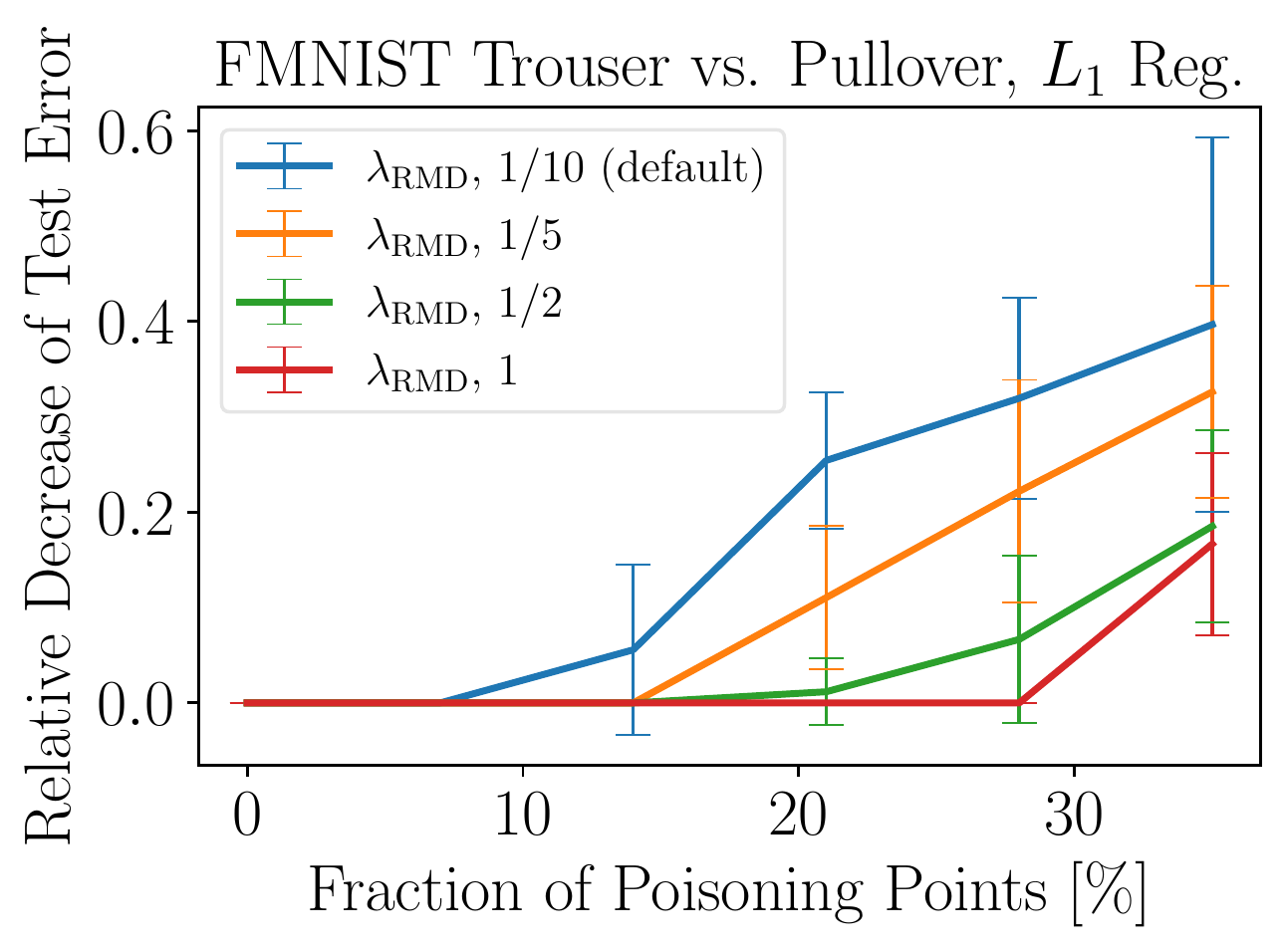}%
\label{fig:lr_val_f}}
\subfloat[]{\includegraphics[width=1.8in]{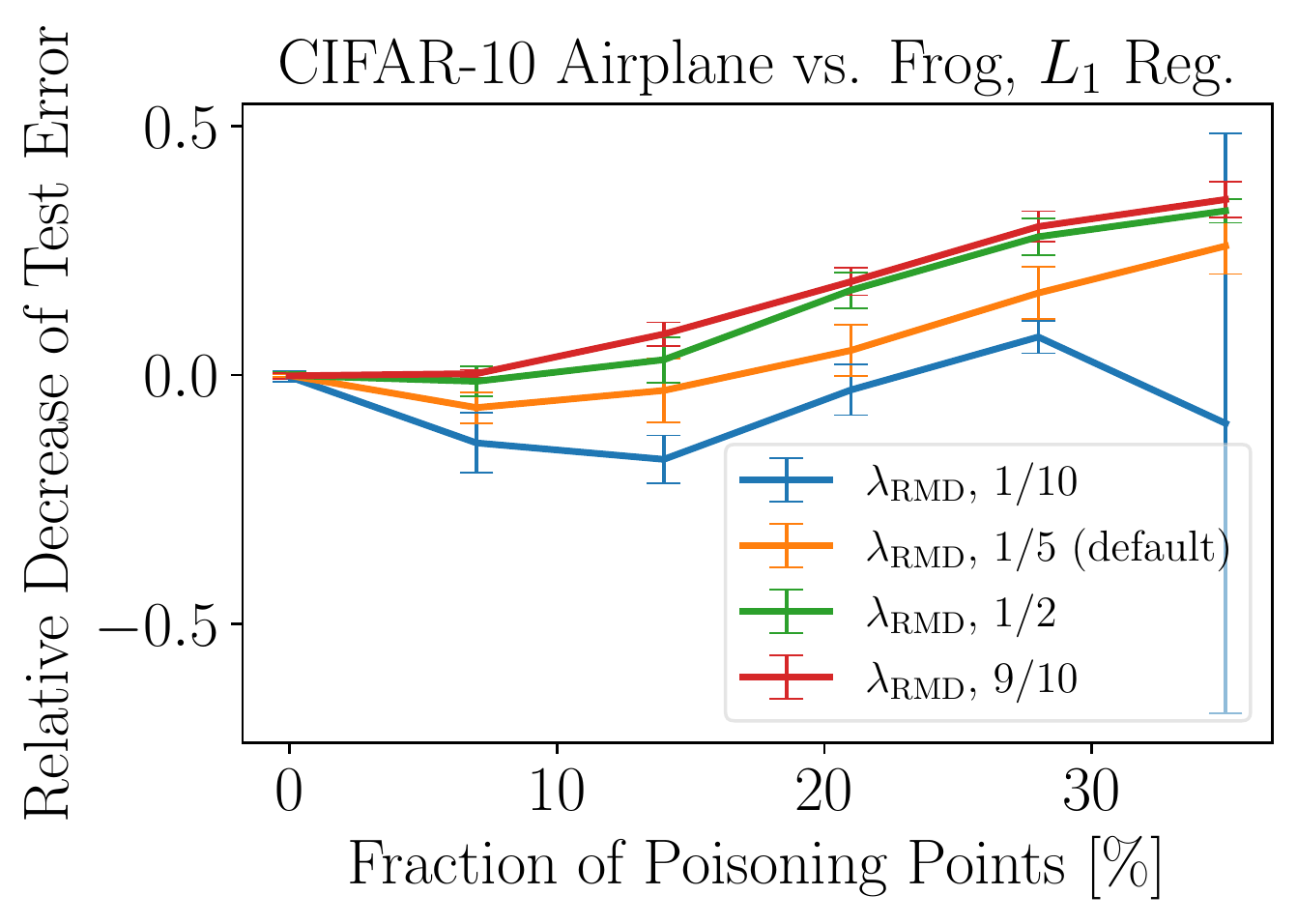}%
\label{fig:lr_val_i}}
\caption{Sensitivity analysis of the size of the validation set for $L_1$ regularization: (a) MNIST, (b) FMNIST, and (c) and CIFAR-10.}
\label{fig:lr_val_2}
\end{figure*}

For the synthetic example in {Fig.~\ref{fig:synthetic}, we sample the attacker's data from two bivariate Gaussian distributions, $\mathcal{N}(\boldsymbol{\mu}_0, \boldsymbol{\Sigma}_0)$ and $\mathcal{N}(\boldsymbol{\mu}_1, \boldsymbol{\Sigma}_1)$ with parameters:

\begin{equation*}
  \begin{aligned}
    \boldsymbol{\mu}_0 & = \begin{bmatrix} -3.0 \\ \hspace{\minuslength}0.0 \end{bmatrix}, & \qquad \boldsymbol{\Sigma}_0 & = \begin{bmatrix} 2.5 & 0.0 \\ 0.0 & 1.5 \end{bmatrix}, \\
    \boldsymbol{\mu}_1  & = \begin{bmatrix} 3.0 \\ 0.0 \end{bmatrix}, &         \boldsymbol{\Sigma}_1 & = \begin{bmatrix} 2.5 & 0.0 \\ 0.0 & 1.5 \end{bmatrix}. \\
  \end{aligned}
\end{equation*}

The attacker uses $32$ points ($16$ per class) for training and $64$ ($32$ per class) for validation, and one poisoning point cloned from the validation set (in the example of the paper, cloned from the set labeled as blue), whose label is flipped. This poisoning point is concatenated into the training set and the features of this point are optimized with RMD. In order to poison the LR classifier, we use $\alpha=0.4$ and $T_{\mathcal{D}_\text{p}}=50$;  $\Phi(\mathcal{D}_\text{p})\in[-9.5,9.5]^2$; $\eta=0.2$, ${T=100}$; and when testing the attack, $\eta_\text{tr}=0.2$, batch size $=32$ (full batch), and $\text{number of epochs} = 100$. When we apply regularization, we fix $\lambda=\log(20)\approx3$.

To plot the colormap in {Fig.~\ref{fig:synthetic}(right), }
the values of $\lambda$ explored for each possible poisoning point are in the range $[-8, 6]$. Then, the optimal value of $\lambda$ is chosen such that it minimizes the error of the model, trained on each combination of the poisoning point (concatenated into the training set) and $\lambda$ in the grid, and evaluated on the validation set.

\section{Additional Results }

\label{sec:addres}

\begin{figure*}[!t]
\centering
\subfloat[]{\includegraphics[width=1.9in]{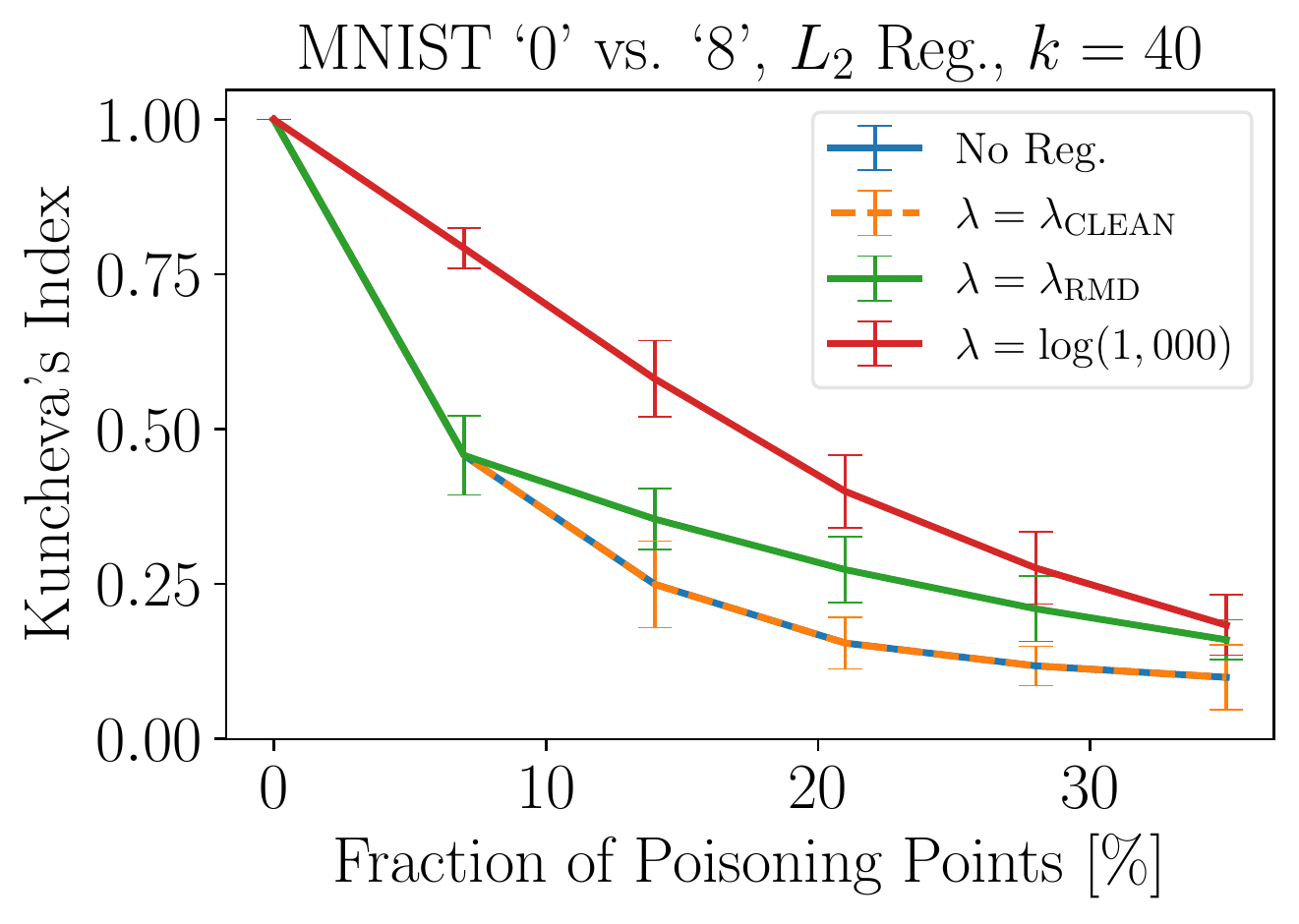}%
\label{fig:kunchevamnist_b}}
\subfloat[]{\includegraphics[width=2in]{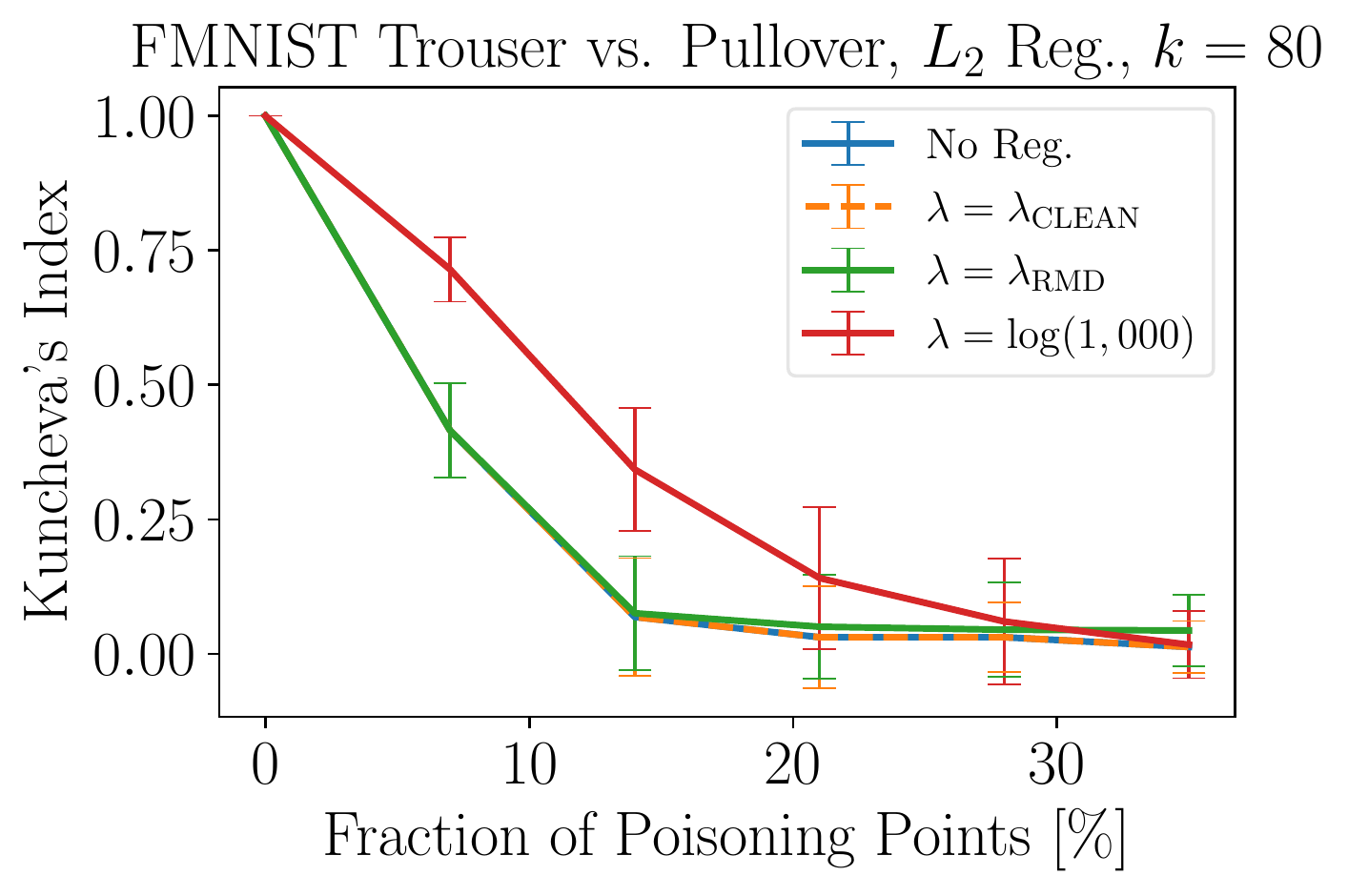}%
\label{fig:kunchevafmnist_b}}
\subfloat[]{\includegraphics[width=2in]{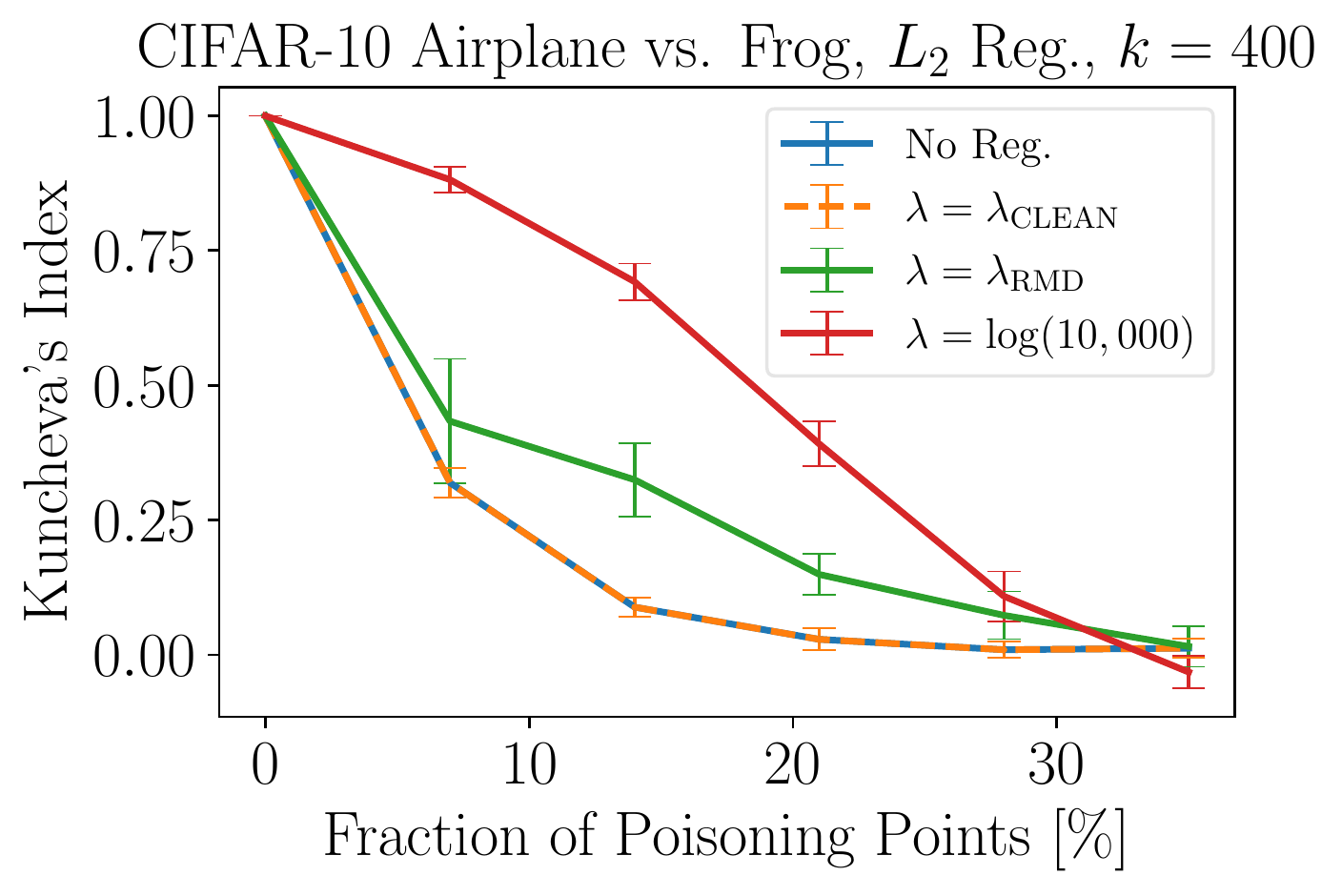}%
\label{fig:kunchevacifar_b}}
\\
\vspace{-.3cm}
\subfloat[]{\includegraphics[width=1.8in]{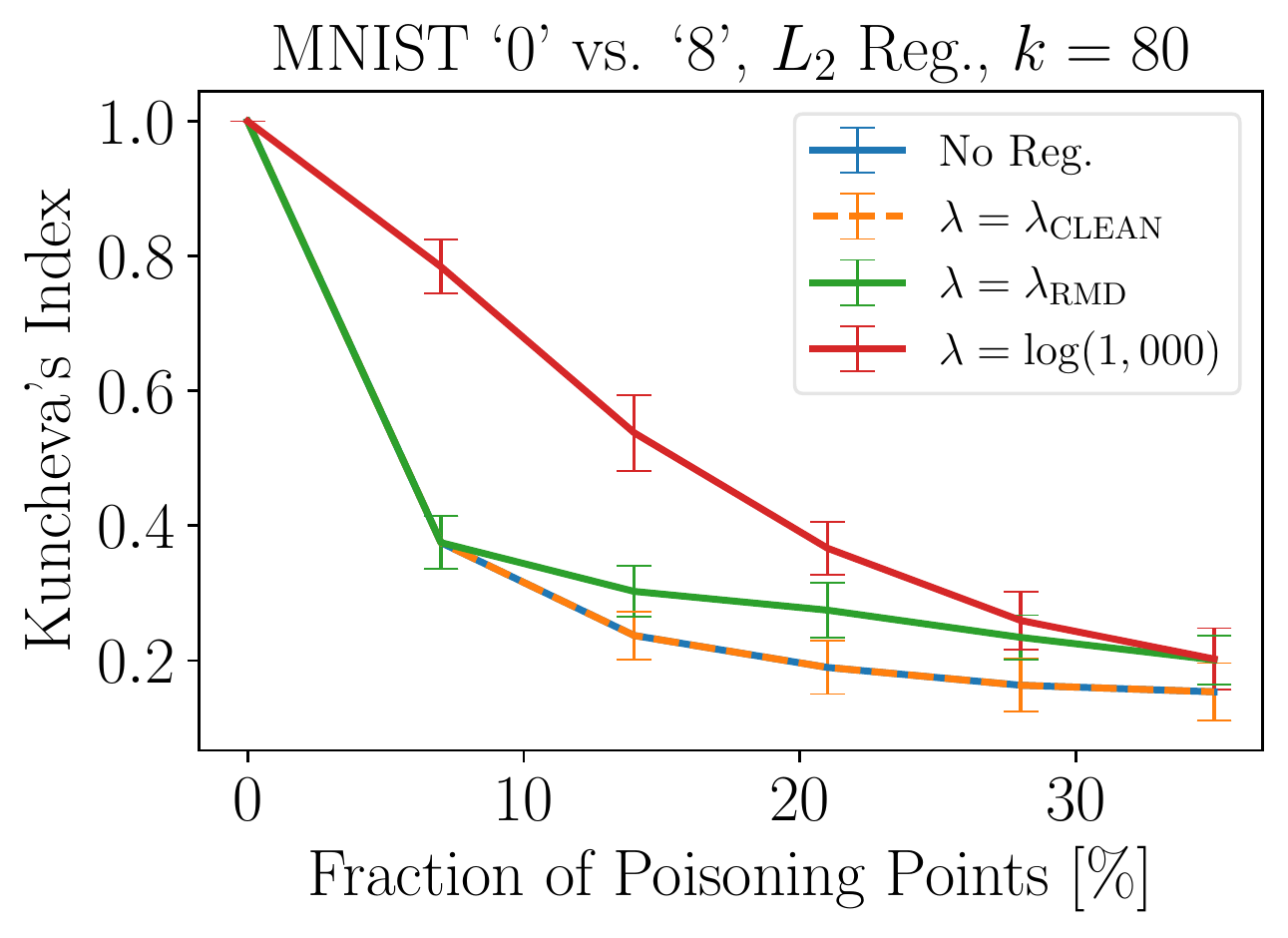}%
\label{fig:kunchevamnist_c}}
\subfloat[]{\includegraphics[width=2in]{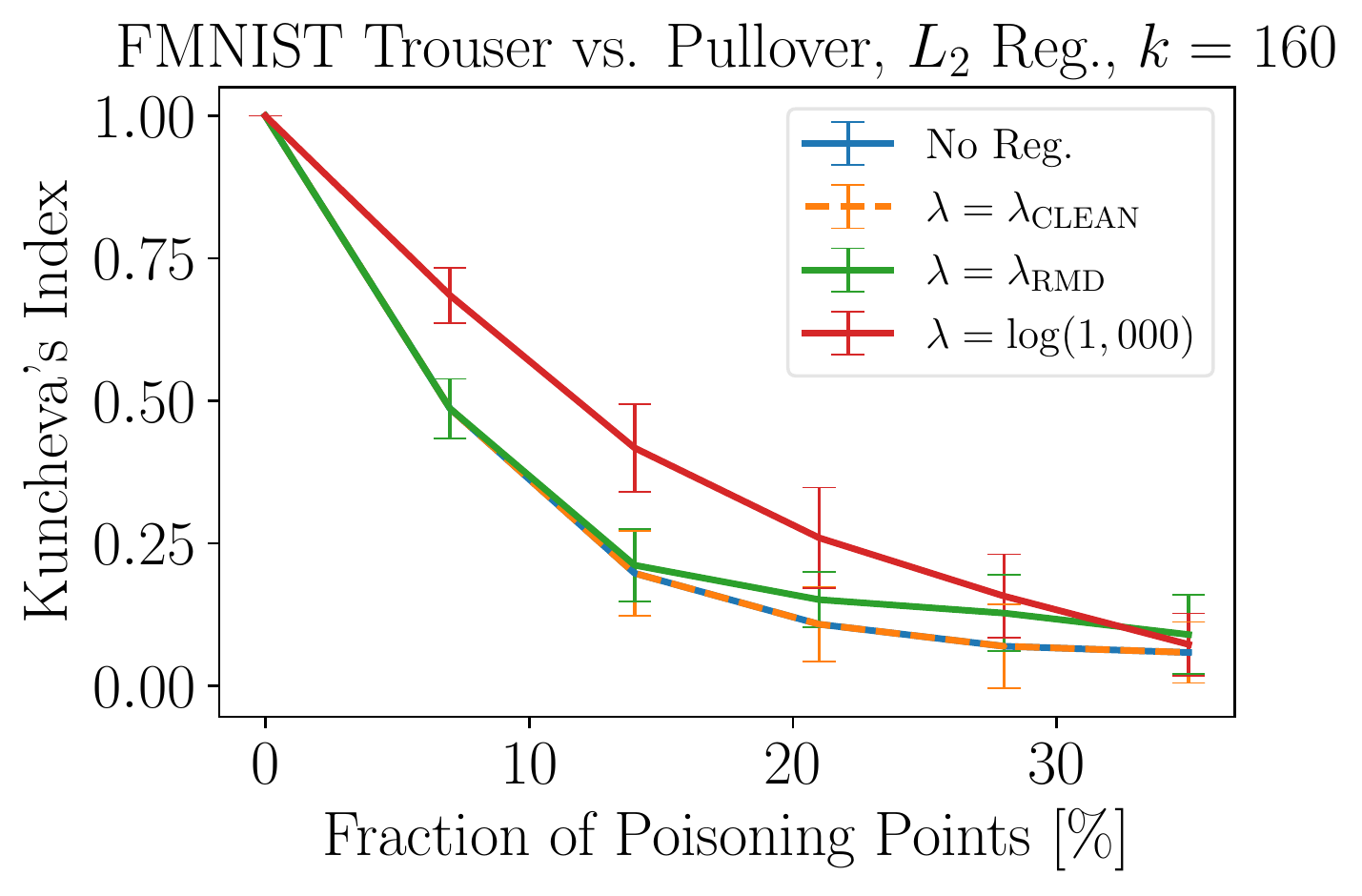}%
\label{fig:kunchevafmnist_c}}
\subfloat[]{\includegraphics[width=2in]{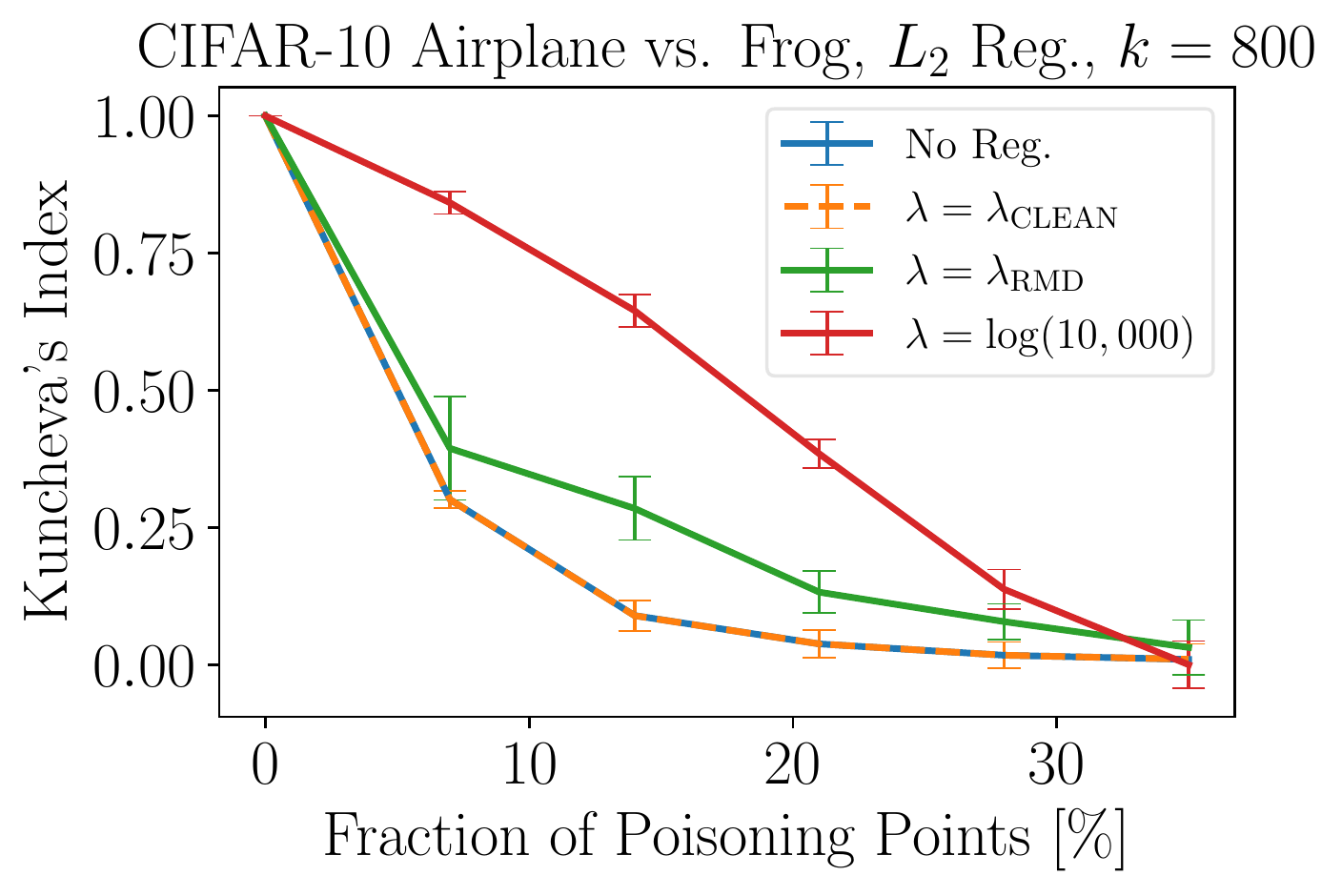}%
\label{fig:kunchevacifar_c}}
\\
\vspace{-.3cm}
\subfloat[]{\includegraphics[width=1.9in]{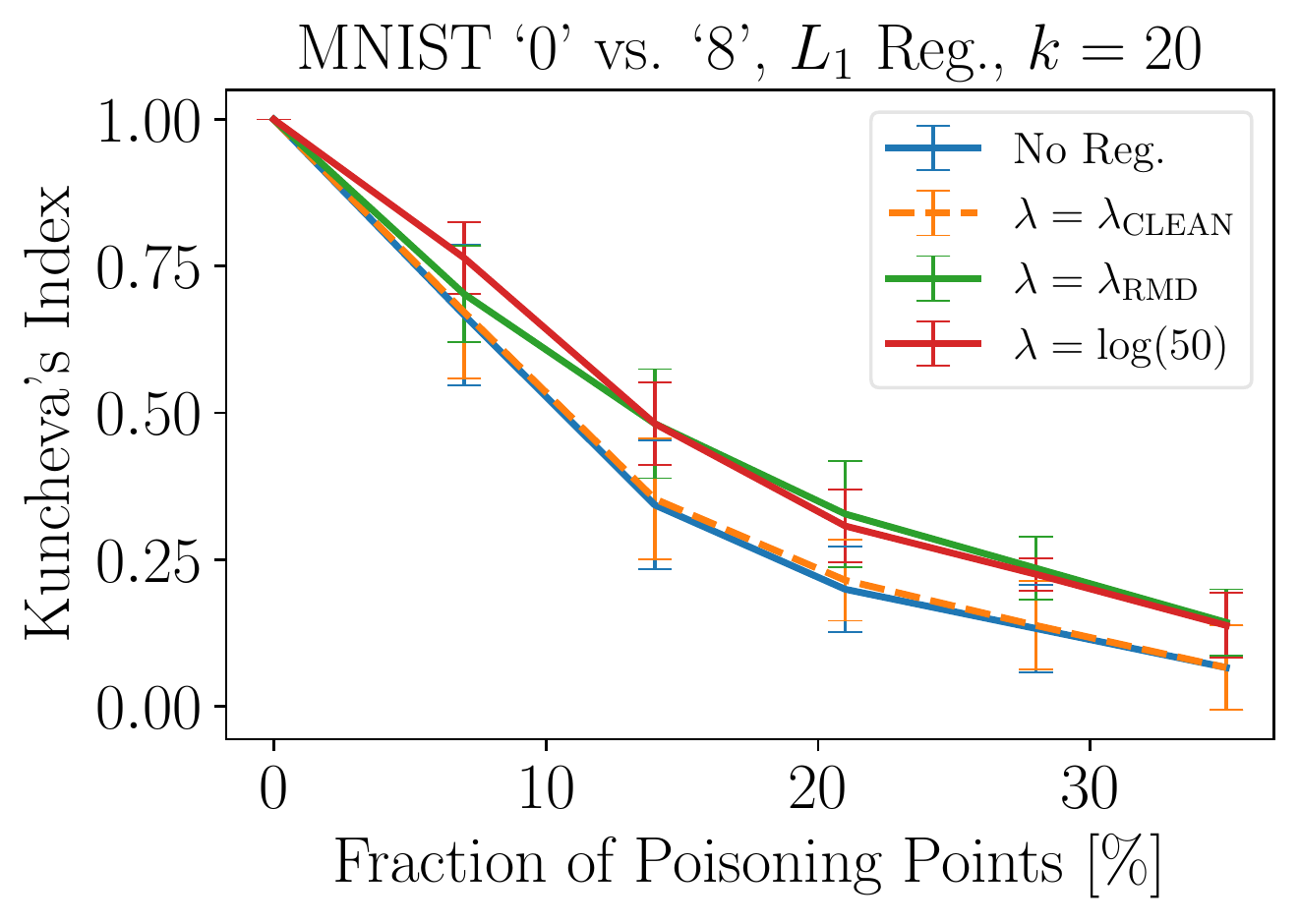}%
\label{fig:kunchevamnist_l1_a}}
\subfloat[]{\includegraphics[width=2in]{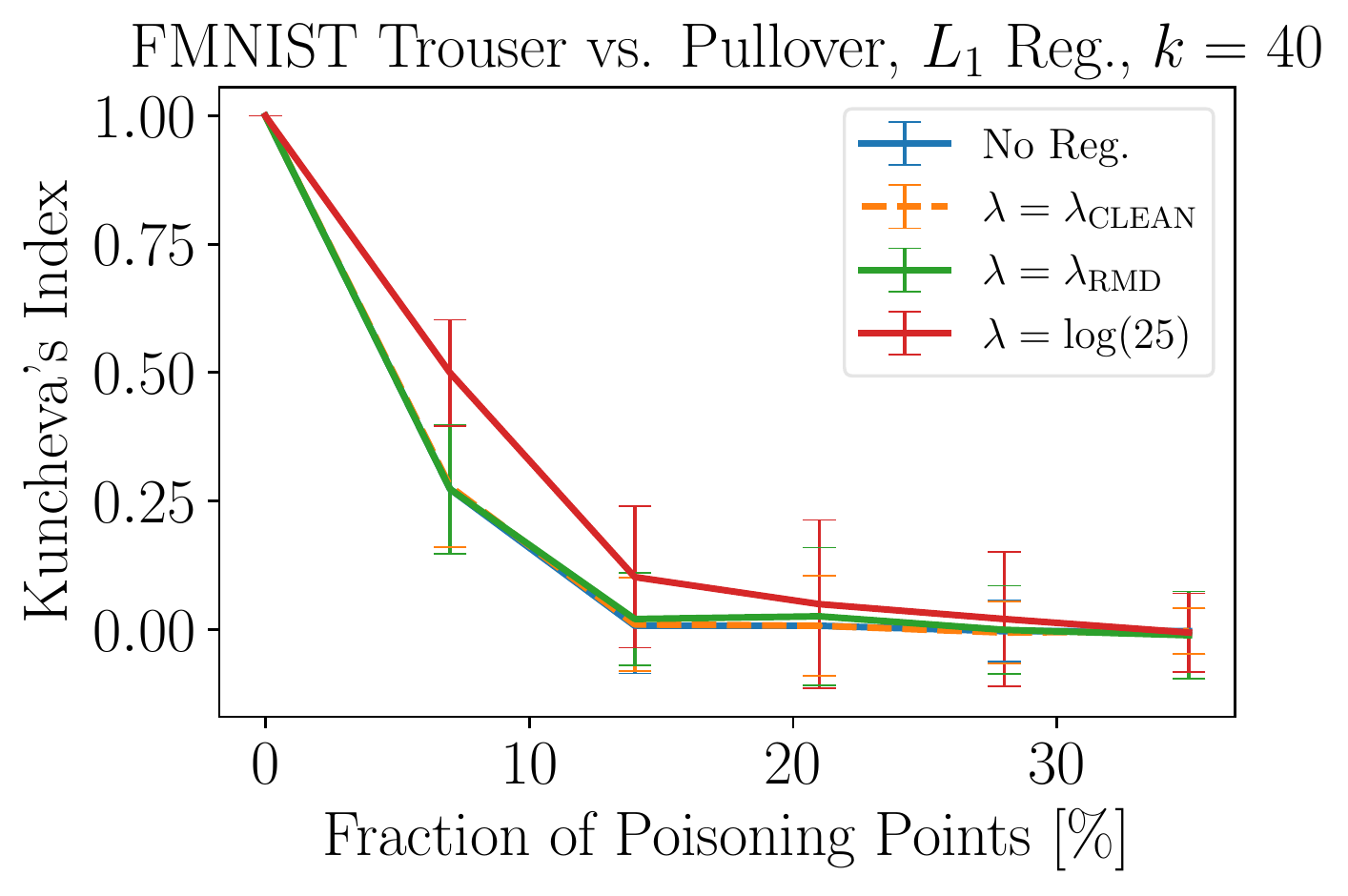}%
\label{fig:kunchevafmnist_l1_a}}
\subfloat[]{\includegraphics[width=2in]{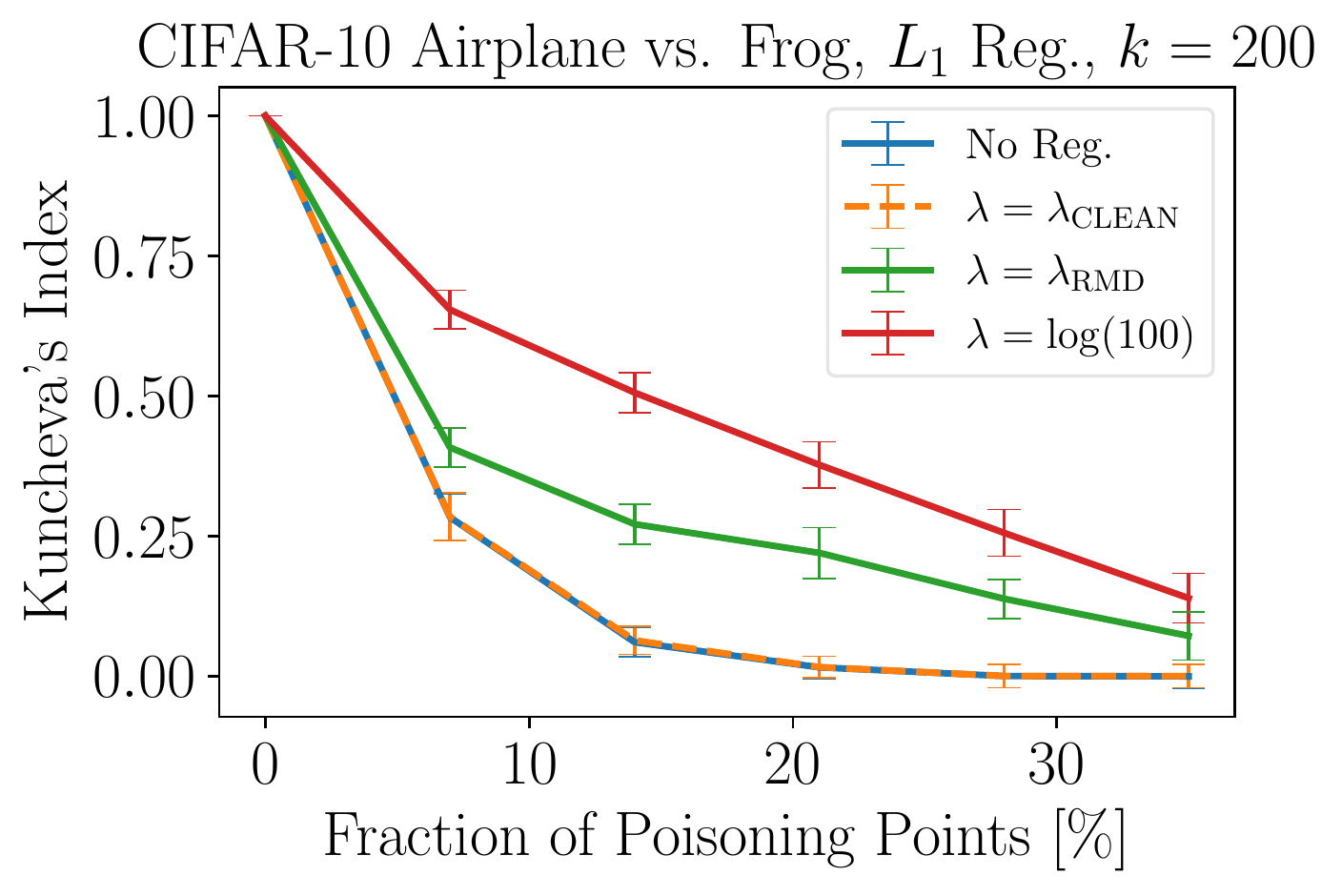}%
\label{fig:kunchevacifar_l1_a}}
\\
\vspace{-.3cm}
\subfloat[]{\includegraphics[width=2in]{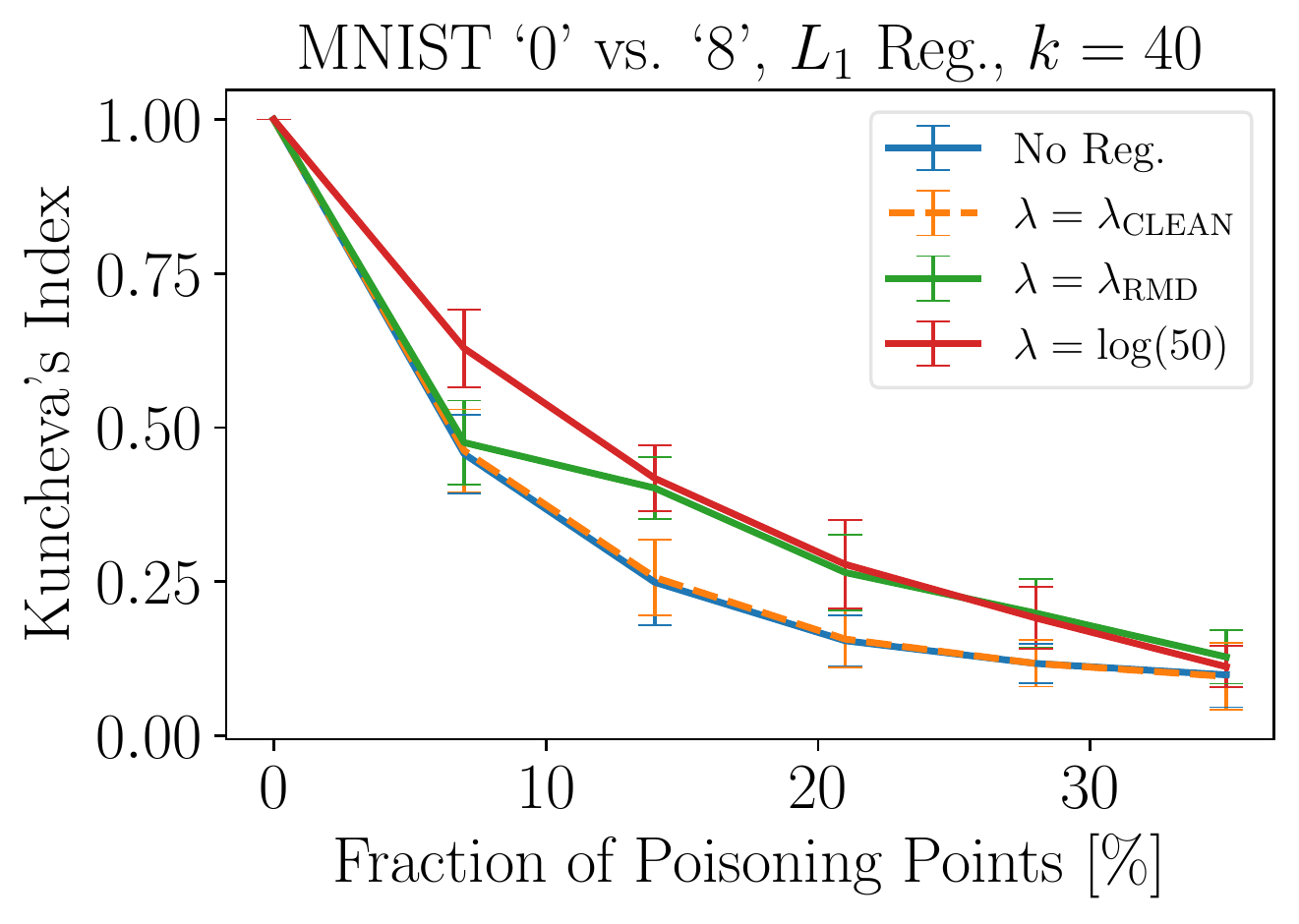}%
\label{fig:kunchevamnist_l1_b}}
\subfloat[]{\includegraphics[width=2in]{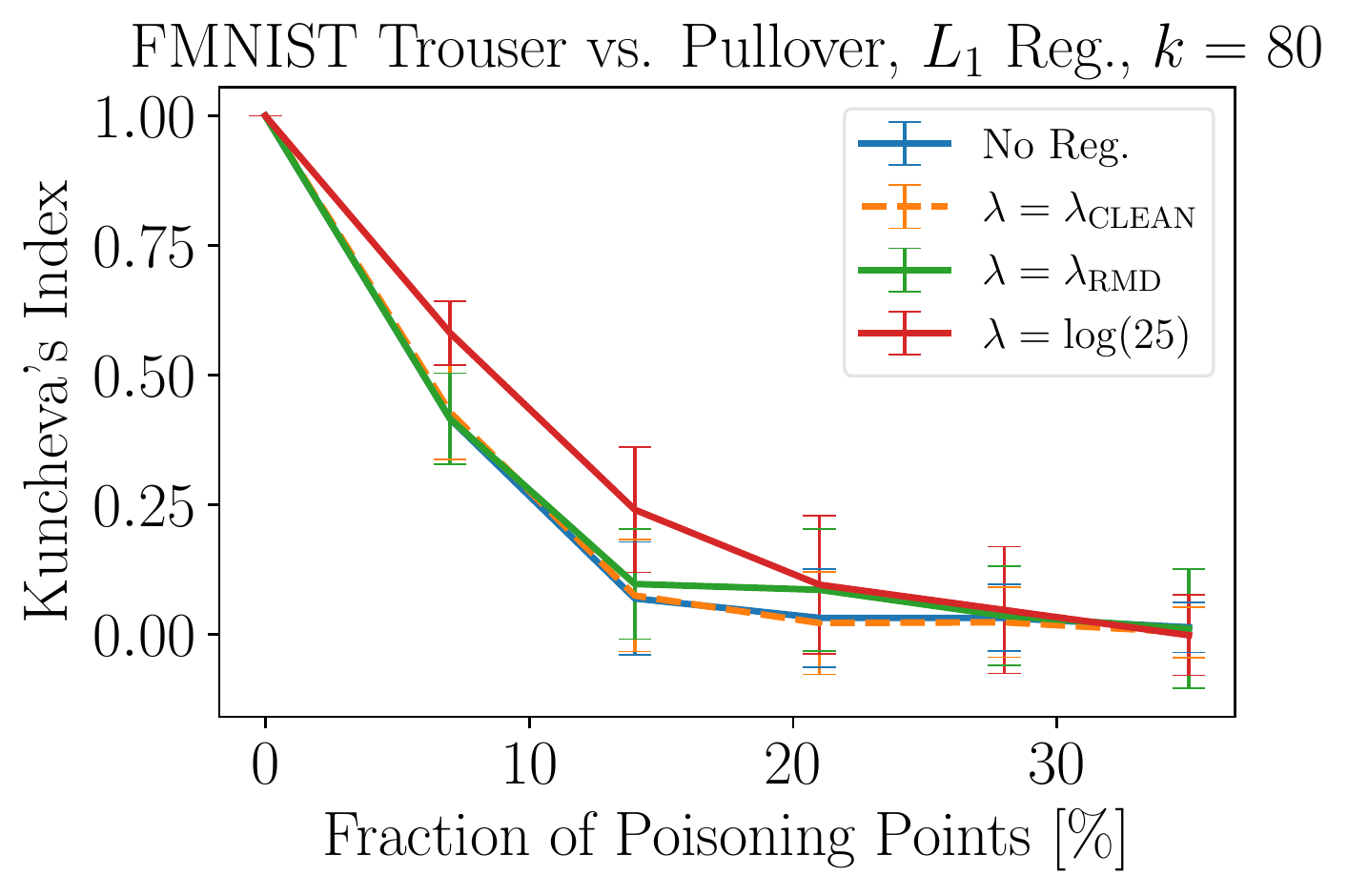}%
\label{fig:kunchevafmnist_l1_b}}
\subfloat[]{\includegraphics[width=2in]{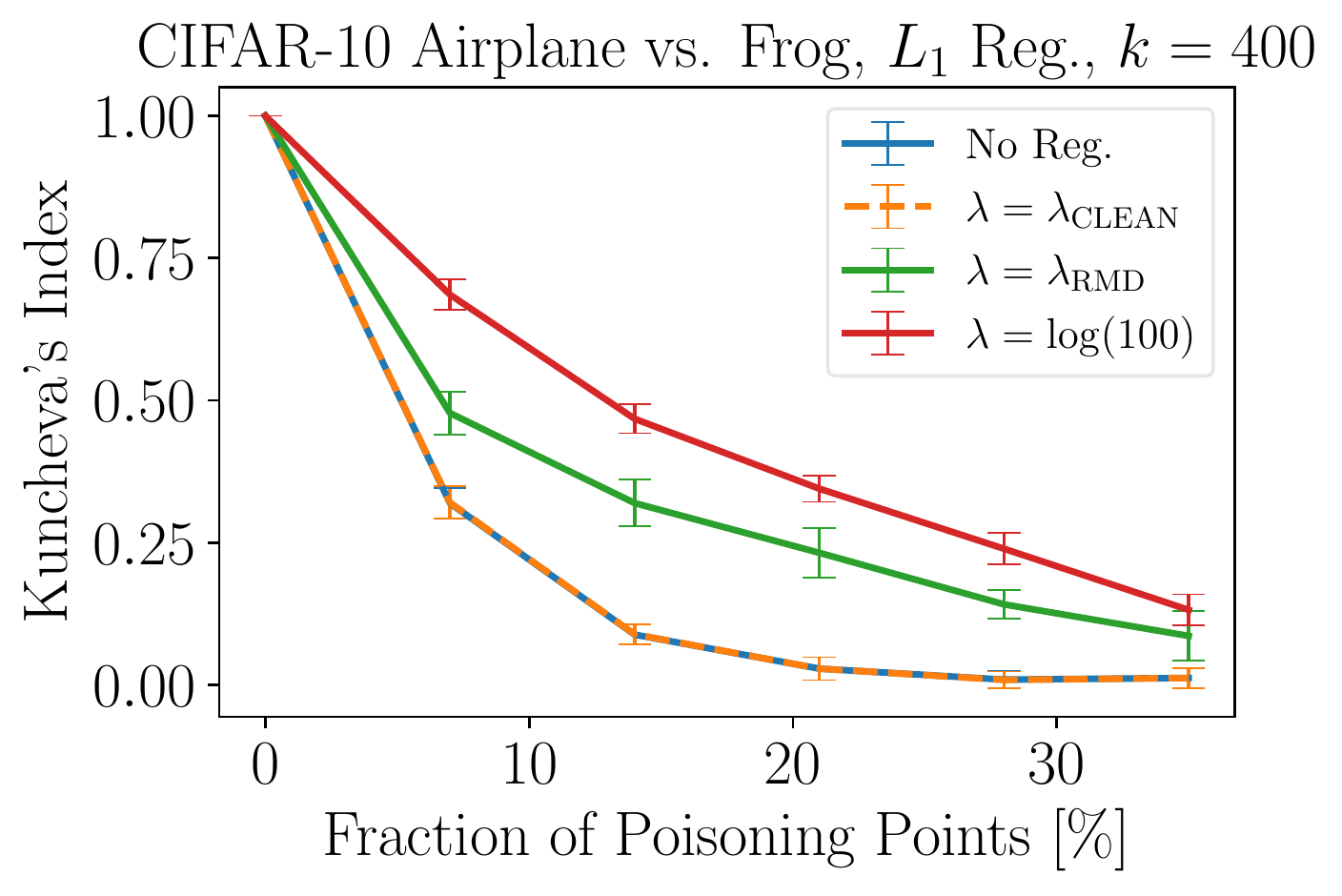}%
\label{fig:kunchevacifar_l1_b}}
\\
\vspace{-.3cm}
\subfloat[]{\includegraphics[width=1.8in]{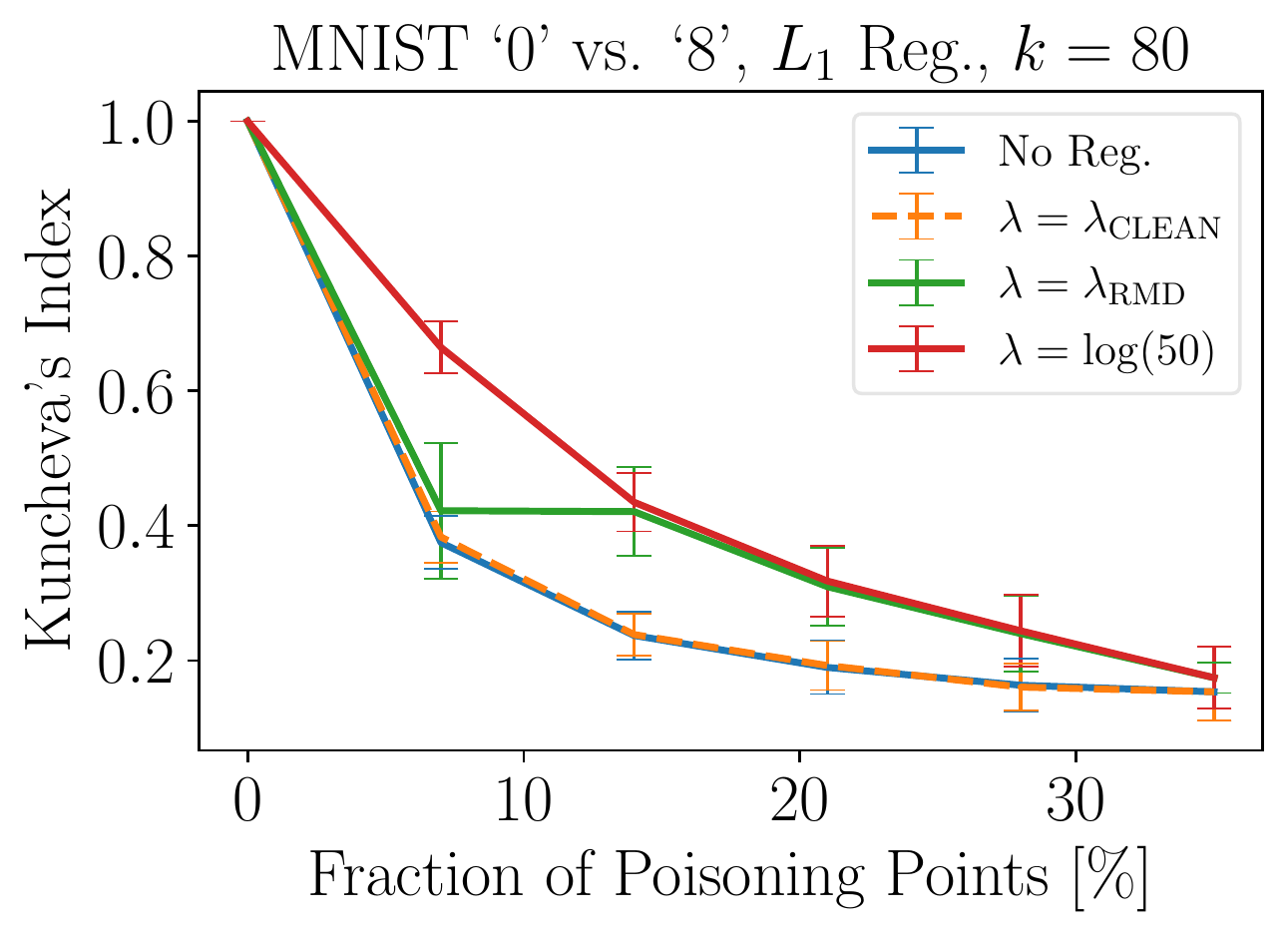}%
\label{fig:kunchevamnist_l1_c}}
\subfloat[]{\includegraphics[width=2in]{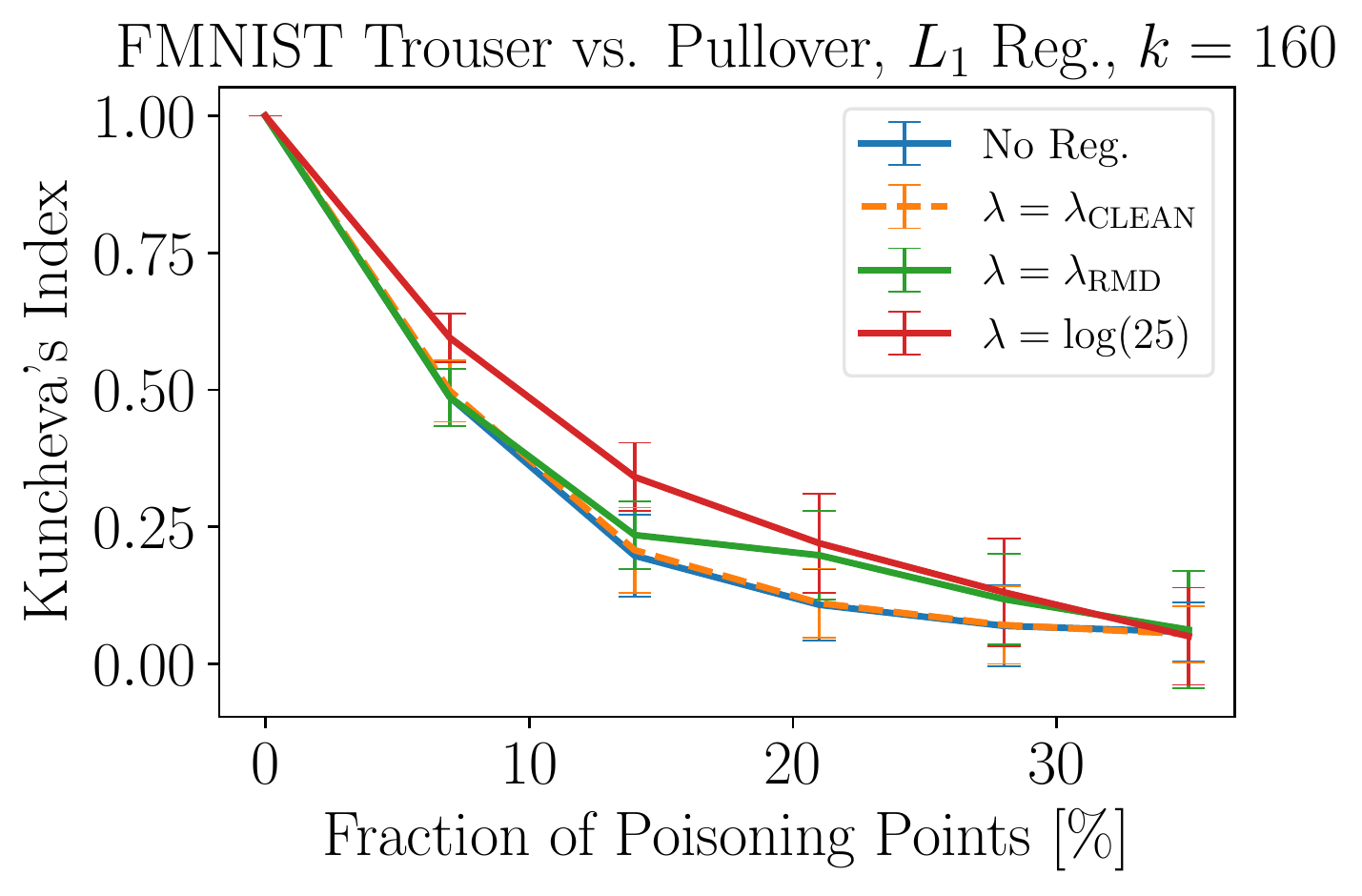}%
\label{fig:kunchevafmnist_l1_c}}
\subfloat[]{\includegraphics[width=2in]{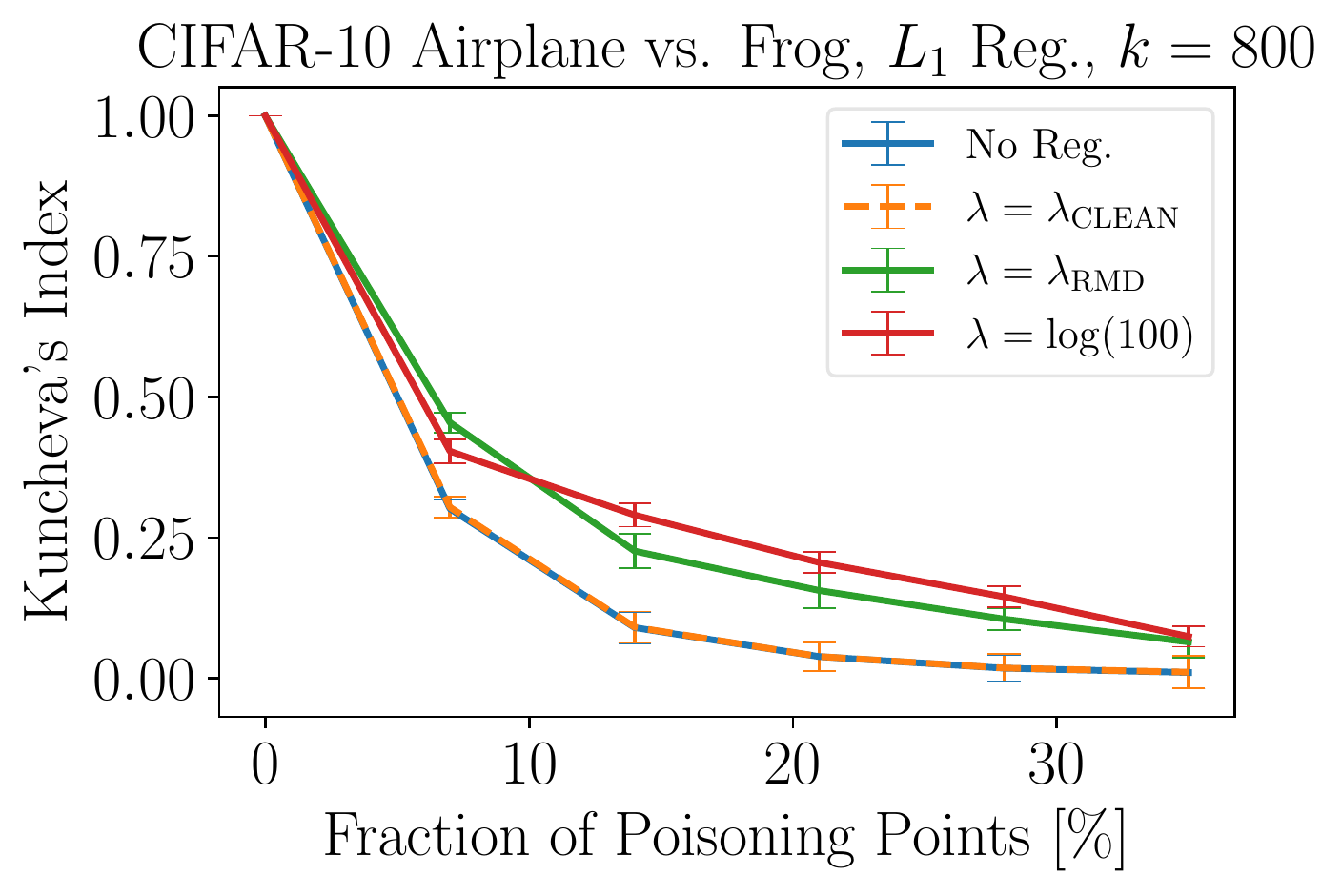}%
\label{fig:kunchevacifar_l1_c}}
\caption{Average Kuncheva's consistency index for the optimal attack against LR, using $L_2$ regularization (first and second rows); and using $L_1$  regularization (third to fifth rows), on MNIST (first column), FMNIST (second column) and CIFAR-10 (third column), for different numbers of {selected} features, $k$:  (a) $k=40$, (b) $k=80$, (c) $k=400$, (d) $k=80$, (e) $k=160$, (f) $k=800$, (g) $k=20$, (h) $k=40$, (i) $k=200$, (j) $k=40$, (k) $k=80$, (l) $k=400$, (m) $k=80$, (n) $k=160$, and (o) $k=800$.}
\label{fig:kuncheva_l2_2}
\end{figure*}

\begin{figure*}[!t]
\centering
\subfloat[]{\includegraphics[width=2.2in]{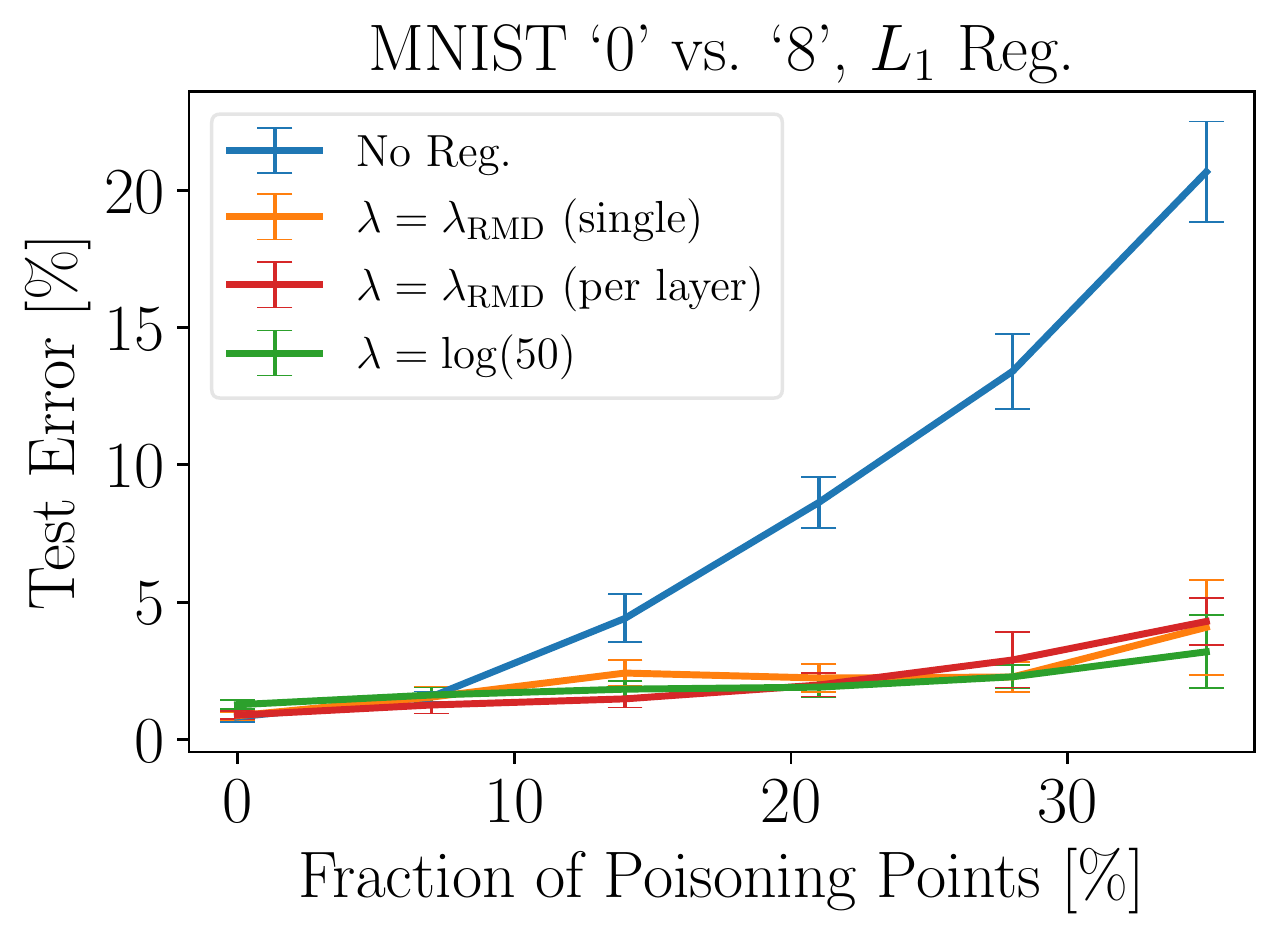}%
\label{fig:dnnopt_l1_a}}
\subfloat[]{\includegraphics[width=2.2in]{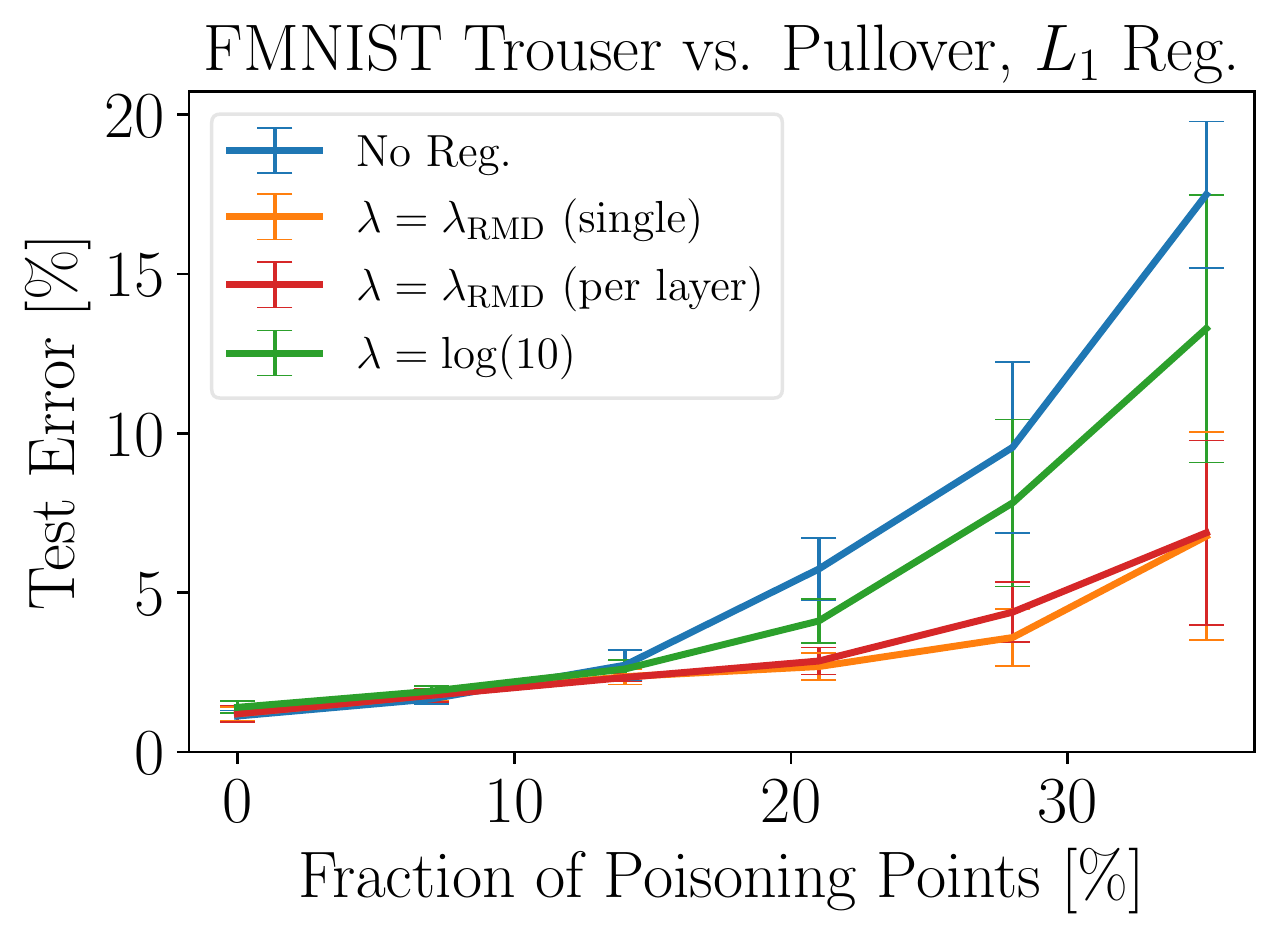}%
\label{fig:dnnopt_l1_b}}
\subfloat[]{\includegraphics[width=2.2in]{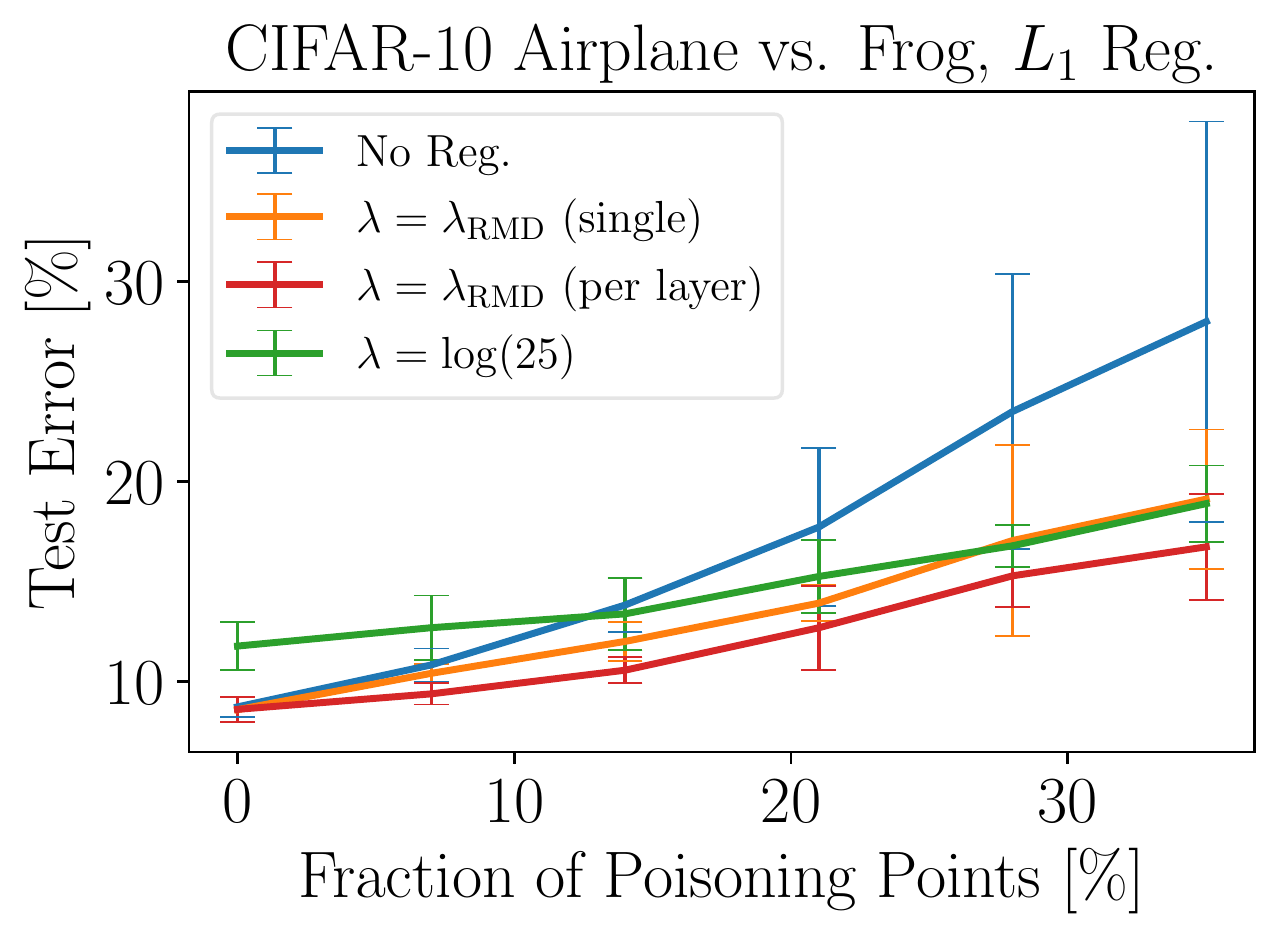}%
\label{fig:dnnopt_l1_c}}
\caption{Average test error for the optimal attack against the DNNs using $L_1$ regularization on (a) MNIST, (b) FMNIST, and (c) CIFAR-10.}
\label{fig:dnnopt_l1}
\end{figure*}

\begin{figure*}[!t]
\vspace{0.2cm}
\centering
\subfloat[]{\includegraphics[width=2.2in]{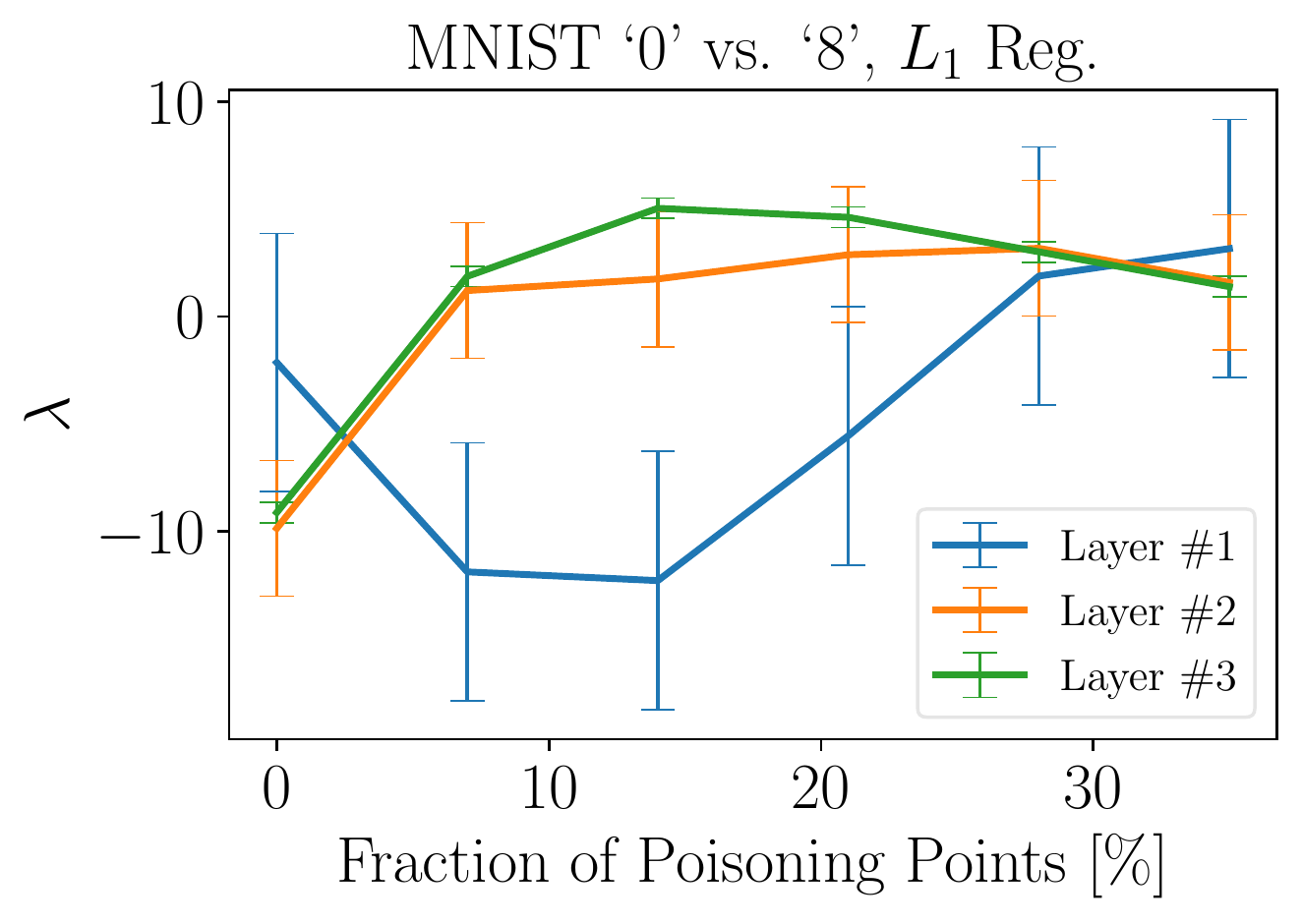}%
\label{fig:dnnlambd_pl_d}}
\subfloat[]{\includegraphics[width=2.2in]{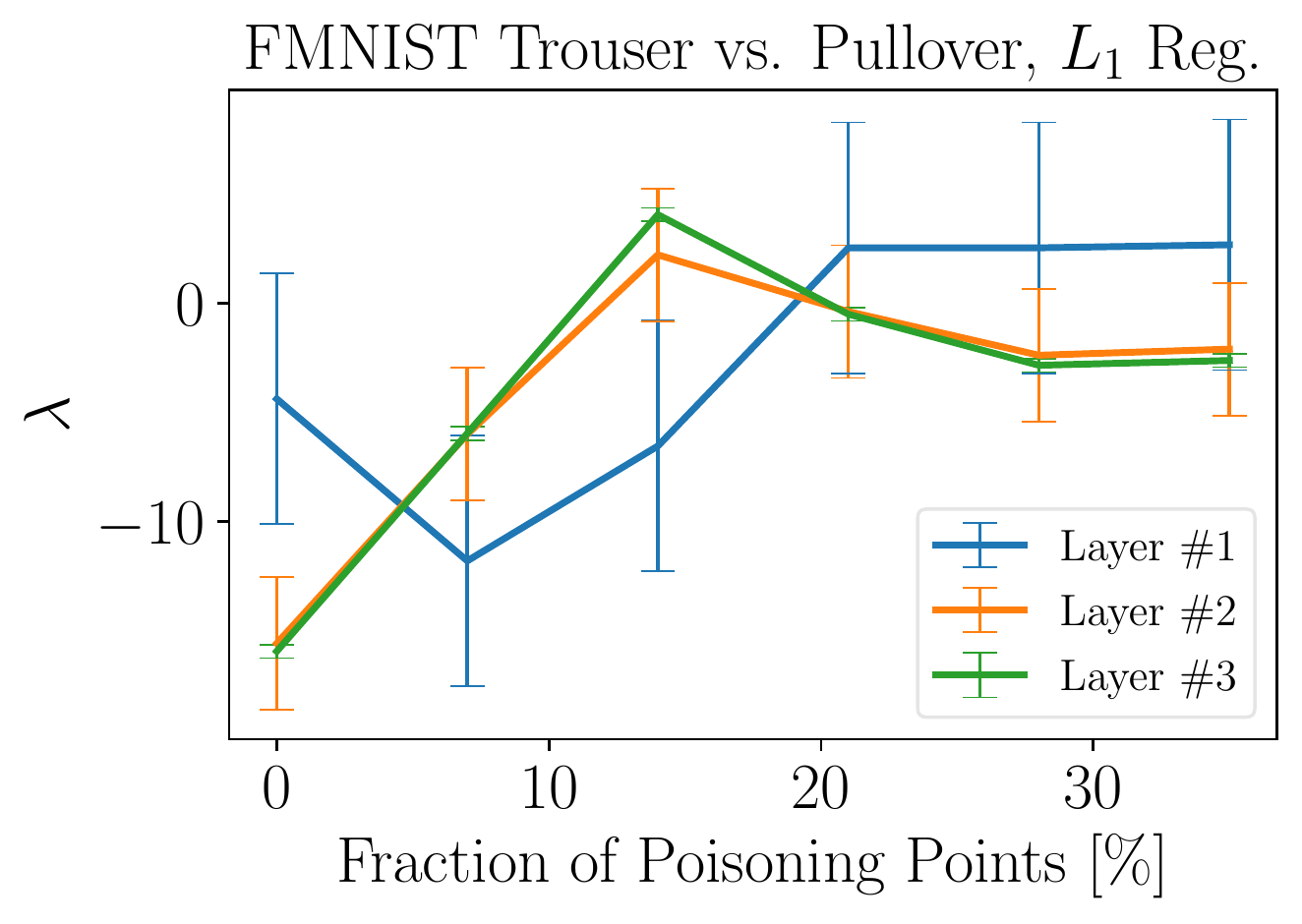}%
\label{fig:dnnlambd_pl_e}}
\subfloat[]{\includegraphics[width=2.2in]{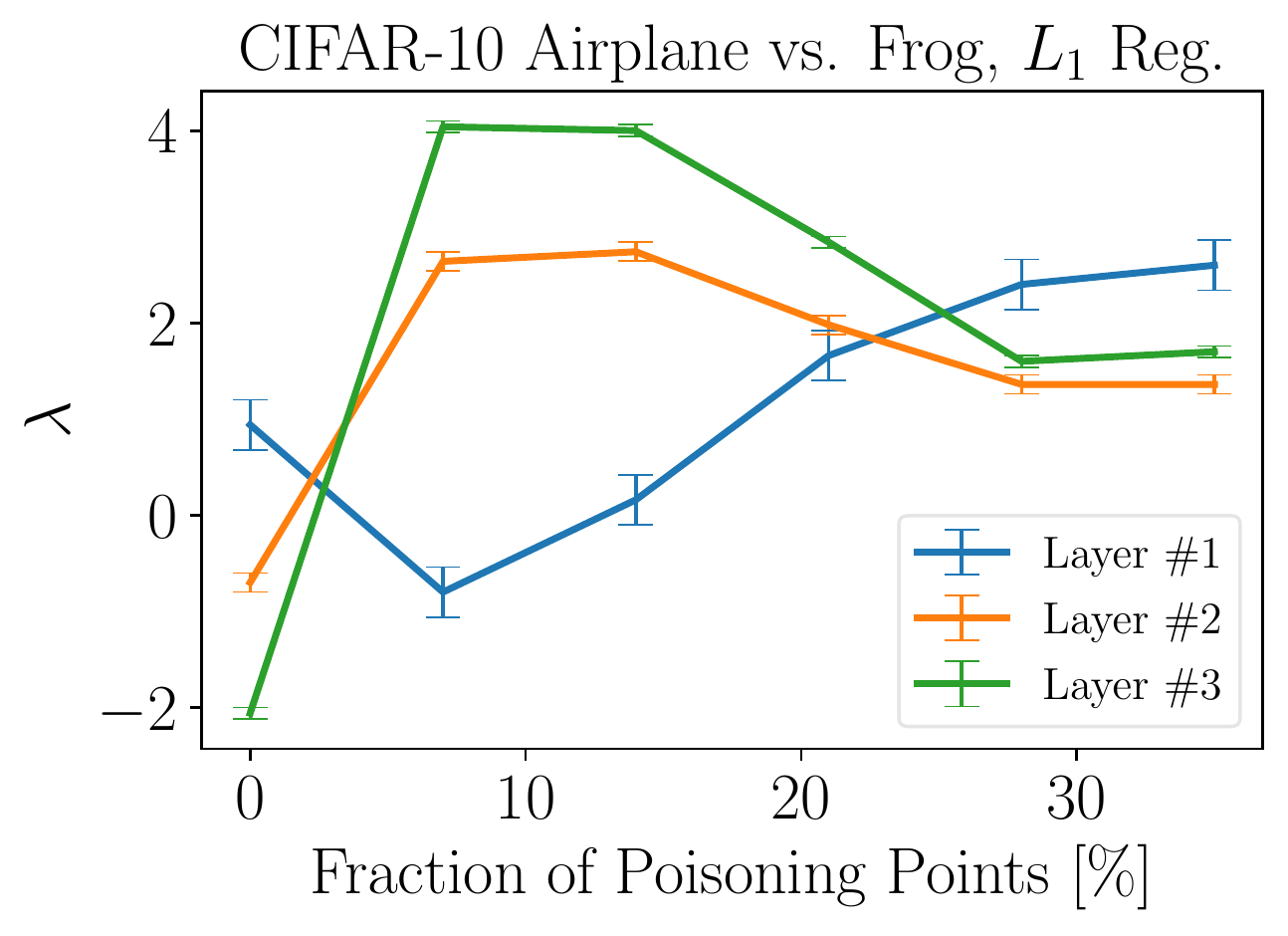}%
\label{fig:dnnlambd_pl_f}}
\caption{Average $\lambda$ learned with RMD for the DNNs using $L_1$ regularization with a different regularization hyperparameter at each layer. The left, central and right figures represent the results for MNIST, FMNIST and CIFAR-10, respectively. }
\label{fig:dnnlambd_pl_l1}
\end{figure*}

\begin{figure*}[!t]
\vspace{0.2cm}
\centering
\subfloat[]{\includegraphics[width=2.2in]{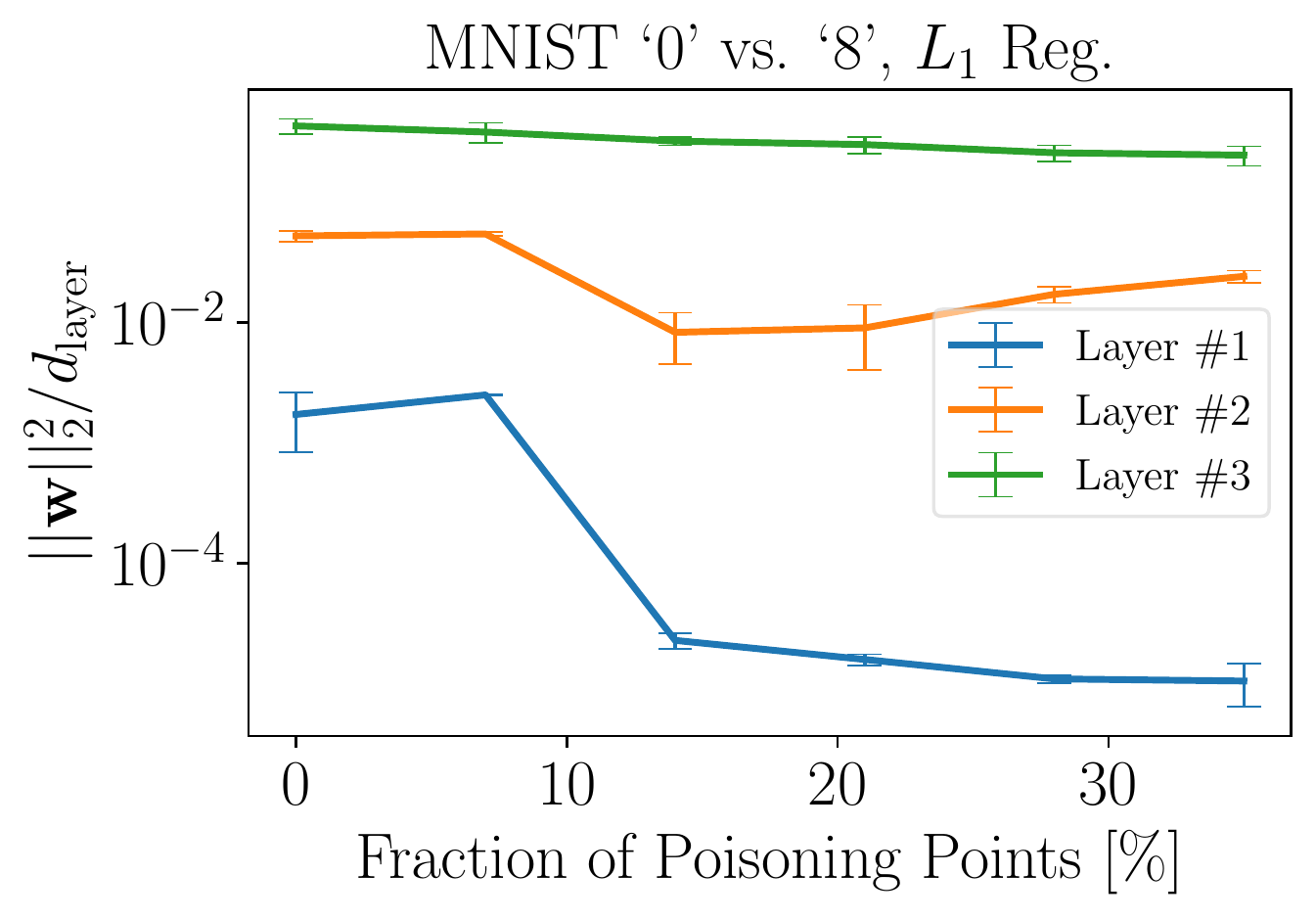}%
\label{fig:dnnlambd_pl_j}}
\subfloat[]{\includegraphics[width=2.2in]{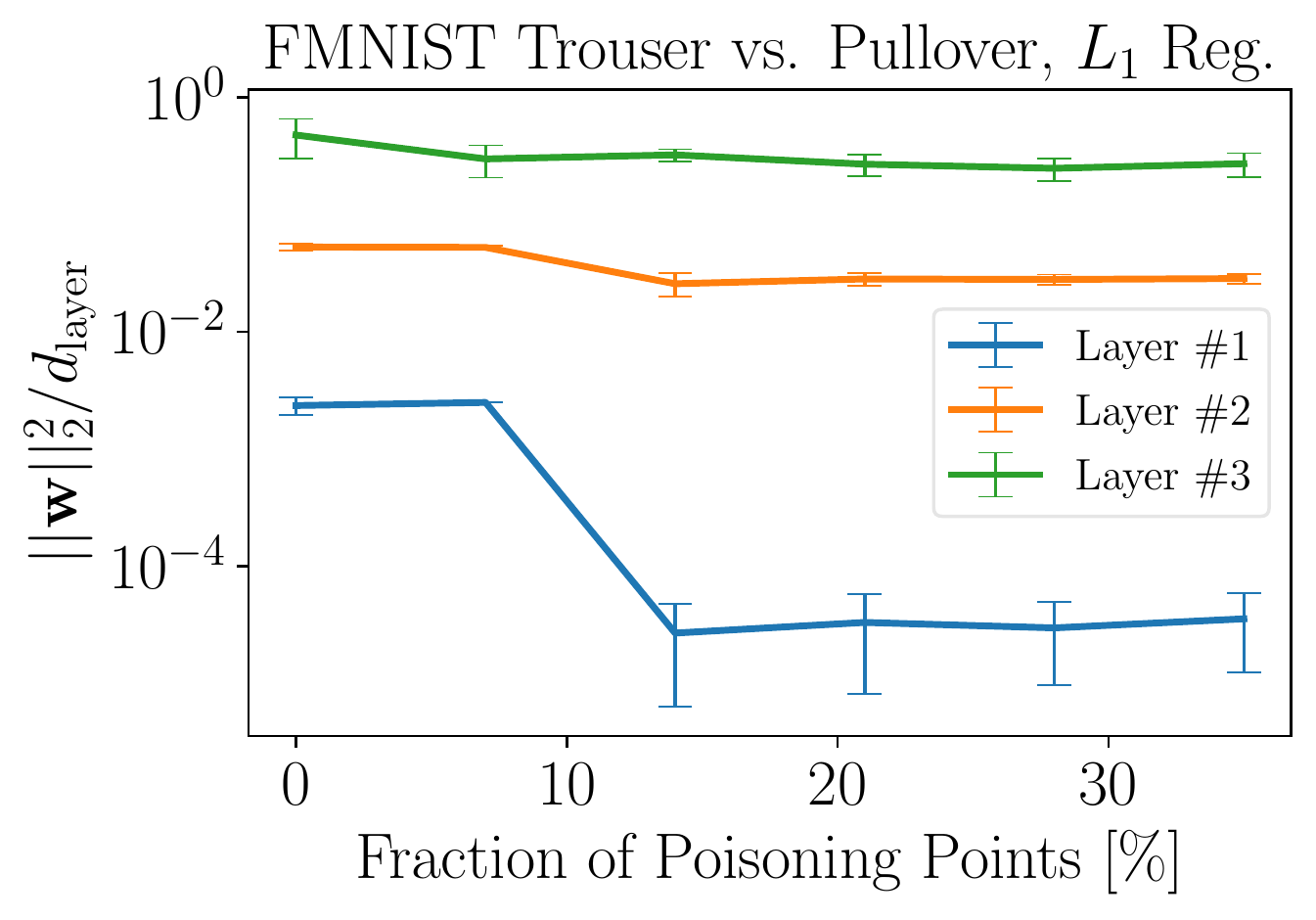}%
\label{fig:dnnlambd_pl_k}}
\subfloat[]{\includegraphics[width=2.2in]{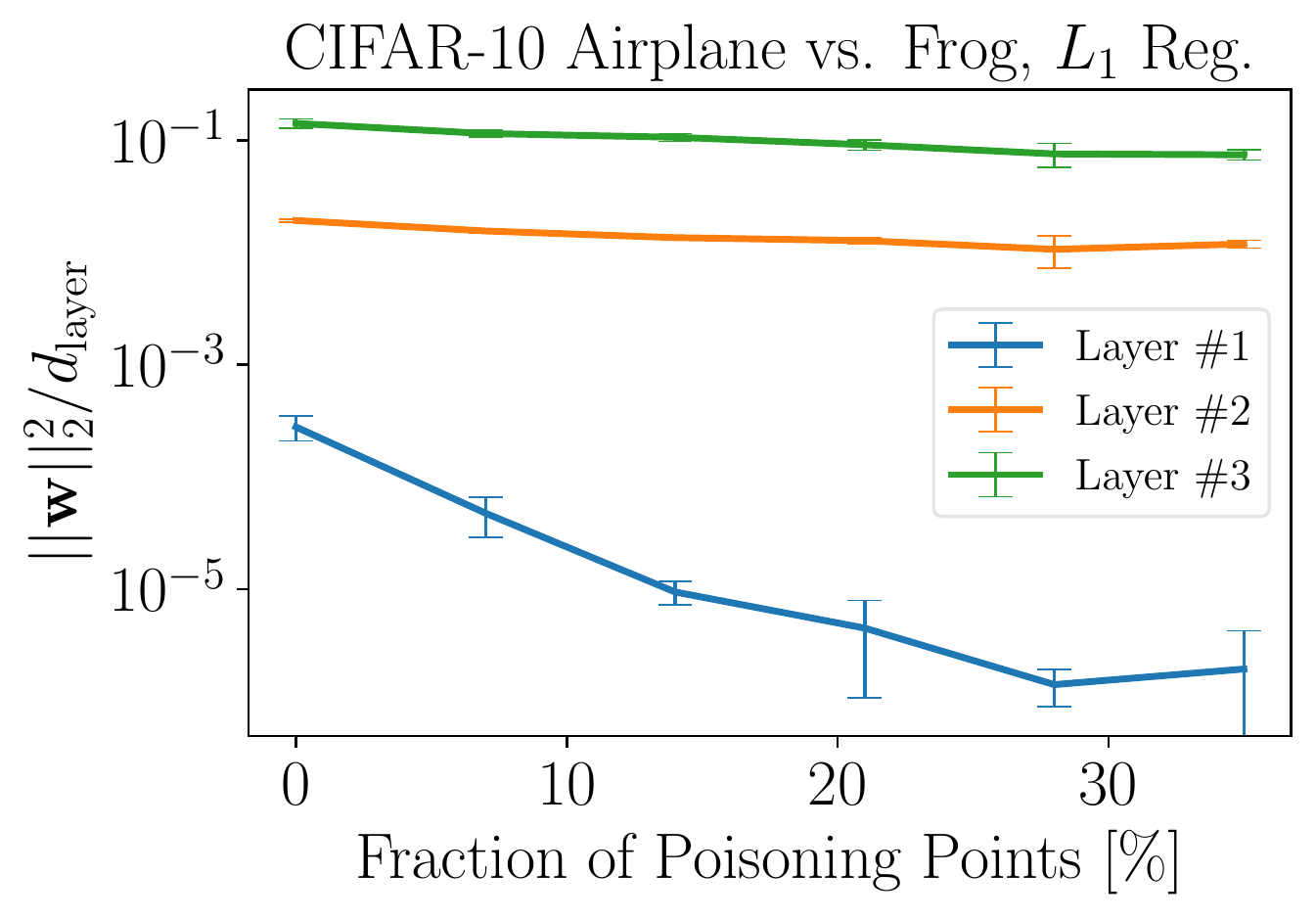}%
\label{fig:dnnlambd_pl_l}}
\\
\subfloat[]{\includegraphics[width=2.2in]{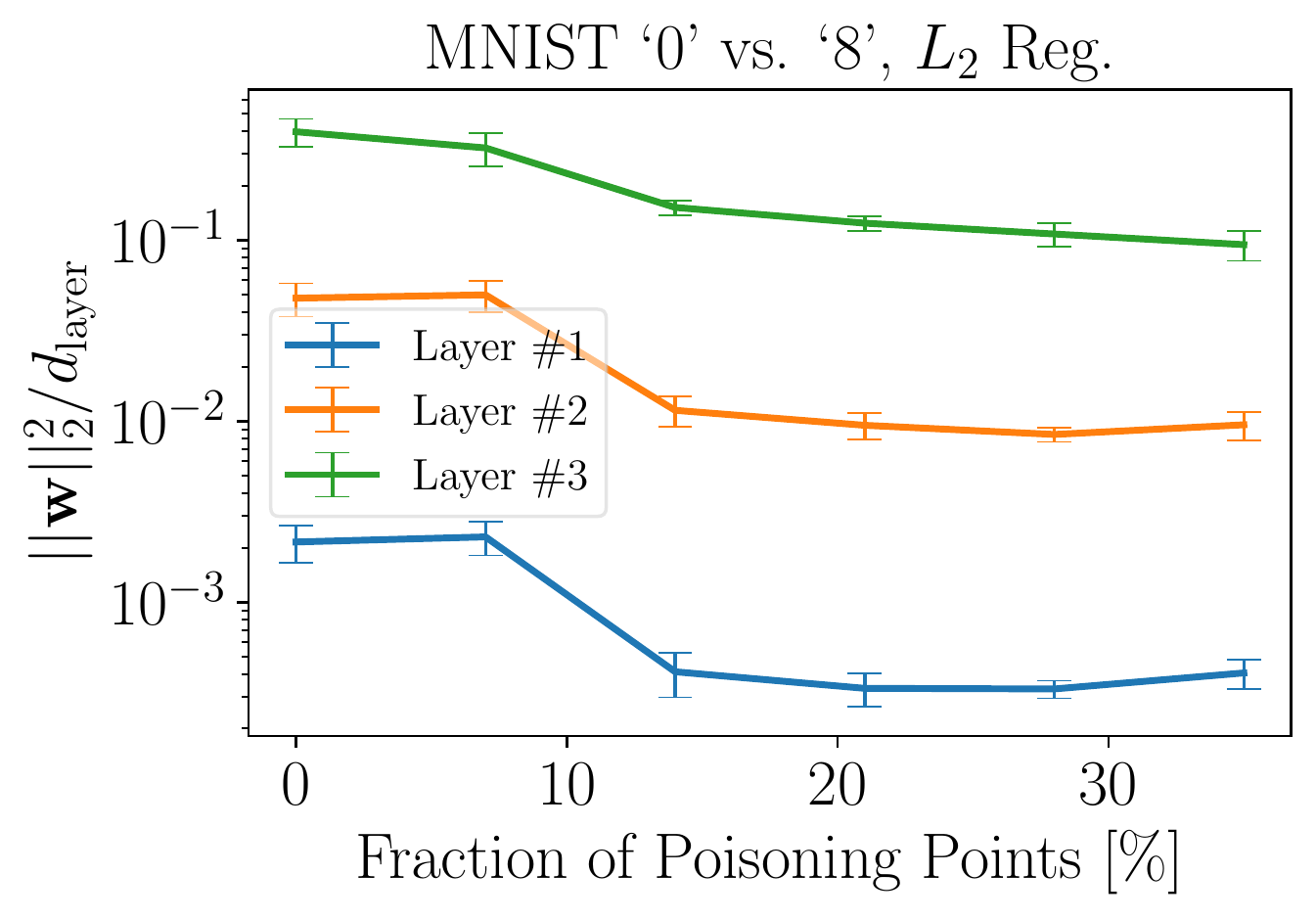}%
\label{fig:dnnlambd_pl_g}}
\subfloat[]{\includegraphics[width=2.2in]{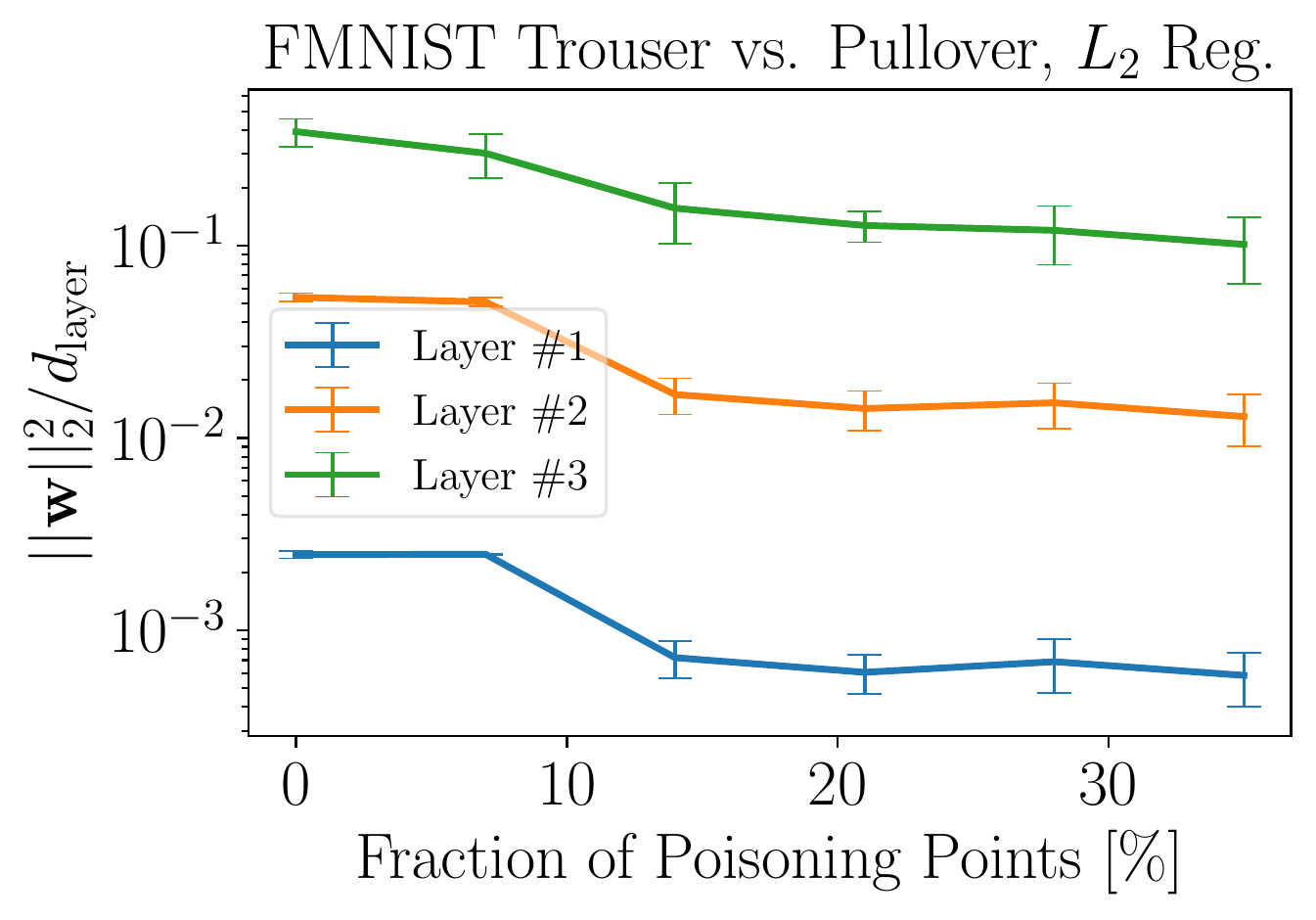}%
\label{fig:dnnlambd_pl_h}}
\subfloat[]{\includegraphics[width=2.2in]{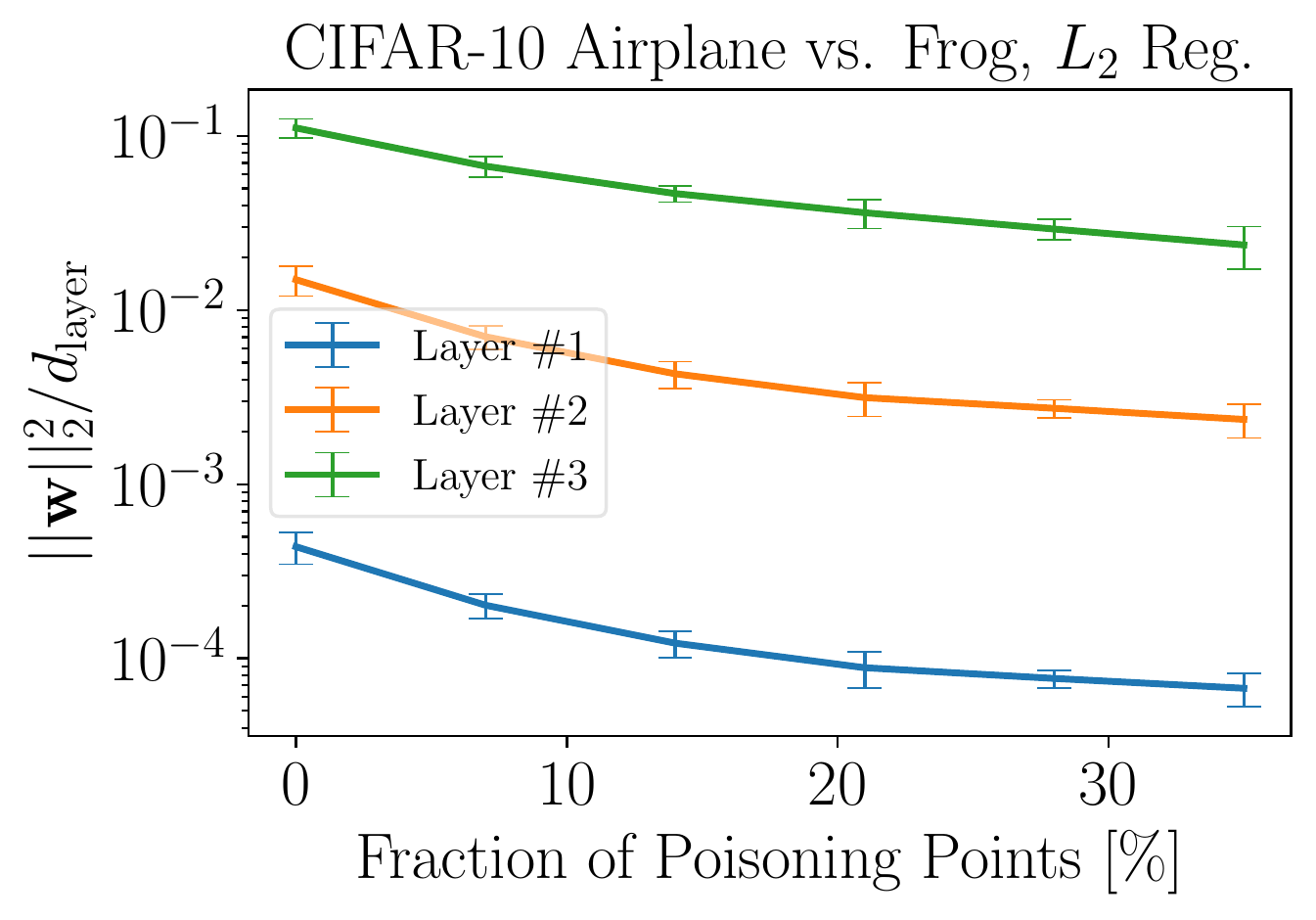}%
\label{fig:dnnlambd_pl_i}}
\caption{Average $||{\bf w}_\text{layer}||_2^2/d_\text{layer}$ for the DNNs using $L_2$ regularization (first row) and $L_1$ regularization (second row) with a different regularization hyperparameter at each layer, where $d_\text{layer}$ represents the number of parameters of the corresponding layer. This normalization allows comparing $||{\bf w}_\text{layer}||_2^2$ for each layer regardless of the number of the parameters. The first, second and third columns represent the results for MNIST, FMNIST and CIFAR-10, respectively. }
\label{fig:dnnlambd_pl_weights_l2}
\end{figure*}

\begin{figure*}[!h]
\centering
\subfloat[]{\includegraphics[width=1.8in]{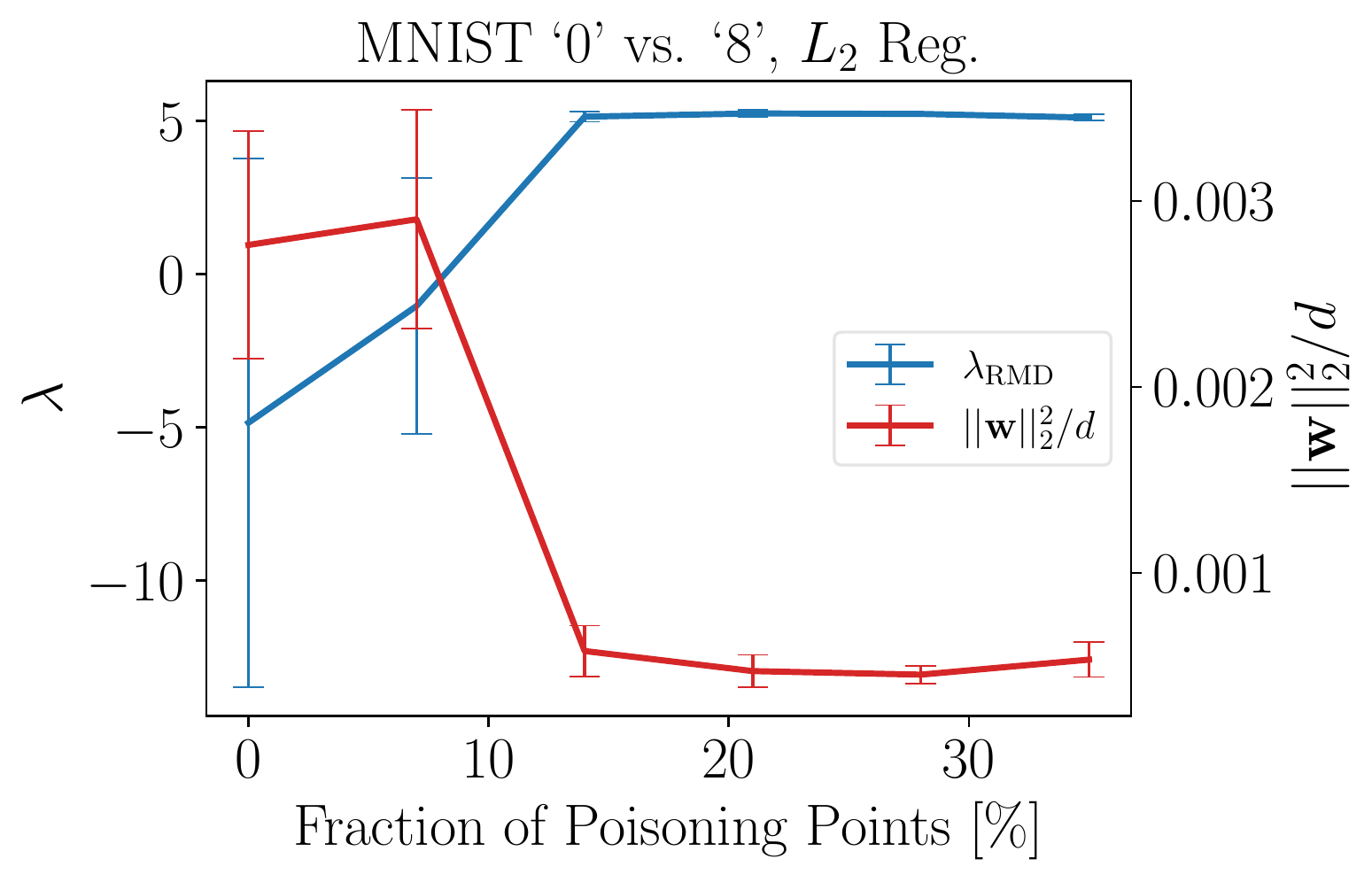}%
\label{fig:dnnlambd_a}}
\subfloat[]{\includegraphics[width=1.8in]{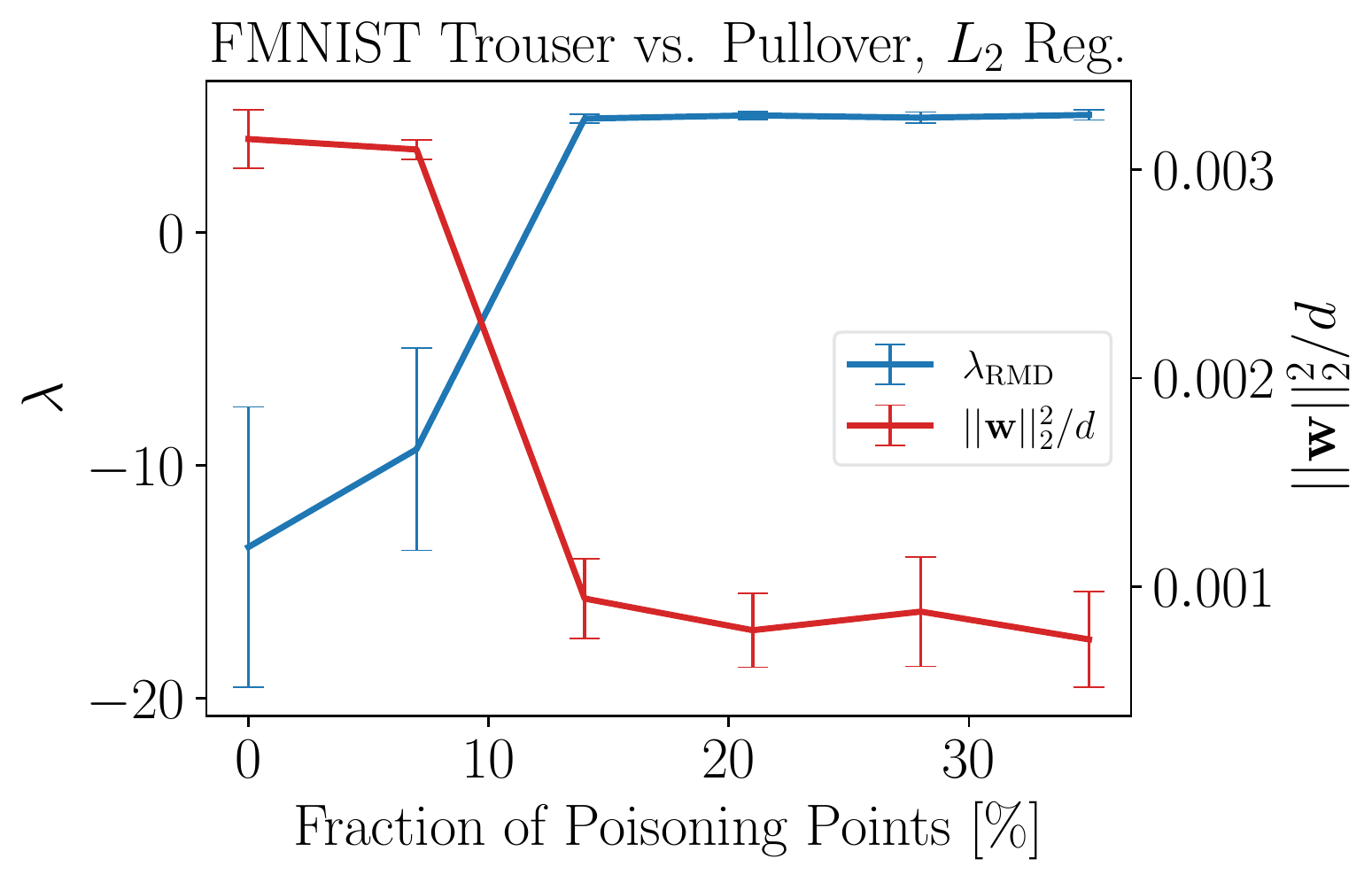}%
\label{fig:dnnlambd_b}}
\subfloat[]{\includegraphics[width=1.8in]{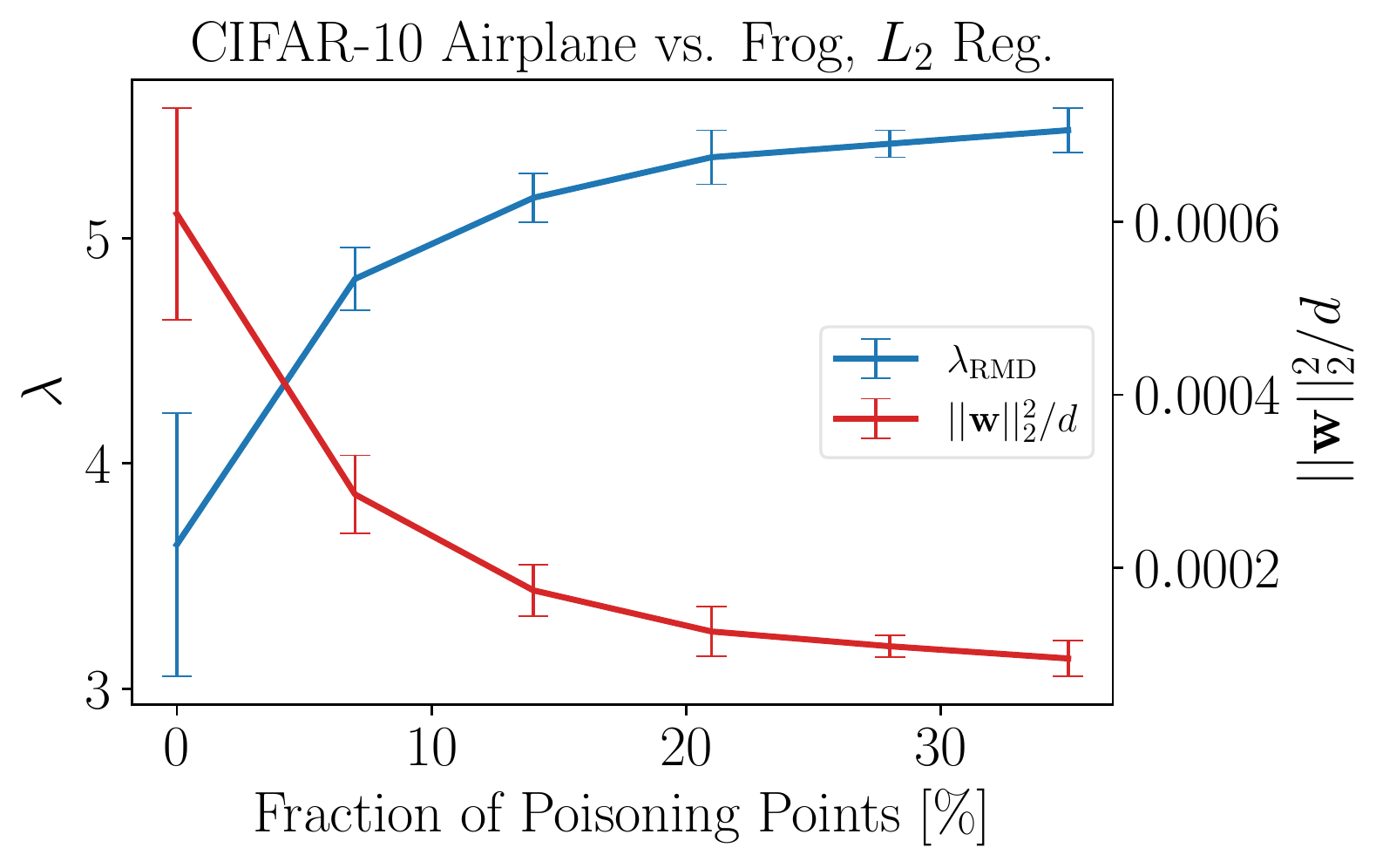}%
\label{fig:dnnlambd_c}}
\\
\vspace{-0.1cm}
\subfloat[]{\includegraphics[width=1.8in]{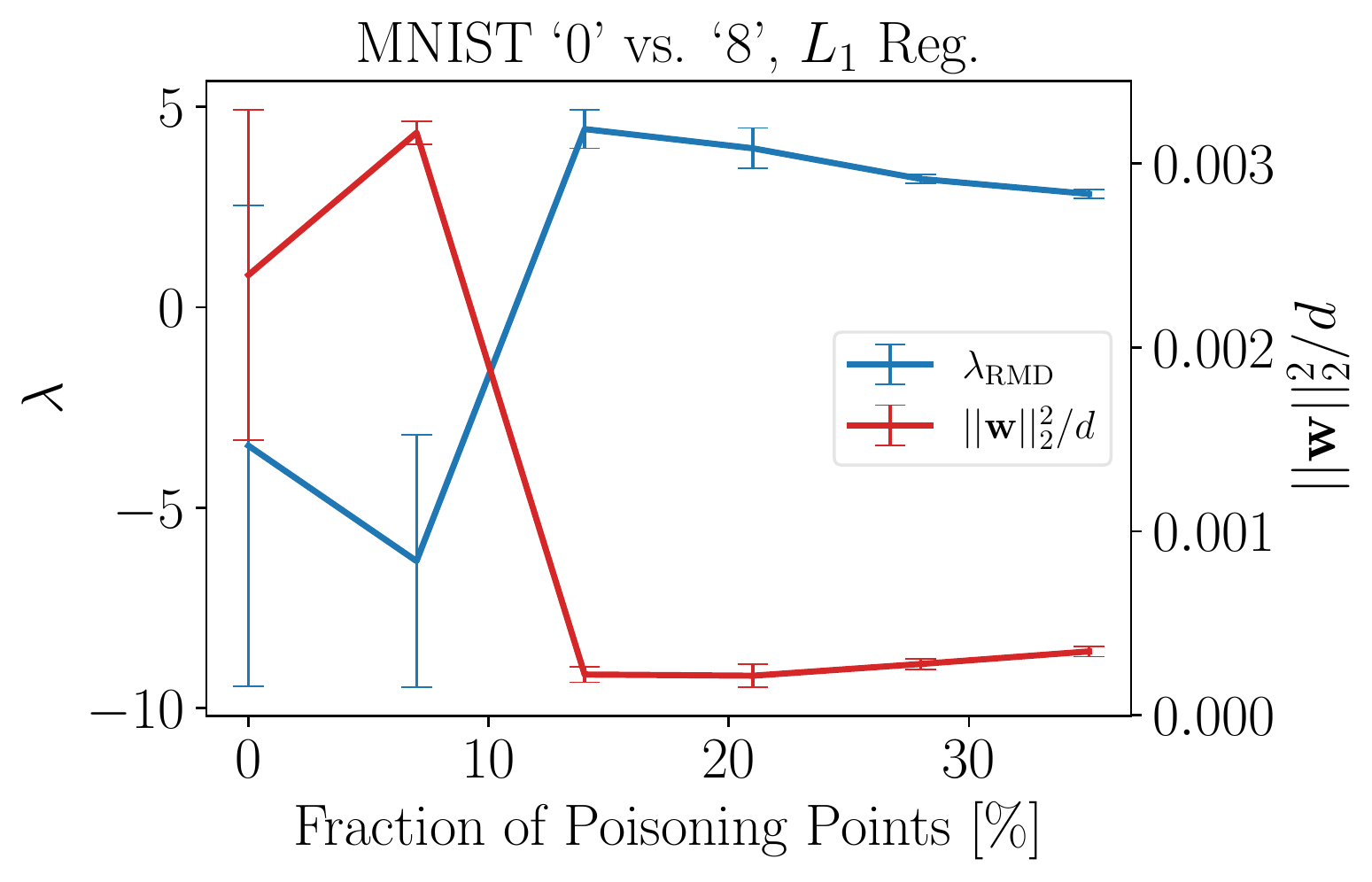}%
\label{fig:dnnlambd_d}}
\subfloat[]{\includegraphics[width=1.8in]{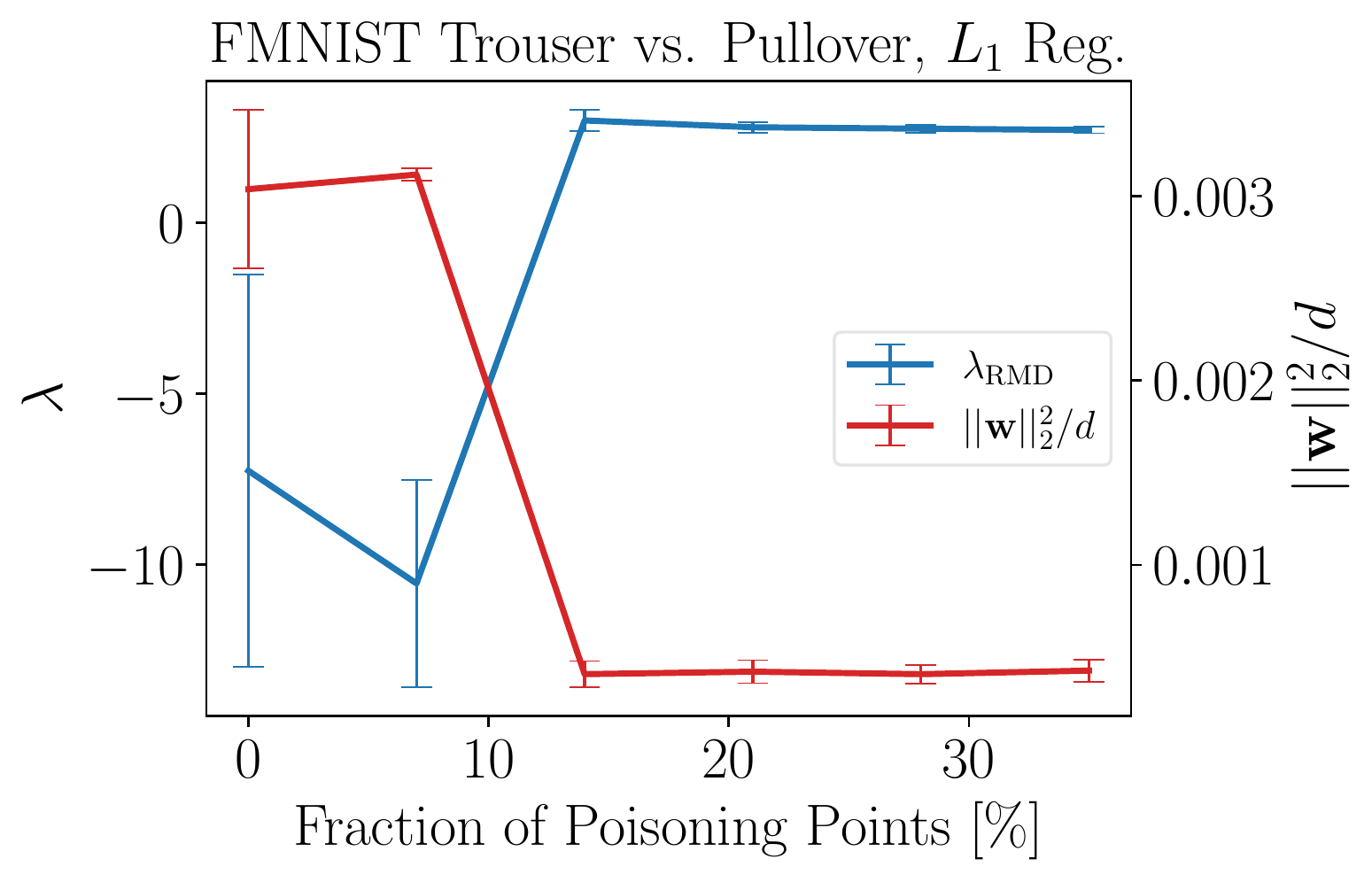}%
\label{fig:dnnlambd_e}}
\subfloat[]{\includegraphics[width=1.8in]{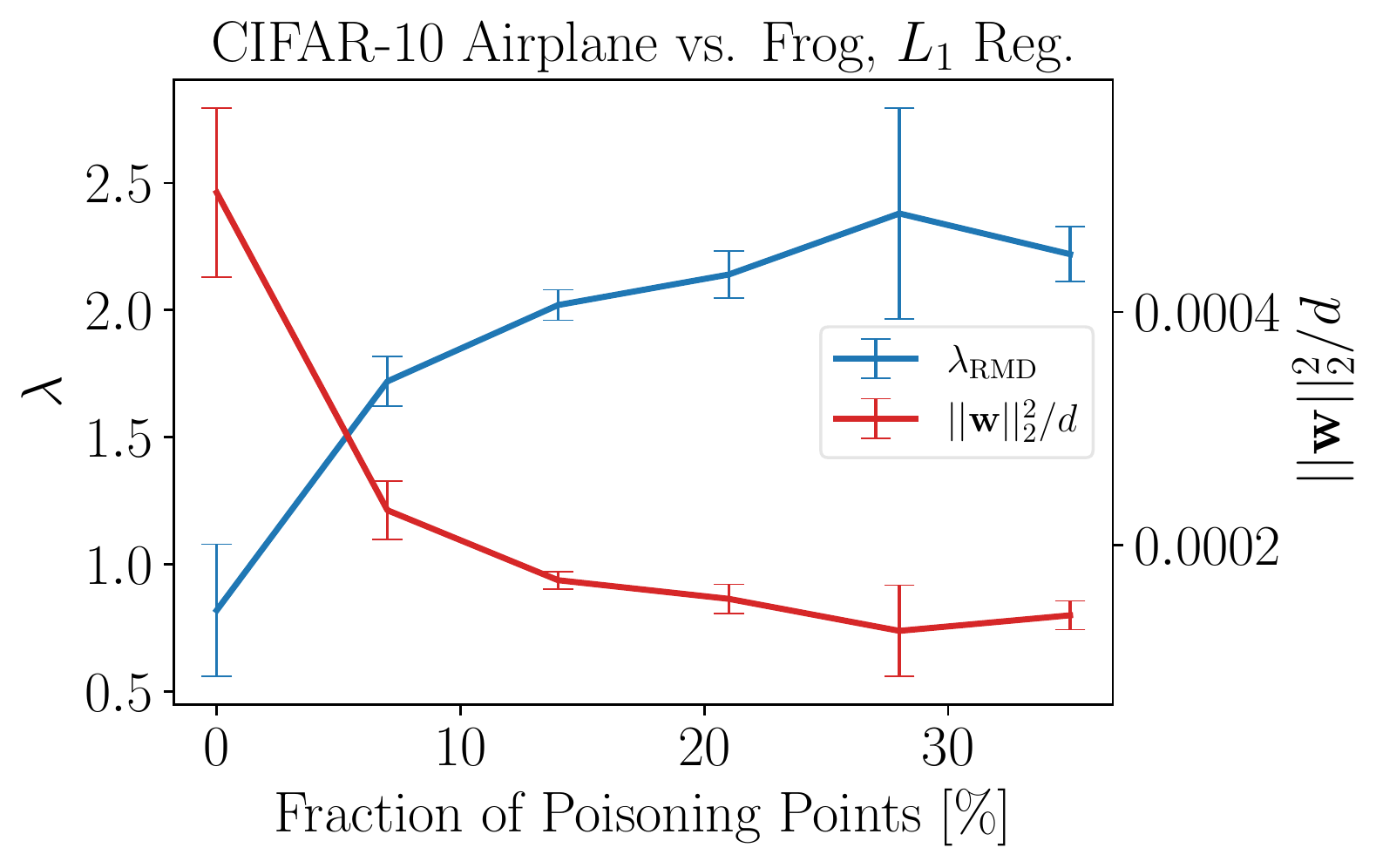}%
\label{fig:dnnlambd_f}}
\caption{Average $\lambda$ and average $||{\bf w}||_2^2/d$ for the optimal attack against the DNNs using a single regularization hyperparameter, where $d$ is the number of parameters of the model. The first row represents the case of $L_2$ regularization on (a) MNIST, (b) FMNIST, and (c) CIFAR-10. The second row represents the case of $L_1$ regularization on (d) MNIST, (e) FMNIST, and (f) CIFAR-10.}
\label{fig:dnnlambd}
\end{figure*}

Here we show additional results that complement and are coherent with the ones discussed in Sect.~\ref{sec:experiment}.

In Fig.~\ref{fig:lrlambd_2} we show the value of $\lambda$ learned and the norm of the model's parameters divided by the number of parameters, $||{\bf w}||^2_2/d$, as a function of the fraction of poisoning points injected, for LR using $L_2$ and $L_1$ regularization on FMINST and CIFAR-10 (see also Fig.~\ref{fig:lrlambd}). We observe that the regularization hyperparameter increases, and then, saturates as we increase the fraction of poisoning points. Comparing $L_2$ and $L_1$, we observe that both regularization techniques provide similar mitigation effects against the attack. %

In Fig.~\ref{fig:lr_val_2} we show the sensitivity analysis of the size of the validation set for LR (see also Fig.~\ref{fig:lr_val}). When $\lambda$ is learned using $L_1$ regularization, for MNIST and FMNIST the test error decreases when the validation set is smaller, whereas for CIFAR-10, the opposite occurs. This shows that having a larger validation set is not always advantageous. Our results show that, with this interplay between the learner and the attacker, the net benefit for the learner depends on the specific classification task, the size of the validation set and the attack strength. However, it is also important to note that, across all experiments, there is a clear benefit for using regularization to mitigate the impact of the attack in all cases and, especially, for strong attacks.

The results for Kuncheva's consistency index for LR using $L_1$ regularization on MNIST, FMNIST and CIFAR-10 are shown in Fig.~\ref{fig:kuncheva_l2_2} (which can be compared with Fig.~\ref{fig:kuncheva_l2}). 
We observe that, in all cases, the consistency index decreases with the ratio of poisoning. This means that, to succeed, the attack naturally modifies the importance of the features of the training set, so that the poisoned model pays more attention to less relevant features. It is also clear that if the model is not regularized, the features selected are less consistent, and regularization helps to increase the feature stability under poisoning. For $\lambda_\text{RMD}$, it is generally bounded between the cases of no regularization and large value of $\lambda$, showing that the algorithm sacrifices some feature stability to decrease the test error. Compared to $L_1$, $L_2$ regularization provides greater feature stability when using a large regularization hyperparameter.

Fig.~\ref{fig:dnnopt_l1} shows the test error for the optimal attack against the DNNs using $L_1$ regularization (see also Fig.~\ref{fig:dnnopt_l2}). These results are consistent with those obtained for the case of $L_2$ regularization and LR. When there is no regularization, the algorithm is vulnerable to the poisoning attack and its test error increases significantly. For a large value of $\lambda$, the algorithm's performance remains quite stable, but the clean error is higher. For $\lambda_{\text{RMD}}$ the test error increases only moderately, and the results when using a single hyperparameter or a different hyperparameter at each layer are very similar. From Fig.~\ref{fig:dnnopt_l1} we can see that when there is no attack, the test error for $\lambda_{\text{RMD}}$ is smaller than in the other two cases. Although over-regularizing may be appealing to make the algorithm more robust to poisoning, the performance in the absence of attacks may be significantly worse. Learning $\lambda$ evidences this trade-off. On the other hand, it is evident that the mitigating effect of regularization is more prominent in the case of DNNs. As the capacity of the DNN (compared to LR) is higher, the attackers can have more flexibility to manipulate the decision boundary. Hence, having regularization in place, in combination with the trusted validation set, is even more important in the case of the DNNs.

Fig.~\ref{fig:dnnlambd_pl_l1} shows the value of $\lambda$ when using a different regularization term at each layer, for $L_1$ regularization (refer also to Fig.~\ref{fig:dnnlambd_pl_l2}). We observe that the $\lambda$ learned for the second and output layers increases faster than the one for the first layer and, for FMNIST and CIFAR-10, this increase is faster for the first hidden layer from $20\%$ of poisoning. This suggests that the latter layers can be more vulnerable to the attacks. These poisoning attacks try to produce more changes in those layers and, at the same time, the network tries to resist those changes by increasing the value of the corresponding regularization hyperparameters. On the other hand, when the attacks are very strong, their impact appear more uniform across all layers in the DNN, based on the values of $\lambda$ learned for each layer.

In Fig.~\ref{fig:dnnlambd_pl_weights_l2} we can observe that, as in the case of LR, the value of the regularization hyperparameters of the DNNs is also related to the norm of the weights divided by the number of parameters for each layer in the DNN.

Finally, for the sake of completeness, Fig.~\ref{fig:dnnlambd} shows the value of $\lambda$ learned and the norm of the model's parameters divided by the number of parameters, $||{\bf w}||^2_2/d$, as a function of the fraction of poisoning points injected, for the DNNs when using a single regularization term for $L_2$ and $L_1$ regularization. These results are coherent with the ones for LR (Fig.~\ref{fig:lrlambd_2} and Fig.~\ref{fig:lrlambd}). We observe that the regularization hyperparameter increases, and then, saturates as we increase the fraction of poisoning points. However, using a different regularization at each layer can be more insightful to understand at which layer the attack focuses more.

}

%

%








\end{document}